\pdfoutput=1
\documentclass[11pt]{article}

\usepackage[margin=1in]{geometry}
\usepackage{natbib}
\usepackage{amsthm}
\usepackage{pdftexcmds}
\newtheorem{theorem}{Theorem}
\newtheorem{lemma}[theorem]{Lemma}
\newtheorem{proposition}[theorem]{Proposition}

\newtheorem{definition}[theorem]{Definition}

\newtheorem{example}{Example}[section]
\newenvironment{keywords}{\par\vspace{0.5em}\noindent\textbf{Keywords:} }{\par}
\usepackage{hyperref}
\usepackage{url}

\usepackage{xcolor}         
\usepackage{framed}
\usepackage{pdfpages}
\usepackage{graphicx}
\usepackage{float}
\usepackage{amsmath}
\usepackage{bbold,color}
\usepackage{amssymb}
\usepackage{MnSymbol}
\usepackage{wasysym}    

\usepackage{subfig}
\newsavebox{\measurebox}

\usepackage{amsbsy}
\usepackage{parskip}
\usepackage{algorithm}
\usepackage{algpseudocode}
\usepackage{color}
\usepackage[normalem]{ulem}
\usepackage{multirow}
\usepackage{graphicx}
\usepackage{cancel}
 \usepackage{booktabs}
\usepackage[nice]{nicefrac}
\usepackage{nomencl}
\makenomenclature

\newcommand{\scKL}{\textsc{KL}}

\newcommand{\Ind}{\mathbb{I}}

\newcommand{\given}{\,|\,}   
\usepackage{personal}

\newtheorem{assumption}{Assumption}[section]

\begin{document}

\title{Generalized Regret Analysis of Thompson Sampling using Fractional Posteriors}

\author{%
Prateek Jaiswal\thanks{Department of Quantitative Methods, Daniels School of Business, Purdue University, West Lafayette, IN 47906, USA. Email: \texttt{jaiswalp@purdue.edu}}
\and
Debdeep Pati\thanks{Department of Statistics, University of Wisconsin-Madison, Madison, WI 53706, USA. Email: \texttt{dpati2@wisc.edu}}
\and
Anirban Bhattacharya\thanks{Department of Statistics, Texas A\&M University, College Station, TX 77843, USA. Email: \texttt{anirbanb@tamu.edu}}
\and
Bani K. Mallick\thanks{Department of Statistics, Texas A\&M University, College Station, TX 77843, USA. Email: \texttt{bmallick@tamu.edu}}
}
\date{}

\maketitle

\begin{abstract}
Thompson sampling (TS) is one of the most popular and earliest algorithms to solve stochastic multi-armed bandit problems. We consider a variant of TS, named $\alpha$-TS, where we use a fractional or $\alpha$-posterior ($\alpha\in(0,1)$) instead of the standard posterior distribution.  To compute an $\alpha$-posterior, the likelihood in the definition of the standard posterior is tempered with a factor $\alpha$. 
For $\alpha$-TS we obtain both instance-dependent \(\mathcal{O}\left(\sum_{k \neq i^*} \Delta_k\left(\frac{\log(T)}{C(\alpha)\Delta_k^2}  + \frac{1}{2} \right)\right)\) and instance-independent \(\mathcal{O}(\sqrt{KT\log K})\) frequentist regret 
bo- unds under very mild conditions on the prior and reward distributions,
where $\Delta_k$ is the gap between the true mean rewards of the $k^{th}$ and the best arms, and $C(\alpha)$ is a known constant. Both the sub-Gaussian and exponential family models satisfy our general conditions on the reward distribution. Our conditions on the prior distribution ~can be easily satisfied by a density  that is positive, continuous, and bounded. We also establish another instance-dependent regret upper bound 
that matches (up to constants) to that of improved UCB~\citep{Auer2010}. 
Our regret analysis carefully adapts and combines recent theoretical developments in the non-asymptotic concentration analysis and Bernstein-von Mises type results for the $\alpha$-posterior distribution.
  Moreover, our analysis does not require additional structural properties such as closed-form posteriors or conjugate priors.
\end{abstract}


\section{Introduction}

 The stochastic multiarmed bandit (MAB) problem is one of the most popular and oldest frameworks for sequential decision-making. It is widely used to model a variety of sequential decision-making problems such as add placement, website optimization, mobile health, packet routing, clinical trials, revenue management, etc.~\cite[Chapter 1]{Bubeck2012}. The objective of this problem is to maximize the total reward collected by trying out different arms sequentially. In each trial, an  agent (or an algorithm) chooses an arm based on the information collected from past trials/experiences to receive a random reward from the chosen arm. While choosing an arm, the agent can either use the observed information to try the arm with the highest chances of generating maximum immediate reward (exploitation) or take a risk and try less explored arms to expect that it may produce a higher reward (exploration). MAB models this exploration/exploitation trade-off, which is intrinsic to many sequential decision-making problems. Algorithms for solving the MAB problem are designed to address this exploration/exploitation dilemma. 

Upper Confidence Bound (UCB)~\citep{lai1985asymptotically,lai1987adaptive,Auer2002} and Thompson Sampling (TS)~\citep{THOMPSON1933} are two popular classes of algorithms to solve stochastic MAB problems. In this work, we study TS that uses a Bayesian heuristic to address the exploration/exploitation trade-off in MAB.  In TS, we first posit a prior distribution on the parameters of the reward distribution for each arm and then update the prior belief by computing a posterior distribution using sequentially observed rewards from the previous trials. TS models the exploration/exploitation trade-off by capturing the uncertainty in mean rewards through the posterior distribution. More specifically, TS samples a reward distribution for each arm from the posterior distribution at each step and picks an arm with the best mean reward. In essence, TS randomly selects an arm according to the probability of it being optimal. Therefore, unlike UCB algorithms, TS computes a randomized arm selection policy. Intuitively, TS encourages exploration because there is always a positive posterior probability at each step of TS to sample a set of reward distributions for which a less explored arm would have the highest mean reward. {However, this probability decreases as more rewards are generated from the optimal arm. } 

The theoretical performance of these algorithms is typically studied by computing an upper bound on the cumulative regret, where the regret is defined as the expected loss in reward when playing the arm selected by the algorithm instead of the best oracle arm. The regret bounds are further classified as instance-dependent and independent. A regret bound is called instance-dependent when it depends on the difference between the true mean rewards for each arm pair, and this difference is considered to be a constant. In contrast, the instance-independent bound has no such dependence; therefore, the true mean rewards for two arms can be arbitrarily close. 
In this work, we will focus on computing both types of regret bounds for TS-type algorithms. 

The germinal work by~\cite{agarwal2017effective} derived both  instance-dependent and independent regret bounds for TS that closely match the respective lower bounds for such problems. However, their analysis exploited a specific analytic form of the sampling (posterior) distribution. In a similar vein,~\cite{Kaufmann2013} derived instance-dependent bound for exponential family models with Jeffrey's prior, which also have an analytic form of the posterior. Recall that the TS algorithm is more versatile and can be used with any prior and reward combinations (\cite{li2012open,chapelle2011empirical,urteaga2018nonparametric,hong22b}). Therefore, \cite{gopalan2013thompson} analyzes TS with no closed-form posterior distribution but under the assumption that the prior and reward distributions have discrete support.~\cite{mazumdar20a} further generalizes the analysis of TS and derives instance-dependent regret bound for log-concave and Lipschitz smooth prior and reward distributions. Their proof technique uses specific properties of log-concave and Lipschitz smooth density functions (e.g., see the proof of Lemma 15 in~\cite{mazumdar20a}), which we find challenging to generalize to other classes of prior and reward distributions.

\textbf{Our contributions:} This work contributes to this growing body of work towards generalizing the analysis of the TS algorithm. We derive regret bounds for the TS algorithm with an $\alpha$-posterior (or fractional posterior) distribution ~\citep{bhattacharya2019bayesian}. We denote this algorithm as $\alpha$-TS. Our analysis of $\alpha$-TS identifies regularity conditions on the reward and prior distribution to derive both (near optimal) instance-independent and dependent regret bounds without assuming any analytical form of the posterior distribution. Our general condition on the reward distribution includes both sub-Gaussian and exponential family reward distributions. Moreover, our theory works for a general class of prior distributions under very mild assumptions, and in particular, one does not need to fix it to an exponential family or a log-concave distribution.  Our condition on the prior distribution requires the prior measure of a ball centered at the true parameter to be bounded above and below by a term that scales with the radius of the ball at the same rate. This type of prior condition was first designed  in the Bayesian posterior concentration literature by~(\cite{GGV}) and are satisfied by a large class of finite-dimensional distributions. The class of priors spanned by this condition is general enough to include Gaussian mixtures or other complex priors as considered in a few recent works~(\cite{hong22b,urteaga2018nonparametric}). Notably, our work also addresses an open problem of generalizing the regret analysis of TS for a broader class of priors, as noted by~\cite[page 69, line 2]{Russo20}. Moreover, we also establish an instance-dependent regret bound for $\alpha$-TS that matches to that of the improved UCB~\citep{Auer2010}. This bound has $\log(\text{Constant}\times T \Delta_k^2)$ term in the numerator instead of just $\log(T)$, which guarantees a better control on the expected regret with respect to $T$ for smaller $\Delta_k$. 

To compute an $\alpha$-posterior distribution, the likelihood in the Bayes formula is tempered with a factor $\alpha\in(0,1)$, unlike the standard posterior where $\alpha=1$.
\nomenclature[a]{$\alpha\in (0,1)$}{The temperature parameter in a fractional posterior}
Smaller values of $\alpha$ inflates the posterior variance while the larger values have an opposite effect. Introducing $\alpha$ provides a principled way to control the amount of exploration through posterior variance. Our work shows that TS algorithm can achieve state-of-the-art regret guarantees even with fractional posterior. We would also like to note that we also use $\alpha$-posteriors as a convenient technical device
to obtain regret bounds under a minimal prior mass condition.  However, our theory does not recover the result for standard TS, i.e., when $\alpha$ is set to 1, which is out of the scope of the current work. Extension to the case of $\alpha =1$ is possible along the lines of Section 3.2 of \cite{Yang2020} with a few additional assumptions on the identifiability of the model and the complexity of the parameter space. We provide guidelines to extend our analysis for $\alpha=1$ case in the~Appendix~\ref{app:StanTS}.

En route to deriving regret bounds, we also provide a different proof of $\alpha$-posterior concentration than the one derived by~\citet{bhattacharya2019bayesian}. This new proof technique enables us to derive posterior concentration results in expectation and also yields the minimax optimal convergence rate for finite-dimensional models, which are crucial for computing near-optimal regret bounds in expectation. Our general analysis leverages recent developments in the theory of posterior concentration~\citep{GGV,Kleijn2006,Ghosal_2007,bhattacharya2019bayesian,Yang2020,ZG,Alquier2020} and their finite sample Bernstein von-Mises type analysis~\citep{Spokoiny2012,Panov2015}
in the Bayesian statistics community. Nonetheless, our analysis also borrows essential ideas from the work on regret bounds for TS in~\cite{agrawal2013further}, which introduced a novel and simplified way to quantify the sufficiency of the (expected) number of times the best arm is sampled. A critical step in the proof technique developed in~\cite{agrawal2013further} requires the posterior probability of over-estimating the mean of the best arm (by sampling from the posterior distribution) to be bounded away from zero (in expectation). Establishing such lower bounds requires studying the second-order properties of the $\alpha$-posterior. While it is convenient to establish such lower bounds when the posterior is Gaussian as assumed in~\cite{agrawal2013further} or under some restrictive conditions as stated in~\cite{mazumdar20a}, it is challenging to establish it for more general posteriors. In our setting with more general conditions on the prior and the likelihood, we appropriately adapt the finite sample Bernstein-von Mises results developed in~\cite{Spokoiny2012,Panov2015} to compute a lower bound on such posterior probability.

Complementing the theory, our experiments (Appendices~\ref{app:More}--\ref{app:Fancy}) give an operational reading of the tempering parameter. First, $\alpha$ acts as an explicit \emph{prior-retention dial}: because online observations enter the $\alpha$-posterior with effective weight $\alpha$ against a full-weight prior. Smaller $\alpha$ retains prior information longer, beneficially when the prior is informative and at a controlled cost when it is misspecified. Second, $\alpha$ shapes the \emph{dispersion} of the regret distribution: whenever committing to a close suboptimal competitor is the dominant risk, we find that the upper quantiles of the regret are heaviest near $\alpha = 1$, where standard TS sits, and are lighter under moderate tempering, both with an informative prior and with a non-informative one. Third, $\alpha$-TS is empirically competitive with strong Bernoulli and categorical baselines, including KL-UCB, kl-UCB++, IMED, and $\varepsilon$-exploring TS, and runs unchanged under non-conjugate priors such as Gaussian-mixture and logit-normal priors.

The layout for the rest of the paper is as follows. In section~\ref{sec:Prob}, we will discuss the formulation of the multi-armed bandit problem and introduce $\alpha$-TS along with related works. Section~\ref{sec:RB} will present our main results that characterize the regret bound evaluating the performance of $\alpha$-TS, provided with a set of regularity conditions on the prior and reward distributions, along with a proof sketch. In section~\ref {sec:Exam}, we will demonstrate that two broad classes of distributions, sub-Gaussian and exponential families, satisfy our regularity conditions. The detailed proofs and experiments can be found in the appendix.

\section{Problem setup}~\label{sec:Prob}
We consider a stochastic MAB problem in which an agent is given a slot machine with $K$ arms.
At each time step $t\in\{1,2,\ldots\}$, the agent pulls an arm $k\in [K]$ and receives a random reward, where for any $n \in \bbN$, $[n]$ denotes the set $\{1,2,\ldots,n\}$. Let $X_{k,0}$ is the collection {of $n_{k,0}$ number} of warm-up samples from the $k^{th}$ arm and $X_{k,t}=\{x_{k,1},x_{k,2},\ldots,x_{k,n_k(t)} \} \bigcup X_{k,0} $ denote the rewards observed from the arm $k$ till time step $t$, where $n_{k}(t)$ is the number of times arm $k$ has been pulled up to and including time $t$, excluding the initial warm-up samples. 
\nomenclature[X]{$X_{k,t}$}{collection of reward observed from arm $k$ till time step $t$ including warmup samples}
\nomenclature[n]{$n_{k}(t)$}{number of times arm $k$ is pulled after warmup}
\nomenclature[k]{$K$}{Number of arms} 
\nomenclature[K]{$[K]$}{$\{1,2,3\ldots,K\}$}
\nomenclature[n]{$n_{k,0}$}{number of warm-up samples for arm $k$}
\nomenclature[x]{$X_{k,0}$}{collection of warm-up samples for arm $k$}

\hspace{-3em} Let $X_t:=\{X_{1,t},X_{2,t},\ldots,X_{K,t}\}$. \nomenclature[x]{$X_t$}{Collection of all the observations from each arm including warm-up samples} At each time step $t$, a stochastic MAB algorithm determines the arm $i_t \in [K]$  using the observations $X_{t-1}$. We formally define $n_k(t):= \sum_{s=1}^{t}\Ind(i_s = k)$, where $\Ind(\cdot)$ is an indicator function. For each arm $k\in [K]$, we assume that $x_k$ is a random variable defined on a probability space 
$(\Omega,\mathcal{S},P_{\theta_k})$, where $\Omega\subseteq \bbR$ and $\pmb \theta =\{\theta_1,\ldots,\theta_K\} \in\Theta^{K}$, where $\Theta \subseteq \bbR^p$ \nomenclature[t]{$\pmb \theta$ ($\pmb \theta_0$)}{Collection of parameters (true) for $K$ arms} is an open set of parameters equipped with the subspace topology inherited from $\mathbb{R}^p$. We assume that given $i_t=k$ (for any $k\in[K]$)\nomenclature[t]{$i_t$}{the arm number pulled at time $t$}, the reward generated from the $k^{th}$ arm is independent of any of the previous rewards. 
\nomenclature[r]{$\bbR$($\bbN$)}{Set of real (natural) numbers}
\nomenclature[t]{$\theta_i$}{a parameter for $i^{th}$ arm}
\nomenclature[t]{$\theta_{0,i}$}{true but unknown parameter for arm $i$}
\nomenclature[t]{$\Theta$}{parameter space}
\nomenclature[p]{$P_{\theta_{0,k}}$}{the reward distribution for arm $k$}
For any $\pmb \theta \in \Theta$, we denote $P_{\pmb \theta} = \prod_{k=1}^{K} P_{\theta_k}$. We assume that there exists (but unknown) a true parameter $\pmb \theta_0 = \{\theta_{0,1},\ldots,\theta_{0,K}\} \in \Theta^K $. For any $k\in [K]$, we denote $\mu_k$ and $\mu_{0,k}$ as the mean of the reward distributions $P_{\theta_k}$ and $P_{\theta_{0,k}}$ respectively. \nomenclature[m]{$\mu_k$($\mu_{0,k})$}{the mean of the reward distributions $P_{\theta_k}$($P_{\theta_{0,k}})$} We also assume that for each distribution $P_{\theta_k}$ there exists a density function $dP_{\theta_k}$ or $p(\cdot|\theta_k)$. 
We assume that all densities are well-defined with respect to the Lebesgue measure. For any $\alpha >0$, we denote the $\alpha-$R\'enyi divergence between a reward density $p(\cdot|\theta_k)$ and the true reward density $p(\cdot|\theta_{0,k})$ for the $i^{th}$ arm as
        \(D_{\alpha}(\theta_k,\theta_{0,k}) := \frac{1}{\alpha-1} \log \int p(x|\theta_k)^{\alpha} p(x|\theta_{0,k})^{1-\alpha} dx. \) 
\nomenclature[d]{$D_{\alpha}(\theta,\nu)$}{$\frac{1}{\alpha-1} \log \int p(x\given \theta)^{\alpha} p(x\given \nu)^{1-\alpha} dx$ for any $\alpha>0$}
\nomenclature[p]{$dP_{\cdot} \text{ or } p(\cdot\given \cdot)$}{density of $P_{\cdot}$} 
The (pseudo-)regret after $T$ time steps is defined as 
\begin{align}
    \text{Regret}(T) : =  \sum_{t=1}^{T} \left(\mu_{0,i^*} - \mu_{0,i_t} \right) ,
\end{align}  
where $i^* = \arg\max_{k\in[K]} \mu_{0,k} $. 

\nomenclature[i]{$i^*$}{arm with maximum true mean reward}
Typically, the objective of any stochastic MAB algorithm is to 
minimize $\bbE[\text{Regret}(T)]$, where expectation is taken with respect to the randomness in choosing $i_t$. Since $i_t$ is typically a mapping from the generated rewards (data) to an arm in $[K]$, this expectation is with respect to the true reward-generating distribution. 
In the next few lines, we describe the true reward-generating process in the stochastic MAB setting. 
Fix an algorithm that computes (or samples) $i_t \in [K]$ using $X_{t-1}$,  at any given $t\geq 1$. Let us assume that we first collect $n_{k,0}$ rewards from each arm and then start using  a MAB algorithm. 
So at time $t=1$, we have a set of $n_{k,0}$ rewards (denoted as $X_{k,0}$) sampled independently from $P_{\theta_{0,k}}$ for each $k\in [K]$. 
Thereafter, the algorithm uses this information $\{X_{k,0}\}_{k\in[K]}$ to compute (or sample) $i_1\in [K]$, samples a reward from $P_{\theta_{0,i_1}}$, and updates it to $X_1$. Similarly, a sequence of rewards is generated from $P_{\pmb \theta_0}$ using $X_t$ at each time step $t+1$ and then updating $X_{t+1}$. So, in essence, the true reward generation process depends on $P_{\pmb \theta_0}$ and the algorithm.

Following the standard in the bandit literature~\citep{lattimoreandszepesvari_2020}, throughout, we say that $f(T)=\mathcal{O}(g(T))$ or $g(T)=\Omega(f(T))$, if $ \left |\frac{f(T)}{g(T)}\right | < \text{C}$, where $C$ is some constant. Note that the original definition of Bachmann-Landau~\citep{bachmann1923analytische,landau1909handbuch} requires $ \limsup_{T\to\infty} \left |\frac{f(T)}{g(T)}\right | < \text{C}$. 

\subsection{$\alpha$-Thompson Sampling}~\label{sec:AlphaTS}

Thompson sampling (TS) was proposed by~\cite{THOMPSON1933} to solve a MAB problem, where the exploration/exploitation dilemma is addressed using a Bayesian heuristic. It is also referred to as \textit{probability matching} or \textit{posterior sampling} in some of the literature~\citep{russo2014learning}. Unlike the standard TS algorithm, we use an $\alpha$-posterior instead of the standard posterior distribution ($\alpha=1$), and call the resulting algorithm $\alpha$-TS. 
The pseudo-code of $\alpha$-TS is provided in Algorithm~\ref{alg:TS}.  In $\alpha$-TS, we first assume a prior distribution $\Pi_k$ on $\theta_k$ with density $\pi_k$. \nomenclature[p]{$\Pi_k(\cdot)$($\pi_k(\cdot)$)}{prior distribution (density)}\nomenclature[p]{$\Pi(\cdot)$($\pi(\cdot)$)}{joint prior distribution (or density).} For each $k\in [K]$, $t\geq 1$, and $\alpha\in(0,1)$, we define the $\alpha$-posterior  density as
\begin{align*}
    \pi_{\alpha,k}(\theta_k|X_{t}) 
   &= \frac{\pi_k(\theta_k) \prod_{u=1}^{n_{k,0}}p(x_{k,u}|\theta_k)^{\alpha}\prod_{s=1}^{t} \left[p(x_{k,
        s}|\theta_{k})q_s^k \right]^{\alpha\ind(i_s=k)}}{ \int \pi_k(\theta_k) \prod_{u=1}^{n_{k,0}}p(x_{k,u}|\theta_k)^{\alpha} \prod_{s=1}^{t} \left[p(x_{
            i_s}|\theta_{i_s})q_s^k\right]^{\alpha \ind(i_s=k)} d\theta_k }
        \\
           & = \frac{\pi_k(\theta_k) \prod_{u=1}^{n_{k,0}}p(x_{k,u}|\theta_k)^{\alpha}\prod_{s=1}^{n_k(t)} \left[p(x_{k,
        s}|\theta_{k}) \right]^{\alpha}}{ \int \pi_k(\theta_k) \prod_{u=1}^{n_{k,0}}p(x_{k,u}|\theta_k)^{\alpha}\prod_{s=1}^{n_k(t)} \left[p(x_{k,
        s}|\theta_{k}) \right]^{\alpha} d\theta_k },
\end{align*}
 where using the joint posterior distribution denoted as $\Pi_{\alpha}(\cdot|X_t)$, we define $q_s^k:=\Pi_\alpha(\mu_k = \max_j \mu_j |X_{s-1})$ is the probability of choosing arm $k\in [K]$ at time $s$ using $X_{s-1}$. \nomenclature[q]{$q_s^k$}{posterior probability of choosing arm $k\in [K]$ at time $s$} \nomenclature[p]{$\Pi_{\alpha,k}(\cdot\given X_t)$( or $\pi_{\alpha,k}(\cdot\given X_t)$)}{posterior distribution (or density) for arm $k$}%
 \nomenclature[p]{$\Pi_{\alpha}(\cdot\given X_t)$( or $\pi_{\alpha}(\cdot\given X_t)$)}{joint posterior distribution (or density).}
 The $K$ events $\{\mu_k = \max_{j} \mu_j\}$, $k\in[K]$, are almost surely disjoint under $\Pi_{\alpha}(\cdot\given X_{s-1})$, so that $\sum_{k\in [K]} q_s^k = 1$ and the categorical distribution below is well defined. Indeed, $\theta_1,\ldots,\theta_K$ are independent under the $\alpha$-posterior, since both the prior and the likelihood factorize across arms, and each $\theta_k$ admits a density because $\pi_k$ does (Assumption~\ref{Ass:Prior2}). As $\mu_k$ is a continuous and strictly increasing function of $\theta_k$ for the reward models considered here, each $\mu_k$ is atomless, whence $\Pi_{\alpha}(\mu_j = \mu_k\given X_{s-1}) = 0$ for every $j \neq k$ and ties carry no posterior mass. 
 
 In particular at each step $t\geq 1$, the $\alpha$-TS algorithm samples the optimal arm from the categorical distribution $\text{Cat}\left( \left\{q_t^k \right\}_{k\in [K]} \right) $.
 Observe that the definition of the $\alpha$-posterior distribution is unaffected by the probability of choosing the best arm as $k$ at each time step $t\geq 1$. 
 
\begin{algorithm}
     \caption{$\alpha$-Thompson Sampling}\label{alg:TS}
    \begin{algorithmic}[1]
    \Require A joint prior distribution $\pi$ on $\pmb \theta$ and $n_{k,0}$ 
    \State Set $X_{k,0}$ by sampling $n_{k,0}$ samples from each arm $k$, 
    \For{$t=1,2,\ldots T$}
        \State Compute $\alpha$-posterior $\pi_{\alpha}(\pmb \theta|X_{t-1})$,
        \State Sample an arm $i_t \sim  \text{Cat}\left( \left\{q_t^k \right\}_{k\in [K]} \right) $,
        \State Observe new reward $x_{i_t,n_{i_t}(t)}\sim P_{\theta_{0,i_t}}$,
        \State Update $X_{i_t,t}$ and $X_t$.
    \EndFor
    \end{algorithmic}
\end{algorithm}
It is evident from the algorithm that the true likelihood of generating data in~$\alpha$-TS is \begin{align}
    p_{\pmb \theta_0}^t(X_t) := \prod_{k=1}^{K} \prod_{u=1}^{n_{k,0}} p(x_{k,u}|\theta_{0,k})\prod_{s=1}^{t}\left[p(x_{
    k,s}|\theta_{0,k})q_s^k \right]^{ \ind(i_s=k)},
\end{align}
and we denote the corresponding data-generating distribution as $P_{\pmb \theta_0}^T$ till time $T$.\nomenclature[p]{$P_{\pmb \theta_0}^T$($p_{\pmb \theta_0}^T$)}{$\alpha$-TS data-generating distribution (density)}
Thus, unlike frequentist UCB algorithms, where $i_t$ is a mapping from $X_{t-1}\to [K]$, in $\alpha$-TS (or TS), we have a predictive distribution over $i_t$. 
Consequently, in our analysis, we establish bounds on the expectation (with respect to $P_{\pmb \theta_0}^T$) of the following conditional regret
\begin{align}
    \text{Regret}(T, \text{$\alpha$-TS} ) =  \sum_{t=1}^{T}  \bbE \left[\left(\mu_{0,i^*} - \mu_{0,i_t} \right) |X_{t-1}\right],
    \label{eq:TS-regret}
\end{align}  
where the expectation in the expression above is with respect to the predictive distribution of $i_t$.
For brevity, we denote the expectation with respect to the true data generating distribution as $\bbE[\cdot]$ and with respect to the predictive distribution of $i_t$ as $\bbE[\cdot|X_{t-1}]$.

Next, we describe the $\alpha$-TS algorithm  for a simple Bernoulli reward with Beta prior to draw parallels between TS and $\alpha$-TS. 
\begin{example}{Beta-Bernoulli MAB}

        In this example, we model the reward function as Bernoulli distribution, that is, for any $k\in [K]$, $x_k \sim Bernoulli(\theta_k)$, where $\theta_k\in [0,1]$ and $\theta_{0,k}$ is the true ( but unknown) parameter.
        We posit a $\text{Beta}(1,1)$ prior on $\theta_k$, for each arm. Let $S_k(t)$ and $F_k(t)$ denote the total number of successes and failures respectively for arm $k$ up to and including time $t$. Note that $S_k(t)+F_k(t)=n_k(t)$. Using these notations, the $\alpha$-posterior distribution for each arm $k$ can be expressed as 
        $\pi_{\alpha}(\theta_k|X_{t},\alpha-\text{TS}) \equiv \text{Beta}(S^{\alpha}_k(t),F^{\alpha}_k(t))$, where $S^{\alpha}_k(t)=1+\alpha S_k(t)$ and $F^{\alpha}_k(t)= 1+\alpha F_k(t)$. Note that in this example, $\mu_k=\theta_k$. For simplicity, let us assume $K=2$. Now the probability of $\{\theta_1-\theta_2>0\}$ is
\begin{align*}
    p^1_t= \pi_{\alpha} (\theta_1=\max (\theta_1,\theta_2) | X_{t}) &= \pi_{\alpha} (\theta_1 -\theta_2>0|X_{t})
    \\
    &= \frac{\int_{0}^{1} B(x|S^{\alpha}_2(t),F^{\alpha}_2(t)) \pi_{\alpha}(x|S^{\alpha}_1(t),F^{\alpha}_1(t)) dx }{B(S^{\alpha}_2(t),F^{\alpha}_2(t))},
    \end{align*}
        where $B(a,b)$ is the Beta function and $B(x|a,b)$ is the incomplete Beta function. So, in essence, $p_\alpha^1$ is a mapping from  $X_t$ to $[0,1]$ and $p^2_t=1-p^1_t$. Hence, $i_t$ has a predictive distribution  $\text{Bernoulli}(p_\alpha^1|X_t)$. 

        In Figure~\ref{fig:Bernoulli}, we plot the Regret$(T,\alpha\text{-TS})$ for various values of $\alpha$. As noted in the introduction, we observe that the performance of $\alpha$-TS algorithm for $\alpha$ values closer to 1 is similar to the performance of $\alpha=1$ (TS). 
\begin{figure}[ht]
\centering
    \includegraphics[width=\textwidth]{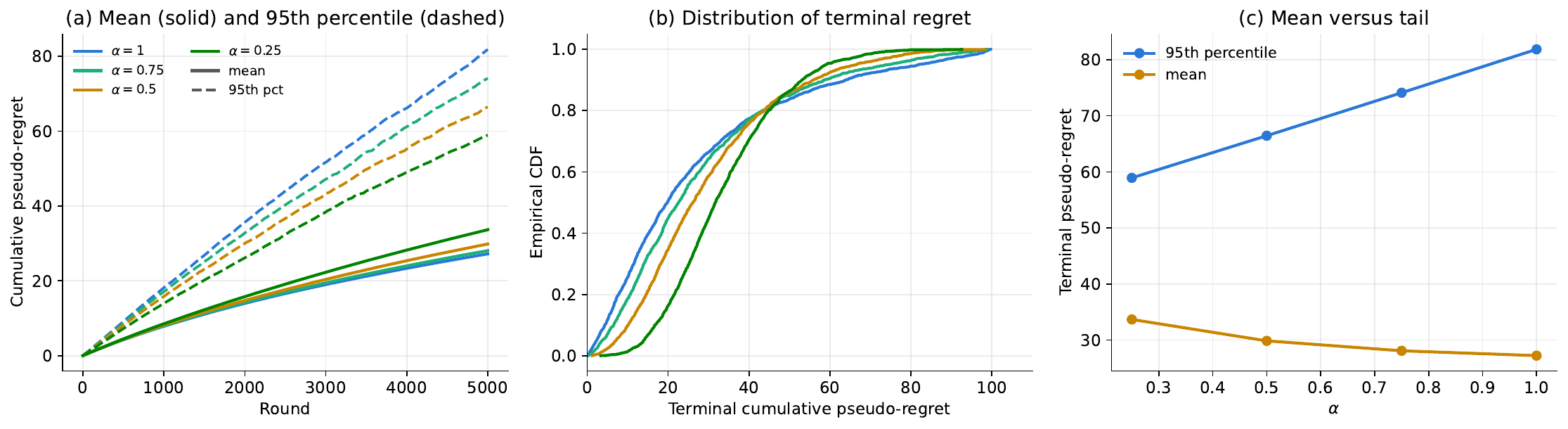}
    \caption{Beta--Bernoulli $\alpha$-TS on a two-armed bandit with true means $(0.60, 0.58)$ and a non-informative $\mathrm{Beta}(1,1)$ prior ($T = 5{,}000$, $2{,}000$ replications on shared reward streams); this is the instance used throughout Appendix~\ref{app:AlphaRole}. (a) Mean (solid) and $95$th percentile (dashed) of cumulative pseudo-regret against the round. The mean curves stay bundled together, rising from $27.2$ at $\alpha = 1$ to $33.7$ at $\alpha = 0.25$, whereas the $95$th-percentile curves separate progressively as the horizon grows, falling from $81.9$ to $58.9$. (b) Empirical CDF of terminal regret: the distributions cross near the $80$th percentile, so standard TS ($\alpha = 1$) is better on typical runs and worse on unlucky ones. (c) Mean and $95$th percentile of terminal regret against $\alpha$: as $\alpha$ decreases the mean rises by about $24\%$ while the $95$th percentile falls by about $28\%$.}
        \label{fig:Bernoulli}
\end{figure}

This example also exposes the operational meaning of $\alpha$. Since the $\alpha$-posterior is $\mathrm{Beta}(a_0 + \alpha S_k(t), b_0 + \alpha F_k(t))$, the prior pseudo-counts enter at full weight while every observation is discounted by $\alpha$: after $n$ pulls the data contribute an \emph{effective sample size} of $\alpha n$, so a prior of strength $n_0 = a_0 + b_0$ governs behavior for roughly $n_0/\alpha$ rounds. In this sense $\alpha$ is an explicit \emph{prior-retention dial}, and halving it doubles the horizon over which the prior dominates the posterior. Figure~\ref{fig:Bernoulli} illustrates the consequence on two closely separated arms under a non-informative prior, where there is no prior information to retain. Tempering raises mean regret moderately, yet it substantially lightens the upper tail of the regret distribution. Standard TS is better on typical runs and worse on unlucky ones, in which it commits early to the suboptimal arm and explores too little to recover. The two-armed instance isolates this mechanism, since confusing the two arms is the only way to incur regret. With many arms the same concentration occurs, in that tempering still reduces over-commitment to a close runner-up in the unlucky runs, but it is masked in the total regret by the additional exploration of clearly-suboptimal arms; Appendix~\ref{app:AlphaRole} reports this decomposition. The converse situation, in which the prior carries information and tempering improves the tail in absolute terms, is examined in 
Appendices~\ref{app:More}--\ref{app:Fancy}.
\end{example}
For computationally efficient sampling, one can also use other prior-likelihood combinations (other than Gaussian--Gaussian and Beta--Bernoulli) where $\alpha$-posterior has a closed form. For instance, Dirichlet--Multinomial/ Categorical, Gamma--Poisson/ exponential, Inverse Gamma--Gaussian (with known mean), Beta--Binomial/negative Binomial/ Geometric has closed form $\alpha$-posterior distributions and thus can be efficiently sampled from. Note that these are special examples of Diaconis--Ylvisaker conjugate priors in exponential families.
However, our theory applies to more general prior and reward combinations that are not necessarily conjugate, such as Gaussian mixture priors~\cite{hong22b}. In such cases, efficient posterior approximate sampling techniques can be used to sample from the intractable posterior distribution \textit{\`a la}~\cite{mazumdar20a} and~\cite{hong22b} to solve the MAB problem.
We also compare the empirical performance of $\alpha$-TS for Bernoulli reward with ~UCB~\citep{Auer2002}, UCBV~\citep{audibert2009exploration} and MOSS~\citep{audibert2009minimax} and provide the results in Appendix~\ref{app:More}. In addition, we benchmark $\alpha$-TS against KL-UCB~\citep{garivier2011kl}, KL-UCB++~\citep{menard2017minimax}, IMED~\citep{honda2015imed}, and $\varepsilon$-TS~\citep{jin2023eps} for Bernoulli and categorical rewards (Appendix~\ref{app:Baselines}); we study the empirical role of $\alpha$ and its effect on the tail of the regret distribution (Appendix~\ref{app:AlphaRole}); and we demonstrate $\alpha$-TS with structured and non-conjugate priors (a Gaussian-mixture prior and a logit-normal prior) for which posterior sampling remains simple while standard TS analyses do not apply (Appendix~\ref{app:Fancy}). 

\subsection{Related Works}
\phantom{a}
 Lai and Robbins, in their pioneering work \citep{lai1985asymptotically} establish a generic (for any algorithm) asymptotic lower bound for one-parameter reward models. They also introduce the popular UCB methodology to achieve the derived asymptotic lower bound for Laplace and Gaussian (with known variance) reward models.~\cite{burnetas1996optimal} generalizes the asymptotic lower bound to handle multi-parameter and nonparametric settings.~Assuming consistency of the algorithm, they show that
\begin{align}
   \liminf_{T\to \infty} \frac{\bbE[\text{Regret}(T)]}{\log (T)} \geq \sum_{k: \mu^0_k < \mu_{0,i^*}}  \frac{\Delta_k }{\inf_{\theta: \mu> \mu_{0,i^*}} \text{KL}(P_{\theta_k}^0\| P_{\theta} )},
   \label{eq:LB}
\end{align}
where $\mu$ is the mean under $P_{\theta}$, $\text{KL}(p\|q):=\int p(x) \log \frac{p(x)}{q(x)}dx $ is the Kullback-Liebler divergence between distributions $p$ and $q$, and $\Delta_k=\mu_{0,i^*} - \mu_{0,k}$. There are a number of works~both for UCB~\citep{Auer2002,honda2010asymptotically}  and TS~\citep{Kaufmann2012,agrawal2012analysis,honda2014optimality,Kaufmann2013} 
that establish asymptotically optimal regret bounds that are instance-dependent in the sense of~\eqref{eq:LB}.

Moreover, \cite{li2012open} posed establishing problem-independent (that is independent of $\mu_{0,i^*} - \mu_{0,k}$) regret bound for TS that is close to non-asymptotic lower bound $\Omega(\sqrt{KT})$ as an open problem. Subsequently,~\cite{agrawal2013further}, establish instance-independent regret bounds  of the TS algorithm for two popular MAB settings with Beta and Gaussian priors, which are near optimal. The result in~\cite{agrawal2013further} assumes apriori that the sampling distribution at each stage of their TS algorithm is Gaussian (or Beta) with mean and variance updated in a non-Bayesian fashion. This enables them to claim their regret bound to hold for any reward distribution with bounded support and where they use the well-defined Gaussian structure in their proof. Nonetheless, many ideas developed in~\cite{agrawal2013further} are general enough to include Bayesian posterior with Gaussian (or Bernoulli) reward.  However, to the best of our knowledge, no result establishes frequentist regret bound for the TS algorithm for general prior and reward distributions without leveraging conjugacy or specific distributional assumption on the sampling distribution. The closest are the works by~\cite{mazumdar20a} and~\cite{gopalan2013thompson}.

\cite{mazumdar20a} provide two efficient Langevin MCMC algorithms for TS with provable and (optimal) instance-dependent regret bound under restrictive structural conditions on the prior and the reward models. In ~\cite{gopalan2013thompson}, the authors assume that the prior and reward distributions are discretely supported and compute an asymptotically optimal instance-dependent regret bound. In addition, frequentist regret bounds with general prior and reward distributions for various complex extensions of MAB such as MAB with finite but dependent arms, linear Bandit with infinite arms and linear reward, and Gaussian process bandit, are still open. Another work by~\cite{honda2014optimality} establishes finite time regret bounds for TS with Gaussian rewards with unknown mean and variances and shows that their bounds are not asymptotically optimal in the sense of~\eqref{eq:LB} for some class of non-informative priors. Table~\ref{tab:my-table} summarizes the existing non-asymptotic bounds on the frequentist regret of TS. Interestingly, there is a recent work by~\cite{jin21d} where they propose a minimax optimal version of the Gaussian TS algorithm that clips the samples from the Gaussian posterior distribution; however, their approach requires the knowledge of time horizon in advance in addition to a closed form sampling distribution. Moreover,~\cite{jin2022finite,Tianyuan2023} combines the idea of $\e$-greedy exploration with TS to propose a novel algorithm that is both asymptotic and minimax optimal. There is a parallel line of work that considers Bayesian regret for TS and computes non-asymptotic bounds~\citep{russo2014learning}. Our bounds are not comparable to this work as it is computed for a different definition of the regret.  

Furthermore, there is a recent interest in studying limiting distributional properties of the random regret (and arm sampling distribution) instead of just computing a finite time bound on the expected regret (which summarizes all the information in the regret distribution to a single performance measure). In particular,~\cite{Glynn21,Wager21} show that $\text{Regret}(T,\text{TS})$ converge weakly to a diffusion limit, which is characterized as the solution to a stochastic differential equation. The authors establish these results in a specific diffusion-asymptotic regime where $\Delta_k$ depends on $T$ and is of the order  $\frac{1}{\sqrt{T}}$. However, the results in~\cite{Wager21} also assume that the prior variance scales with the time horizon and asymptotically becomes non-informative, which is uncommon in TS literature where fixed priors are used. \cite{Wager21} also note that the existing state-of-the-art instance-dependent regret bound derived in~\cite{agrawal2013further} diverges in the  diffusion-asymptotic regime. We also address this gap by deriving a new instance-dependent bound~(see Theorem~\ref{thm:PIRB0}), under our general setting.
Also,~\cite{Kalvit21} studies the behavior of sampling distribution of the arms for the Beta-Bernoulli TS, when both the arms have reward distribution Bernoulli($\theta$). They observe that the sampling distribution in this case either converges to the uniform distribution ($\text{Beta}(1,1)$) for $\theta=1$ or a Dirac measure at $1/2$ for $\theta=0$ and also provide theoretical justification for this phenomenon. We believe that the theoretical development in this paper can be used to generalize the result in~\cite{Kalvit21} for TS to general prior and reward distribution. 

\begin{table*}[t]
\centering
\resizebox{\textwidth}{!}{%
\begin{tabular}{@{}ccccc@{}}
\toprule
\multicolumn{3}{c}{\textbf{Assumptions}} &
  \multicolumn{2}{c}{\textbf{Regret bounds}} \\ \midrule
\textbf{Prior} &
  \textbf{Reward} &
  \textbf{Posterior} &
  \textbf{Instance-dependent} &
  \textbf{Instance-independent} \\ \midrule
Beta &
  \multirow{2}{*}{\begin{tabular}[c]{@{}c@{}}Any distribution \\ with bounded support\end{tabular}} &
  Beta &
  \begin{tabular}[c]{@{}c@{}}$(1+\e) \left( \sum_{k\neq i^*} \frac{\Delta_k }{\text{KL}(\theta_{0,k},\theta_{i^*}^0)}\right) \log(T) +\mathcal{O}(\frac{K}{\e^2})$, \\ for any $\e>0$ \citep{agrawal2013further,agrawal2012analysis}\end{tabular} &
  \begin{tabular}[c]{@{}c@{}}$\mathcal{O}(\sqrt{KT \log(T)})$\\ \citep{agrawal2013further,agrawal2012analysis}\end{tabular} \\ \cmidrule(r){1-1} \cmidrule(l){3-5} 
Gaussian &
   &
  Gaussian &
  -- &
  \begin{tabular}[c]{@{}c@{}}$\mathcal{O}(\sqrt{KT \log(K)})$\\ \citep{agrawal2013further}\end{tabular} \\ \midrule
Jeffrey's &
  Exponential Family &
  \multirow{2}{*}{\begin{tabular}[c]{@{}c@{}}Standard \\ posterior\end{tabular}} &
  \begin{tabular}[c]{@{}c@{}}$\frac{1+\e}{1-\e}\left( \sum_{k\neq i^*} \frac{\Delta }{\text{KL}(\theta_{0,k},\theta_{i^*}^0)}\right) \log(T) + C(\e,B) $,\\ for any $\e\in(0,1)$ and a instance-dependent \\ constant $B$. \citep{Kaufmann2013}\end{tabular} &
  -- \\ \cmidrule(r){1-2} \cmidrule(l){4-5} 
Discrete support &
  \begin{tabular}[c]{@{}c@{}}Discrete support \\ (with dependent arms)\end{tabular} &
   &
  \begin{tabular}[c]{@{}c@{}}$B + C(\log T)$, where $B$ is a problem-\\ dependent constant and C is the solution\\ of an optimization  problem.\citep{gopalan2013thompson}\end{tabular} &
  -- \\ \midrule
\begin{tabular}[c]{@{}c@{}}Log-concave \\ and Lipschitz smooth \end{tabular} &  
  \begin{tabular}[c]{@{}c@{}}Log-concave \\ and Lipschitz smooth \end{tabular} & \begin{tabular}[c]{@{}c@{}}Standard \\ posterior \end{tabular}
   &
  \begin{tabular}[c]{@{}c@{}} $\mathcal{O}\left(\sum_{k\neq i^*} \left[\frac{\log(T)}{\Delta_k} + \Delta_k\right]\right)$ \citep{mazumdar20a}\end{tabular} &
  -- \\ \midrule
\textbf{General prior} 
&
  \textbf{\begin{tabular}[c]{@{}c@{}}Any reward with sub-Gaussian error\\ or exponential-family\\ with bounded variance\end{tabular}} &
  \textbf{\begin{tabular}[c]{@{}c@{}}$\alpha$- \\ posterior\end{tabular}} &
  \textbf{\begin{tabular}[c]{@{}c@{}}
  $\sum_{k \neq i^*} \Delta_k\left(9(r_0+1) \frac{\log(T)}{C(\alpha)\Delta_k^2}  + \frac{(2r_0+3)}{2} \right)$
  (Theorem~\ref{thm:RB-PD})\\ 
  $\sum_{k\neq i^*, A(K,T,\alpha)} \left(\frac{9(r_0+1) \max\{\log(T C(\alpha)\Delta_k^2/9),\,1\}}{C(\alpha)\Delta_k} + r_0 \Delta_k + \frac{27}{2 C(\alpha) \Delta_k}  \right)$
  \\
           $ + \sum_{k\neq i^*, A(K,T,\alpha)^\mathtt{c}} \Delta_k T$,
  (Theorem~\ref{thm:PIRB0}) 
  \end{tabular}} 
  &
  \textbf{\begin{tabular}[c]{@{}c@{}}$\mathcal{O}(\sqrt{KT \log(K)})$\\ (Theorem~\ref{thm:PIRB})\end{tabular}} 
  \\ \bottomrule
\end{tabular}%
}
\caption{Existing results on the finite sample frequentist regret bounds for TS for stochastic MAB. $r_0$ is a constant that depends on the prior and $\alpha$. ( Note that, we derive two instance-dependent bound, the second one is similar to the instance-dependent regret bound of improved UCB~\cite{Auer2010} and also have appropriate limiting behaviour under diffusion asymptotic regime~\cite[ Sec. 4(pg. 24)]{Wager21})}
\label{tab:my-table}
\end{table*}

\section{Regret Bounds}~\label{sec:RB}
   In this section, we identify regularity conditions on the prior and reward distributions to compute a non-asymptotic bound on the $\alpha$-TS regret as defined in~\eqref{eq:TS-regret}. We carefully adapt and modify the first and second-order finite sample analysis of $\alpha$-posterior to compute the required regret bounds. 
    To study first-order properties of $\alpha$-posterior, we essentially use a variant of prior-mass condition (Assumption~\ref{ass:prior2}) specified in{~\cite[(2.9)]{GGV}} and derive an $\alpha$-posterior concentration bound in expectation instead of high-probability . Assumption~\ref{ass:prior2} in this work is stronger than the prior-mass assumption required in ~\cite{bhattacharya2019bayesian}, but yields the correct $\alpha$-posterior concentration rate of order $\sqrt{\nicefrac{1}{n_k(t)}}$ instead of $\sqrt{\nicefrac{\log n_{k}(t)}{n_{k}(t)}} $ for finite-dimensional reward models. We will later see that establishing $\sqrt{\nicefrac{1}{n_{k}(t)}}$ is crucial for the regret analysis as it enables us to appropriately control the initial number of samples $n_{k,0}$. 
Then, we combine our $\alpha$-posterior concentration bounds to derive regret bounds for $\alpha$-TS, by generalizing the arguments in~\cite{agrawal2013further}.

As we note at the end of introduction, a critical step is to compute the posterior probability of over-estimating the mean of the best arm to be bounded away from zero (in expectation). 
We briefly discuss the relevant results from~\cite{Spokoiny2012,Panov2015} in Section~\ref{sec:FBvM} and Lemma~\ref{lem:FBVM} that we use to compute a lower bound on such posterior probability.

        We first discuss regularity conditions on the reward and prior distributions.
\nomenclature[c]{$C(\alpha)$}{concentration constant $D(1-\alpha)\min(2\alpha,\,1-\alpha)/16$ appearing in the regret bounds}%
\nomenclature[d]{$\Delta_k$}{suboptimality gap $\mu_{0,i^*}-\mu_{0,k}$ of arm $k$}%
\nomenclature[d]{$\delta_k$}{analysis radius $\Delta_k/3$}%
\nomenclature[d]{$D_2(\cdot \Vert \cdot)$}{R\'enyi divergence of order $2$}%
\nomenclature[n]{$\bar n_k(t)$}{$n_k(t)+n_{k,0}$, pulls of arm $k$ up to time $t$ including the warm-up samples}%
\nomenclature[m]{$\mu_k^t$}{mean corresponding to one draw from the $\alpha$-posterior of arm $k$ at time $t$}%
\nomenclature[l]{$L_k(T)$}{analysis threshold $\log(MT)/(C(\alpha)\delta_k^2)$}%
\nomenclature[a]{$A(K,T,\alpha)$}{set of arms with $\Delta_k > e\sqrt{K\log K}/\sqrt{TC(\alpha)}$}%
\nomenclature[t]{$\tau_u$}{time step at which the optimal arm is pulled for the $(u+n_{i^*,0})^{th}$ time}%
\nomenclature[k]{$\kappa$}{prior-thickness constant in Assumption~(B1)}%
\nomenclature[k]{$\mathrm{KL}(\cdot \Vert \cdot)$}{Kullback--Leibler divergence}%
\nomenclature[r]{$r_0$}{constant from Lemma~\ref{lem:Int2} bounding the odds of under-estimating the optimal arm}%
\nomenclature[m]{$M$}{free parameter of Lemmas~\ref{lem:Int4} and~\ref{lem:Int2}, required to satisfy $Mn_{k,0} \geq e$}%
\nomenclature[f]{$\mathcal{F}_s$}{natural filtration $\sigma\left(X_s, \{i_r\}_{r=1}^s\right)$ of the algorithm}%
\nomenclature[u]{$u_k$, $y_k$}{analysis-time thresholds $\mu_{0,k}+\delta_k$ and $\mu_{0,i^*}-\delta_k$}%

        \begin{assumption}~\label{ass:reward}
            We assume that there is a unique optimal arm $i^*\in[K]$. 
        \end{assumption}

        The assumption above is standard in the TS literature. 
        It implies that for each $k\in [K]\setminus \{i^*\}$, there exist $u_k$ and $y_k$ such that $\mu_{0,k} < u_k < y_k < \mu_{0,i^*}$. To be precise, we define $\delta_k=\Delta_k/3$ and $y_k:=\mu_{0,i^*} -\delta_k$ and $u_k:= \mu_{0,k} +\delta_k $, that we use later in our proofs. Moreover,~\cite{agrawal2012analysis} show that if there are more than one optimal arms then the expected regret will further decrease. The argument in \cite[Appendix A]{agrawal2012analysis} is algorithm-agnostic and relies only on the definition of regret. Consequently, it extends directly to $\alpha$-TS as well as to other multi-armed bandit algorithms. Therefore, we can assume uniqueness of the best arm without loss of generality. 
        
      We next specify the assumptions on the reward distributions. 
      In addition to the assumptions on the reward distributions listed in Section~\ref{sec:FBvM},  we also require that the squared difference in respective means (evaluated at two of its parameters) by the corresponding $\alpha$-R\'enyi divergence between them. 
      In particular, we assume the following
      \begin{assumption}[Renyi divergence]~\label{ass:Renyi}
      For any $\alpha\in(0,1)$ and $\theta,\vartheta \in \Theta$, there exists a constant $D>0$, such that
     \(   |\bbE_{\theta}[r]- \bbE_{\vartheta}[r]| \leq D\sqrt{\frac{2}{\alpha} D_{\alpha}(\theta,\vartheta)},\) where $\bbE_{\theta}[r]$ is the mean reward under reward distribution $p(\cdot|\theta)$. 
\end{assumption}
    Assumption~\ref{ass:Renyi} is needed to adapt the existing posterior concentration analysis in the Bayesian literature for the regret analysis. It connects model discrepancy to the difference in mean. In posterior concentration analysis, typically, the model discrepancy is measured using the divergence measure of a reward distribution from the true reward distribution, whereas regret is defined using mean rewards at respective actions. This connection is reflected in Assumption~\ref{ass:Renyi} which requires the squared difference in mean rewards to be bounded by the R\'enyi divergence between respective models. In Lemmas~\ref{lem:Pin} and~\ref{lem:Exp}, we show that Assumption~\ref{ass:Renyi} holds for sub-Gaussian (Definition~\ref{def:SubReward}) and exponential families (Definition~\ref{def:rewardExp}), respectively. While Lemma~\ref{lem:Exp} follows straightforwardly from the definition of the exponential family, Lemma~\ref{lem:Pin} is novel to the best of our knowledge. Lemma~\ref{lem:Pin} generalizes a similar classical result on \textit{transportation cost inequality}~\cite[Lemma 4.18]{Boucheron2013} for KL divergence to R\'enyi divergence.

        Furthermore, in our next assumption, we impose a joint regularity condition on the prior and the reward distribution for each arm, that controls the minimum prior mass provided to the neighborhood of the true model. We use this assumption on the prior thickness to compute the rate of convergence of the $\alpha$-posterior distribution.

\begin{assumption}[Prior thickness]~\label{ass:prior2}
            We assume that, for each $k\in[K]$, $\alpha\in(0,1)$, $j\geq 1$, and $\delta_k^2=\Delta_k^2/9$, for  a ball $B\left(\theta_{0,k}, C(\alpha)\delta_k^2 \right) := \left\{  \theta_k \in \Theta:D_2(\theta_{0,k},\theta_k)\leq  C(\alpha)\delta_k^2 \right\} $ and shell $F_k^j:= \big\{\theta_k \in \Theta: D\frac{\alpha}{2} (j+1) \delta_k^2 \geq D_{\alpha}(\theta_k,\theta_{0,k}) \geq D j \frac{\alpha}{2} \delta_k^2  \big\}$, there exist $\kappa > 0$, such that 
        \begin{align}\tag{B1} \frac{\Pi_k(F_k^j)}{\Pi_k(B\left(\theta_{0,k},C(\alpha)\delta_k^2 \right) )} \leq \kappa(j+1),
            \end{align}
        where $\kappa$ (a constant) depends on the prior, reward distributions and true model parameter.
        \end{assumption}

We show the above assumption holds for the sub-Gaussian and exponential family in Appendix~\ref{App:Assumption34} with any bounded prior distribution with continuous density. In general, the  Assumption~\ref{ass:prior2}  holds for any reward model for which $D _{2}(\theta_k, \theta_{0,k}) \leq  c_1 |\theta-\theta_0|^2$  and $D _{\alpha}(\theta_k, \theta_{0,k}) \geq c_2 |\theta-\theta_0|^2$ (locally) with bounded and continuous prior.~This is demonstrated in the proofs of Proposition~\ref{prop:ExpG} and Proposition~\ref{prop:SG} in the Appendix~\ref{App:Assumption34}.

Assumption~\ref{ass:prior2} is different than the common assumption in the Bayesian literature used for computing the convergence rate of nonparametric standard~\citep{GGV,ZG} as well as $\alpha$-posterior distribution~\citep{bhattacharya2019bayesian}. ~However, it is similar to the condition assumed in~\citet[(2.9)]{GGV} 
and enables us to compute the $\alpha$-posterior concentration rate that satisfies $\delta_k^2 n_{k}(t) \geq 1$. In fact, in Lemma~\ref{lem:Post} we will observe that $\nicefrac{1}{\sqrt{n_{k}(t)}}$, determines the concentration rate of the $\alpha$-posterior distribution.  The prior thickness assumption used in~\citet{bhattacharya2019bayesian}, which is adapted from~\citet[2.3]{GGV} is coarser than Assumption~\ref{ass:prior2} and is good for nonparametric and infinite dimensional models but would yield  $\sqrt{\nicefrac{(\log n_{k}(t))}{n_{k}(t)}}$ concentration rate if we use it for finite-dimensional models.~\citet{GGV} noted this issue and proposed a finer assumption (2.9) in their paper, yielding the minimax optimal rate for finite-dimensional models. Moreover, in our setting, using the common assumption as used in~\cite{bhattacharya2019bayesian}  would yield an initial number of samples $n_{k,0}$ of order $\nicefrac{e^{1/\delta_k^2}}{\delta_k^2}$\footnote{The typically condition on the prior thickness in \citet{bhattacharya2019bayesian} requires $\pi_k(B\left(\theta_{0,k}, \delta_k^2 \right) ) \geq  e^{- n_{k}(t) \delta_k^2 }$ for any $\delta_k^2 \geq \frac{\log(\delta_k^2 n_{k}(t))}{n_{k}(t)}$. To show this condition, the first step is to show that $\pi_k(B\left(\theta_{0,k}, \frac{\log(\delta_k^2 n_{k}(t))}{n_{k}(t)} \right) )\geq \sqrt{\frac{\log(\delta_k^2 n_{k}(t))}{n_{k}(t)}}$ (upto constants). Now it remains to argue that $ \sqrt{\frac{\log(\delta_k^2 n_{k}(t))}{n_{k}(t)}} \geq e^{-n_{k}(t)\delta_k^2}$, which would require $n_{k,0}$ to be of order  $\nicefrac{e^{1/\delta_k^2}}{\delta_k^2}$.} which is undesirable and yields a trivial regret bound. With this finer assumption, we significantly adapt the proof of~$\alpha$-posterior concentration in~\cite{bhattacharya2019bayesian}.  

Recall, from the discussion at the beginning of this section, that we need finite sample Bernstein-von Mises (BvM) to derive a lower bound on the posterior probability of overestimating the mean of the best arm. Therefore, in the next section, we lay down the assumptions required for the second-order analysis of the $\alpha$-posterior distribution, in particular, establishing the finite sample BvM theorem for it. 

\subsection{Finite Sample Bernstein-von Mises}~\label{sec:FBvM}
In this section, we specify the regularity conditions developed in~\cite{Spokoiny2012,Panov2015} for finite sample  BvM theorem to hold for parametric models. 
 First, we specify some of the simplifying notations (by omitting indexes for arms) for this section. Let \( X_n = (x_{1},\ldots,x_{n})^\top \) be an i.i.d. sample from a measure \( P_{\theta_0} \equiv P_0\). We assume that the data generating distribution belongs to a parametric family 
\(
\mathcal{P} = \{ P_\theta : \theta \in \Theta \subseteq \mathbb{R}^p \},
\)
where each $P_\theta$ is a probability measure indexed by the parameter $\theta$. That is, there exists a parameter space $\Theta \subseteq \mathbb{R}^p$ such that the true distribution $P_{\theta_0} \in \mathcal{P}$ for some $\theta_0 \in \Theta$.
 
 Denote the log-likelihood of generating $X_n$ as $\sL(\theta) = \log dP_{\theta}(X_n)  = \sum_{i=1}^{n} \ell(X_i,\theta)$ , where $\ell(X_i,\theta)=\log dP_{\theta}(X_i)$. $\nabla \sL(\theta)$ denotes gradient of $\sL(\cdot)$ evaluated at $\theta$ and $\nabla^2\bbE\sL(\theta)$ stands for the Hessian of the expected log likelihood.  Define 
 \begin{align}
    \label{eq:fisherFullparam}
    \sD_0^{2}
     := &
    - \nabla^{2} \bbE \sL(\theta_0) .
 \end{align}
Note that $\sD_0^{2} = nI(\theta_0)$ , where $I(\theta_0)$ is the Fisher information of $P_{\theta}$ at $\theta_0$. For notational convenience, we also write $\sV_0^2 := I(\theta_0)$, so that $\sD_0^{2} = n \sV_0^2$.
Also denote the maximum likelihood estimator as $\hat{\theta}_n := \arg\max_{\theta \in \Theta} \sL(\theta)$ and $\theta_0 = \arg\max_{\theta \in \Theta} \bbE[\sL(\theta)]$. The stochastic part of the log-likelihood is denoted as $\zeta_i (\theta):= \ell(X_i,\theta) - \bbE[\ell(X_i,\theta)]$.
\nomenclature[l]{$\sL(\theta)$}{Log-likelihood function $\log dP_\theta(X_n)$}
\nomenclature[l]{$\ell(X_i,\theta)$}{Log-density $\log dP_\theta(X_i)$}
\nomenclature[n]{$\nabla \sL(\theta)$}{Gradient of the log-likelihood at $\theta$}
\nomenclature[d]{$\sD_0^{2}$}{Negative Hessian of expected log-likelihood at $\theta_0$; $\sD_0^{2} := -\nabla^2 \bbE\sL(\theta_0)$}
\nomenclature[i]{$I(\theta_0)$}{Fisher information of $P_\theta$ evaluated at $\theta_0$}
\nomenclature[v]{$\sV_0^2$}{Shorthand for $I(\theta_0)$ so that $\sD_0^2 = n\sV_0^2$}
\nomenclature[t]{$\hat{\theta}_n$}{Maximum likelihood estimator (MLE)}
\nomenclature[z]{$\zeta_i(\theta)$}{Stochastic log-likelihood fluctuation $\ell(X_i,\theta)-\bbE[\ell(X_i,\theta)]$}

 The regularity conditions required for the finite sample BvM to hold are split into local and global. The local conditions only describe the properties of the process \(\ell(X_1,\theta)\)
  for \(\theta \in \Theta_0(\mathtt{u}_0) \) with some fixed value \(\mathtt{u}_0\), where
  \begin{align}
    \Theta_0(\mathtt{u}_0)
    :=
    \big\{\theta \in \Theta \colon \|\sV_0(\theta - \theta_0)\| \le  \mathtt{u}_0 \big\}.
  \end{align}
The global conditions have to be fulfilled on the whole \(\Theta\).
\begin{assumption}~\label{ass:Spokoiny}
We start with some exponential moments conditions. 
  \begin{enumerate}
    \item[\({(e\!d_{0})} \)]
      There exists constants \(\nu>0\)
       and \( \mathtt{g} > 0 \) such that
      \begin{align}
        \sup_{\gamma \in \bbR^{p}\setminus \{0\}} \log\bbE \exp\left\{
              \mathtt{m} \frac{\langle \nabla \zeta_1(\theta_0),\gamma \rangle}
              {\| \sV_0 \gamma \|}
              \right\}
        \le
        \frac{\nu^{2} \mathtt{m}^{2}}{2}, ~ |\mathtt{m}| \le \mathtt{g}.
      \end{align}

    \item[\( {(e\!d_{1})} \)]
      There are constants \( \omega > 0 \) and $\mathtt{g}_1>0$ such that for each $\mathtt{u} \leq \mathtt{u}_0$ and $\mathtt{m} \le \mathtt{g}_1$:
      \begin{align}
        \sup_{\gamma \in \bbR^{p}\setminus \{0\}} \sup_{\theta \in \Theta_0(\mathtt{u})} \log \bbE \exp\left\{
        \frac{\mathtt{m}}{\omega} \frac{\gamma^{\top}[\nabla \zeta_1(\theta) - \nabla \zeta_1(\theta_0)  ] }{ \mathtt{u} \|\sV_0 \gamma\|}
        \right\}
        \le
        \frac{\nu^{2} \mathtt{m}^{2}}{2}, ~ |\mathtt{m}| \le \mathtt{g}_1.
      \end{align}
  \end{enumerate}
\end{assumption}

  \begin{enumerate}
    \item[\({(\ell_0)} \)]
      There exists a constant \(\delta^*\) such that it holds for all \(\mathtt{u} \le \mathtt{u}_0\)
      \begin{align}
       \sup_{\theta \in \Theta_0(\mathtt{u})} \bigg|\frac{2\text{KL}(\theta_0,\theta)}{(\theta -\theta_0)^{\top}I(\theta_0)(\theta -\theta_0)}-1 \bigg|
        \le
        \delta^* \mathtt{u},
      \end{align}
      where $\text{KL}(\theta_0,\theta)$ is the KL divergence between model $P_{\theta_0}$ and $P_{\theta}$.
  \end{enumerate}

  The following two are the required global conditions.
  \begin{enumerate}
    \item[\( {(\ell{\mathtt{u}})} \)]
      For any \(\mathtt{u}>0\) there exists a value \(\mathtt{b}(\mathtt{u}) > 0\),
      such that  
      \begin{align}
            \sup_{\theta \in \Theta: \|\sV_0(\theta - \theta_0)\|= u} \frac{\text{KL}(\theta,\theta_0)}{\|\sV_0 (\theta - \theta_0)\|^2} \ge \mathtt{b}(\mathtt{u}).
      \end{align}

      \item[\( {(e\mathtt{u})} \)]
       For each $\mathtt{u} \geq 0$, there exists $\mathtt{g}_1(\mathtt{u}) >0 $, such that for all $|\mathtt{m}| \le \mathtt{g}_1(\mathtt{u})$:
      \begin{align}
        \sup_{\gamma \in \bbR^{p}\setminus \{0\}} \sup_{\theta \in \Theta_0(\mathtt{u})} \log \bbE \exp\left\{
        \mathtt{m} \frac{ \gamma^{\top}\nabla \zeta_1(\theta)   }{ \|\sV_0 \gamma\|}
        \right\}
        \le
        \frac{\nu^{2} \mathtt{m}^{2}}{2}.
      \end{align}
  \end{enumerate}

The discussion regarding these assumptions can be found in Section 5.1 of~\cite{Spokoiny2012} and partly in~\cite{Panov2015}. Recall from Assumption~\ref{ass:prior2} that (effectively) no assumptions are imposed on the reward distribution in order to derive first-order concentration properties of the $\alpha$-posterior distribution. 
 However, when aiming to derive a more-refined second-order BvM type result, a more comprehensive set of local and global moment conditions is required, as specified in Assumption~\ref{ass:Spokoiny}. These conditions impose additional regularity requirements on the structural properties of the reward distributions. In Appendix~\ref{app:Spokoiny}, we show that the exponential family (Definition~\ref{def:rewardExp}) and sub-Gaussian  (Definition~\ref{def:SubReward} with some mild regularity condition on the density of the error distribution) families satisfy all the conditions in Assumptions~\ref{ass:Spokoiny}. More examples can be found in~\cite{Spokoiny2012,Panov2015}.
 
 To extend the lower bound computed in~\cite[Theorem 4]{Panov2015} to general priors, we need to control the prior density on the set $\Theta_0(\mathtt{u}_0)$. In particular, observe that
  \begin{align}
    \pi_{\alpha}(\Theta_0(\mathtt{u}_0)|X_n)  
    =&
    \frac{\int_{\Theta} \exp \big\{ {\alpha} \sL(\theta, \theta_0) \big\} \pi(\theta)
            \Ind\big\{\theta \in \Theta_0(\mathtt{u}_0)\big\} d \theta}
         {\int_{\Theta} \exp \big\{ {\alpha} \sL(\theta, \theta_0) \big\}
            \big\} \pi (\theta) d \theta}.
  \label{eq:concentrBvmTarget}
  \end{align}

Note, that $\Theta_0(\mathtt{u}_0)= \big\{\theta \in \Theta \colon \|\sD_0(\theta - \theta_0)\| \le \sqrt{n} \mathtt{u}_0 \big\}= \big\{\theta \in \Theta \colon \|\sV_0(\theta - \theta_0)\| \le \mathtt{u}_0 \big\} $. So for a fixed $\mathtt{u}_0$, the prior density can easily be lower bounded on $\Theta_0(\mathtt{u}_0)$ by fixed number that depends on $\mathtt{u}_0$ and $\theta_0$. In fact, if $\mathtt{u}_0 =  \frac{c}{\sqrt{n}}$, for some constant $c>0$, then note that  $\min_{\theta\in \Theta_0(\mathtt{u}_0) } \pi(\theta)\geq \min_{\theta\in \Theta_0(c) } \pi(\theta)$, because, $\Theta_0(\mathtt{u}_0) \subseteq \Theta_0(c)$ for all $n\geq 1$. Therefore,

\begin{assumption}[Prior]
\label{Ass:Prior2}
We make the following assumption on the prior distribution.
    \begin{enumerate}
        \item We assume the prior density $\pi$ is continuous and strictly positive on $\Theta.$
        \item There exists a positive constant $\mathtt{M}$, such that the prior density, $\pi(\theta)\leq \mathtt{M}$ for all $\theta \in \Theta$.
       
    \end{enumerate}
\end{assumption}

Since, we are only interested in the lower bound result of~\cite{Panov2015}, we can see that, with the assumption above the result in Theorem~4 of~\cite{Panov2015}, follows easily with a factor of $\frac{\min_{\theta\in \Theta_0(c) } \pi(\theta)}{\mathtt{M}} := \mathbb{C}$. For parametric problems of interest in this work, the two conditions (Assumptions~\ref{ass:prior2} and~\ref{Ass:Prior2}) imposed on the prior distribution are very weak and just requires the prior density to be positive, continuous, and bounded.

Now we present below the relevant part of the~\cite[Theorem 4]{Panov2015} used in this paper.

\begin{lemma}~\label{lem:FBVM}
    Under Assumptions~$(e\!d_0),(e\!d_1),(\ell_0),(\ell\mathtt{u}), (e\mathtt{u})$, on the likelihood model, and Assumption~\ref{Ass:Prior2} on the prior density, for any measurable set $A\subseteq \bbR$ and $\alpha\in(0,1)$,
    \begin{align}
        \pi_{\alpha}(\sD_0(\theta- \theta_n^*)\in A|X_n) \geq \mathbb{C}\exp\left(-2\sqrt{\frac{p+ \eta}{n}} - 8e^{-\eta}\right) \left\{P(\phi/\sqrt{\alpha} \in A)  \right\} -e^{-\eta}, 
    \end{align}
    with $P_0$ probability of at least $1-e^{-\eta}$, where $\phi$ is a standard Gaussian random variable and $\theta_n^* = \theta_0 + \sD_0^{-2} \nabla\sL(\theta_0)$. Moreover, $\|\sD_0(\theta_0-\theta_n^*)\|\leq {C\sqrt{p+\eta}}$ with $P_0$-probability of at least $1-e^{-\eta}$ for some positive constant $C$.
\end{lemma}
The proof of the result above is a direct consequence of~\cite[Theorem 4]{Panov2015} and the result in display (33) of~\cite[Theorem 9]{Panov2015}.

\subsection{Main Results}~\label{sec:Main Results}
        Under the above assumptions, we first establish the following instance-dependent regret bound for $\alpha$-TS that matches to that of the improved UCB~\citep{Auer2010}.
        
        \begin{theorem}~\label{thm:PIRB0}
            For any $k\in[K]$, fix $\delta_k=\Delta_k/3$,  and $\alpha\in(0,1)$.   Then, under Assumptions~\ref{ass:reward},~\ref{ass:Renyi},~\ref{ass:prior2},~\ref{ass:Spokoiny}, and~\ref{Ass:Prior2},  for a warm-up size $n_{k,0} = n_0$, where $n_0 \geq 1$ is any constant independent of $T$ and of the gaps, $\kappa \leq (1-e^{-4})^2/48$ and $C < \sqrt{2}$ (where $C$ is the constant of Lemma~\ref{lem:FBVM}), we have
        \begin{align}
        \nonumber
            \bbE  \left[ \sum_{t=1}^{T} (\mu_{0,i^*} -  \mu_{0,i_t})  \right] 
            \leq&  \sum_{k\neq i^*, A(K,T,\alpha)} \left(\frac{9(r_0+1) \max\{\log(\nicefrac{T C(\alpha)\Delta_k^2}{9}),\,1\} }{C(\alpha)\Delta_k} + r_0 \Delta_k + \frac{27}{2 C(\alpha) \Delta_k}  \right)
            \nonumber
            \\
            & + \sum_{k\neq i^*, A(K,T,\alpha)^\mathtt{c}} \Delta_k T,
            \label{eq:TightPDUB}
        \end{align}
        where $A(K,T,\alpha)=\{k\in[K]:\Delta_k> \nicefrac{e\sqrt{K\log K}}{\sqrt{TC(\alpha)}}\}$ and $C(\alpha)=\nicefrac{D (1-\alpha) \min \left(2\alpha,1-\alpha  \right)}{16}$ and $r_0$ is a known constant introduced in Lemma~\ref{lem:Int2}.  .
        \end{theorem}
         The above bound has $\log(\text{Constant}\times T \Delta_k^2)$ term in the numerator instead of just $\log(T)$, which guarantees a better control on the expected regret with respect to $T$ for smaller $\Delta_k$. Moreover, for the two arm case considered in~\cite{Wager21} under diffusion asymptotic regime, where $\Delta = \frac{\bar \Delta}{\sqrt{T}}$, ( $\bar \Delta$ is the difference between the limiting mean reward of the two arms) the above upper bound reduces to  $\sqrt{T}\left(\frac{9(r_0+1) \log( C(\alpha)\bar \Delta^2/9)}{C(\alpha)\bar \Delta} + \frac{\bar \Delta}{T} + \frac{27}{2 C(\alpha) \bar \Delta}  \right)$ and thus the scaled (by $\sqrt{T}$) regret does not diverge as $\bar \Delta$ converges to infinity in the limit of $T$.~ 
           
         The warm-up size does not enter the bound: the proofs invoke the indicator form~\eqref{eq:postconL} of Lemma~\ref{lem:Post} only at thresholds $\Lambda > L_k(T)$, whose admissibility condition $C(\alpha)\delta_k^2\Lambda > \max\{\log(MT),1\} \geq 1$ holds by the definition of $L_k(T)$ alone. Accordingly $n_{k,0}$ may be any constant $n_0 \geq 1$: it is independent of the horizon $T$ and of the unknown gaps, the algorithm needs no knowledge of $T$, and the burn-in costs $Kn_0 = O(K)$ observations, a lower-order term. The constant $n_0$ influences the analysis only through $r_0$, which enters Lemma~\ref{lem:Int2} through the finite-sample BvM approximation; since $n_0$ appears there only inside a factor that increases to $1$, taking $n_0$ larger cannot reduce $r_0$ below its limiting value, and a moderate constant already realizes it up to a small factor. 
         
         Moreover, note that the factor $9$ in the definition of $\delta_k$ is merely an artifact of our chosen instance, where $\delta_k = \Delta_k / 3$, and can be made substantially smaller without affecting the analysis. Indeed, the analysis only requires $u_k < y_k$, that is $\mu_{0,k} + \delta_k < \mu_{0,i^*} - \delta_k$, which holds precisely when $\delta_k < \Delta_k/2$. Any choice $\delta_k = \Delta_k/c$ with $c > 2$ is therefore admissible, and since $L_k(T) = \max\{\log(MT),1\}/(C(\alpha)\delta_k^2)$, such a choice replaces the factor $9$ by $c^2$ in every leading constant of Theorems~\ref{thm:PIRB0}--\ref{thm:RB-PD}, the infimum being $4$ as $c \downarrow 2$. For instance $c = 2.1$ gives $4.41$ in place of $9$ and $6.615$ in place of $27/2$, improving these constants by roughly a factor of two. We retain $c = 3$ throughout for readability.
          
         The instance-dependent bound above yields the state-of-the-art instance independent regret bound below. 
        \begin{theorem}~\label{thm:PIRB}
            For any $k\in[K]$ fix $\delta_k=\Delta_k/3$ .  Then, under Assumptions~\ref{ass:reward},~\ref{ass:Renyi},~\ref{ass:prior2},~\ref{ass:Spokoiny}, and~\ref{Ass:Prior2},  for a warm-up size $n_{k,0} = n_0$, where $n_0 \geq 1$ is any constant independent of $T$ and of the gaps, $\kappa \leq (1-e^{-4})^2/48$ and $C < \sqrt{2}$, we have
        \begin{align}
            \bbE  \left[ \sum_{t=1}^{T} (\mu_{0,i^*} -  \mu_{0,i_t})  \right] = \mathcal{O}\left(K+\frac{\sqrt{KT\log(K)}}{\sqrt{C(\alpha)}} \right) ,
        \end{align}
        and for $\frac{K}{\log(K)}\leq \frac{T}{C(\alpha)}$, $\bbE  \left[ \sum_{t=1}^{T} (\mu_{0,i^*} -  \mu_{0,i_t})  \right] = \mathcal{O}\left(\frac{\sqrt{KT\log(K)}}{\sqrt{C(\alpha)}} \right) $.
        \end{theorem}
Note that the above regret upper bound is similar to the best known regret bound computed by~\cite{agrawal2013further} for TS algorithm with Gaussian prior. However, our regret bound above generalizes to any prior and reward distribution. Also, our bound is away from the minimax optimality by a factor of $\mathcal{O}(\sqrt{\log K})$. We believe that the ideas developed in~\cite{jin21d} can be combined
with our general $\alpha-$posterior contraction rate in Lemma 3.1 to yield minimax optimal regret bounds. 

Finally, we compute an instance-dependent bound on the expected regret of the $\alpha$-TS algorithm, that is close to the lower bound~\eqref{eq:LB} computed in~\cite{lai1985asymptotically,burnetas1996optimal}. We will later see in the proof that the arguments required to prove the following instance-dependent bound and the one in~Theorem~\ref{thm:PIRB0} are very similar.
\begin{theorem}~\label{thm:RB-PD}
    For any $k\in[K]$, fix $\delta_k=\Delta_k/3$, .  Then, under Assumptions~\ref{ass:reward},~\ref{ass:Renyi},~\ref{ass:prior2},~\ref{ass:Spokoiny}, and~\ref{Ass:Prior2},  for a warm-up size $n_{k,0} = n_0$, where $n_0 \geq 1$ is any constant independent of $T$ and of the gaps, $\kappa \leq (1-e^{-4})^2/48$ and $C < \sqrt{2}$, we have
        \begin{align}
            \bbE  \left[ \sum_{t=1}^{T} (\mu_{0,i^*} -  \mu_{0,i_t})  \right] \leq \sum_{k \neq i^*} \Delta_k\left(9(r_0+1) \frac{\log(T)}{C(\alpha)\Delta_k^2}  + \frac{(2r_0+3)}{2} \right).
        \end{align}
\end{theorem}

The above upper-bound matches (up to a constant factor and additive term) to the lower bound provided in~\cite{lai1985asymptotically,burnetas1996optimal} for all discrete and Gaussian reward (with fixed variance) distributions. It follows because for Gaussian reward with fixed variance $(\Delta_k)^2=2\scKL(\theta_{i^*}^0,\theta_{k}^0)$ and for  discrete rewards $(\Delta_k)^2 $ is bounded below by  $\scKL(\theta_{i^*}^0,\theta_{k}^0)$ (up to a known instance-independent constant) due to reverse Pinsker's inequality~\cite[Lemma 6.3]{Csiszar2006}. However, our bound in~Theorem~\ref{thm:RB-PD} does not exactly recover the lower bounds of~\cite{lai1985asymptotically,burnetas1996optimal} for general reward distributions. Nonetheless, ~\cite{agrawal2012analysis} and~\cite{Auer2002} computes instance-dependent regret bounds that depend on $\Delta_k^2$ instead of $\scKL(\theta_{i^*}^0,\theta_{k})$ for TS and UCB1 respectively. In addition note that, our regret upper bound in Theorem~\ref{thm:RB-PD} also have an additional factor $C(\alpha)^{-1}$ with other constants, that makes it away from being asymptotically optimal (in constants) according to~\cite{lai1985asymptotically}.

\begin{figure}[ht]
\centering
    \includegraphics[width=0.99\textwidth]{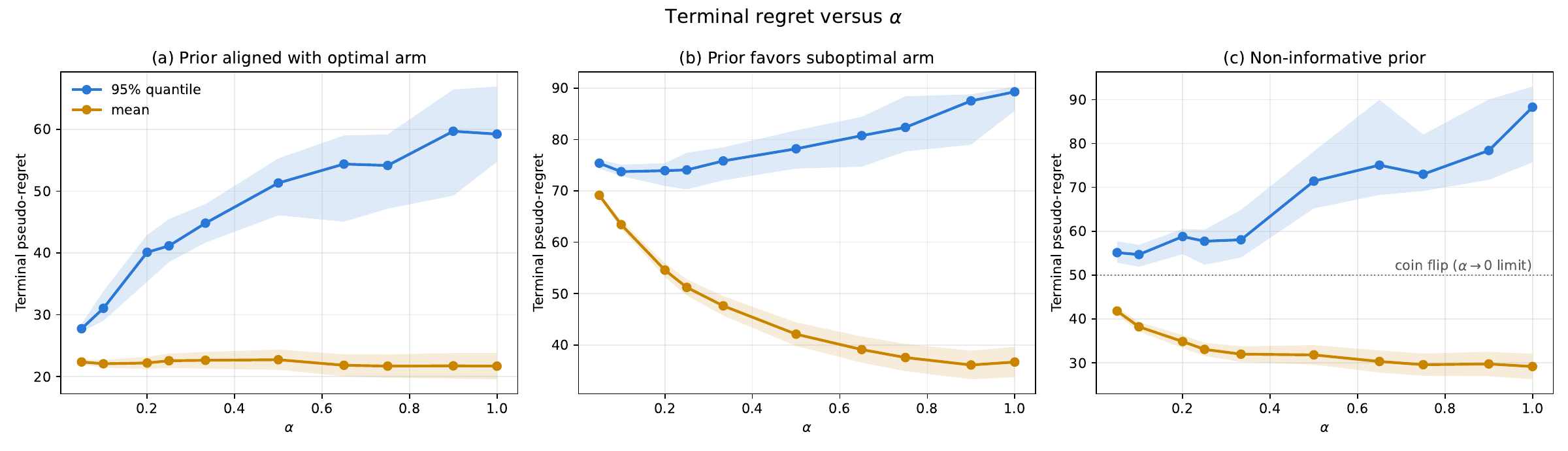}
    
    \caption{Terminal regret versus $\alpha$ on a grid extending down to $\alpha = 0.05$, for a two-armed Bernoulli instance with gap $\Delta = 0.02$ ($T = 5{,}000$, $300$ paired replications). Curves are the mean and the $95$th percentile of terminal regret; shaded bands are $\pm 2$ standard errors and bootstrap confidence intervals respectively. {Panels (a) and (b) use an informative prior.} In the misspecified environment {(b)} the tail is heaviest at $\alpha = 1$ ($q_{95} = 89.3$, versus $\approx 74$ under moderate tempering) while the mean increases as $\alpha$ decreases; in the aligned environment {(a)} the mean is essentially flat while the tail still lightens as $\alpha$ decreases. {Panel (c) is a control run with a non-informative $\mathrm{Beta}(1,1)$ prior: even with no prior to mislead the algorithm the tail is again heaviest at $\alpha = 1$ ($q_{95} = 88.3$, versus $\approx 55$ under moderate tempering), so the heavy tail is produced by the untempered posterior collapsing onto a wrong arm after early unlucky draws rather than by a misspecified prior}}
  \label{fig:TailAlpha}
\end{figure}

Beyond the mean, the parameter $\alpha$ shapes the \emph{distribution} of the regret. Figure~\ref{fig:TailAlpha} reports, on a two-armed Bernoulli instance, the mean and the $95$th percentile of terminal regret as functions of $\alpha$. The upper tail is heaviest for $\alpha$ close to $1$, where standard Thompson Sampling sits, and is lighter under moderate tempering, while the mean moves in the opposite direction when the prior is misspecified. Thus $\alpha$ trades mean regret against regret dispersion, which is the practical reason to prefer $\alpha < 1$ when a single bad run is costly. Appendices~\ref{app:More}--\ref{app:Fancy} develop these experiments in full, including a non-informative prior, comparisons with KL-UCB, kl-UCB++, IMED and $\varepsilon$-TS, and non-conjugate priors.

\subsection{Proof sketch}
First, we present our main lemma, where we compute a finite sample sub-exponential bound, valid at any bounded stopping time, on the expected $\alpha$-posterior measure of a complement of a ball of radius $\delta_k$ centered at the true parameter $\theta_{0,k}$, where the ball is defined using the difference in the observed and the true mean reward.   
        \begin{lemma}[$\alpha$-posterior concentration]~\label{lem:Post}
            Fix {$k\in [K]$,} and $\alpha\in(0,1)$ . Under  Assumptions~\ref{ass:Renyi} and~\ref{ass:prior2}, for  any bounded stopping time $\tau \leq T$ with respect to the filtration $\mathcal{F}_s := \sigma\left(X_s, \{i_r\}_{r=1}^s\right)$, provided $C(\alpha)\delta_k^2 n_{k,0} \geq 1$ (this condition is needed only for the first display below)  and $\kappa \leq (1-e^{-4})^2/48$, 
            \begin{align}
                \bbE& \left[ \bbE\left[ \ind\{|\mu_k^{\tau} -\mu_{0,k} |^2 \geq \delta_k^2 \} | X_{\tau} \right] e^{C(\alpha) \bar n_k(\tau) \delta_k^2} \right] 
                \leq 2^{-1},
                \label{eq:postcon}
            \end{align}
            Moreover, for any deterministic $\Lambda$ satisfying $C(\alpha)\delta_k^2 \Lambda \geq 1$, we also have the indicator form, which does \emph{not} require the condition $C(\alpha)\delta_k^2 n_{k,0} \geq 1$:
            \begin{align}
                \bbE \left[ \bbE\left[ \ind\{|\mu_k^{\tau} -\mu_{0,k} |^2 \geq \delta_k^2 \} | X_{\tau} \right] \ind\{\bar n_k(\tau) \geq \Lambda\} \right] 
                \leq 2^{-1} e^{ - C(\alpha) \Lambda \delta_k^2 },
                \label{eq:postconL}
            \end{align}
            where $\bar n_k(\tau) := n_k(\tau)+n_{k,0}$ counts the pulls of arm $k$ up to time $\tau$ including the warm-up samples, $\mu_k^{\tau}$ denotes the mean corresponding to one draw from the $\alpha$-posterior $\pi_{\alpha}(\cdot|X_{\tau})$ 
            , and $C(\alpha) = D (1-\alpha) \min \left(2\alpha,1-\alpha  \right)/16$ .
        \end{lemma}
         The results in Lemma~\ref{lem:Post} differs from the result in~\citep[Theorem 3.1 and Corollary 3.2]{bhattacharya2019bayesian} due to $\bar n_k(t)$, which is stochastic and also due to the bounds, which are on expectation with respect to the data generating distribution instead of high-probability bounds in~\citep{bhattacharya2019bayesian}.   
         
         The proof of Lemma~\ref{lem:Post} requires no independence of the observed rewards conditionally on the pull counts or on the action sequence. The data-dependent number of pulls $\bar n_k(\tau)$ is handled by an optional-stopping argument: for each fixed $\theta_k$, the tempered likelihood ratio normalized by its per-observation mean is a nonnegative martingale with respect to the natural filtration of the algorithm, irrespective of how the (adaptive, possibly randomized) actions are selected, and the concentration bound follows from the optional stopping theorem, with the random pull count $\bar n_k(\tau)$ retained in the statement through an exponential compensation factor. 
         The proof also uses the observation that the inner expectation in the RHS of~\eqref{eq:postcon} is bounded above by $ \bbE\left[ \ind\{D_{\alpha}(\theta_k,\theta_{0,k}) \geq D \frac{\alpha}{2} \delta_k^2  \} | X_{t} \right]$ due to Assumption~\ref{ass:Renyi}.   We then define a set of rewards
\(A_{\bar n_k(t)} := \left\{ X_{k,t}: \int_{\Theta} \tilde \pi_k(\theta_k) e^{-\alpha \ell_{\bar n_k(t)}(\theta_k,\theta_{0,k})}d\theta_k \leq 4^{-\alpha} e^{-D \frac{(\alpha-\alpha^2)}{8} \bar n_k(t)\delta_k^2}  \right\},\) where for any $B\subset \Theta$, $\tilde \Pi_k(B) = \Pi_k(B\cap \Theta)/\Pi_k(B\left(\theta_{0,k},C(\alpha)\delta_k^2 \right))$ and $\ell_{\bar n_k(t)}(\theta_k,\theta_{0,k})= \log \prod_{i=1}^{\bar n_k(t)}$ $  p(x_{i,k}|\theta_{0,k})
- \log {\prod_{i=1}^{\bar n_k(t)} p(x_{i,k}|\theta_k)}$, and $\bar n_k(t)= n_k(t)+n_{k,0}$, . Intuitively, the set $A_{\bar n_k(t)}$ is a collection of observations for which evidence is exponentially small even when the prior has a sufficient mass around the true model. Then, we decompose the expected $\alpha$-posterior probability of $\{D_{\alpha}(\theta_k,\theta_{0,k}) \geq D \frac{\alpha}{2} \nabla_k^2  \}$ by dividing it on $A_{\bar n_k(t)}$ and its complement. We further decompose the set $\{D_{\alpha}(\theta_k,\theta_{0,k}) \geq D \frac{\alpha}{2} \nabla_k^2  \}$ into union of concentric shells  $\bigcup_{j\geq 1} \Big\{D\frac{\alpha}{2} (j+1) \delta_k^2 \geq D_{\alpha}(\theta_k,\theta_{0,k}) \geq D j \frac{\alpha}{2} \delta_k^2  \Big\}$ and use the definition of $A_{\bar n_k(t)}^\mathtt{c}$, $\tilde \Pi_k(B)$ and union bound with Assumption~\ref{ass:prior2} to compute a summable bound on the term with $A_{\bar n_k(t)}^\mathtt{c}$. The summand in the bound is the ratio of the prior probability of the above shells to that of the ball $B\left(\theta_{0,k},C(\alpha)\delta_k^2 \right)$  (refer display~\eqref{eq:MABDE6.5} in Appendix~\ref{app:Proofs}) . Now, Assumption~\ref{ass:prior2}, which bounds this ratio, enables us to compute a finite and exponentially decaying bound on the term with $A_{\bar n_k(t)}^\mathtt{c}$. 

When the assumption is solely on $\Pi_k(B\left(\theta_{0,k}, C(\alpha)\delta_k^2 \right))$, {\`a} la ~\cite{bhattacharya2019bayesian, ZG}, then in the $\alpha$-posterior concentration proof we can trivially bound prior probability of the union of these shells by 1, and the required bound follows from the assumed lower bound on $\Pi_k(B(\theta_{0,k}, C(\alpha)\delta_k^2))$. However, as noted in the discussion after Assumption~\ref{ass:prior2}, if we assume a similar condition as in~\cite{bhattacharya2019bayesian,ZG}, then the initial number of samples $n_{k,0}$ will be of order $\nicefrac{e^{1/\delta_k^2}}{\delta_k^2}$, which will yield a trivial regret bound. This updated proof technique makes our proof for $\alpha$-posterior concentration significantly different from the proof in~\cite{bhattacharya2019bayesian} to yield a stronger result.

    Now recall $\Delta_k = \mu_{0,i^*} -  \mu_{0,k}$ and note that the expected regret 
    \begin{align*}
       \bbE\left[\sum_{t=1}^{T} \bbE[ \mu_{0,i^*} -  \mu_{0,i_t}|X_{t-1} ]\right ]&= \bbE  \left[ \sum_{t=1}^{T}(\mu_{0,i^*} -  \mu_{0,i_t})  \right] 
       = \sum_{k\neq i^*}  \Delta_k \bbE[n_{k}(T)] 
       \\
       &= \sum_{k\neq i^*} \Delta_k \bbE\left[\sum_{t=1}^{T}\ind( i_t=k )\right].
    \end{align*}
     For the analysis only, fix $\delta_k = \Delta_k/3$ for each $k \neq i^*$ and define the thresholds $u_k := \mu_{0,k} + \delta_k$ and $y_k := \mu_{0,i^*} - \delta_k$; since $\Delta_k = 3\delta_k$, we have $u_k < y_k$. 
     Essentially, in our proofs, we decompose the expectation above on the set $\{\mu_k^t \leq y_k\} $ and its complement and bound them separately. Observe that
    \begin{align} 
        \sum_{t=1}^{T} \bbE\left[  \ind(i_t=k )\right] &=  \sum_{t=1}^{T} \bbE\left[  \ind(i_t=k )  \ind\{\mu_k^t > y_k  \}   \right]
        + \sum_{t=1}^{T} \bbE\left[  \ind(i_t=k )  \ind\{\mu_k^t \leq y_k  \}   \right] .
        \label{eq:MAIN}
        \end{align}
        In essence, the first and the second term accounts for over-estimation of the sub-optimal arms and under-estimation of the best arm respectively. 
         The bound on the first term in~\eqref{eq:MAIN} is a direct consequence of~Lemma~\ref{lem:Post} as it is easier to control the over-estimation of samples from the posterior distribution using its concentration properties. We present below the result that bounds the first term.
        \begin{lemma}~\label{lem:Int4}
            Fix $\alpha\in(0,1)$.    Under Assumptions~\ref{ass:reward}, \ref{ass:Renyi}, and \ref{ass:prior2} , $L_k(T)= \frac{\max\{\log(TM),\,1\}}{C(\alpha)\delta_k^2}$ defined for any $M>0$ , 
            we have for each $k\neq i^*$ , 
            \begin{align}
            \sum_{t=1}^{T} \bbE\left[  \ind(i_t=k )  \ind\{\mu_k^t > y_k  \}   \right] \leq L_k(T)  + \left(2M \right)^{-1}.
            \end{align}
        \end{lemma}
        Observant readers would have noted that $L_k(T)$ is the term that defines the appropriate bound on the expected regret. We anticipate to have a similar bound on the second term in~\eqref{eq:MAIN}. To analyse the second term in~\eqref{eq:MAIN}, we define $p_{k,t}:= \pi_{\alpha}( \mu_{i^*}^{t} \geq y_k | X_{t-1} )$ for any $k \neq i^*$ and bound it using the following lemma. This result is similar to~\cite[Lemma 2.8]{agrawal2013further}, which significantly simplifies the analysis of TS than the prior techniques.
        \begin{lemma}~\label{lem:Int1}
            For any $k\neq i^*$ and $t\in[T]$
            \begin{align}
                E&\left[  \ind(i_t=k )  \ind\{\mu_k^t \leq y_k  \} |X_{t-1}  \right] 
                \leq \frac{1-p_{k,t}}{p_{k,t}} P(i_t=i^* , \mu_k^t \leq y_k | X_{t-1} ).
            \end{align}
        \end{lemma}

In particular, the result above bounds the posterior probability of playing the sub-optimal arm (when the best arm is under-estimated) by a linear function of the posterior probability of playing the best arm times the odds ratio of under-estimating the best arm to over-estimating it. 
    Thereafter, we use the above lemma to show that $\sum_{t=1}^{T} \bbE \left[   \ind(i_t=k )  \ind\{\mu_k^t \leq y_k  \}   \right] \leq \sum_{j=0}^{T-1}  \bbE \left[   \frac{1-p_{k,\tau_{u}+1}}{p_{k,\tau_{u}+1}}     \right]$, where
    $\tau_{u}$ is the time-step at which the total number of observations from the best arm $i^*$, including the $n_{i^*,0}$ warm-up samples, first equals $u$, so that $\bar n_{i^*}(\tau_u) = u$. 
    
To bound $  \bbE \left[   \frac{1-p_{k,\tau_{u}+1}}{p_{k,\tau_{u}+1}}     \right]$, for $u>L_k(T)$, observe that $\frac{1-p_{k,\tau_{u}+1}}{p_{k,\tau_{u}+1}}$ is the mean of a geometric random variable denoting the number of consecutive independent trials until $\mu_{i^*}^t > y_k$ given $u$ observations from the best arm and then expressing this mean as an infinite sum of the tail probabilities of this geometric random variable.  The upper bound for $u>L_k(T)$ follows by using Lemma~\ref{lem:Post} in bounding the above described geometric tail probability and using the definition of $L_k(T)$. However, note that we also need to bound $ \bbE \left[ \frac{1-p_{k,\tau_{u}+1}}{p_{k,\tau_{u}+1}}     \right]$ for $u\leq L_k(T)$. Since the concentration results do not hold in this regime, we have to resort to some finite sample analysis of posterior distribution that can upper bound $\bbE \left[   p_{k,\tau_{u}+1}^{-1}    \right]$ for all such $u$. This is the most critical part of the TS analysis that evidently requires posterior anti-concentration bounds.
    
The work in~\cite{agrawal2013further} assumes Gaussian posterior, which enables them to conveniently compute this bound, and on the other hand~\cite{mazumdar20a} assumes sufficient structural assumption on the prior and the likelihood model so that posterior can be lower bounded by a Gaussian distribution and thus enabling them too to compute the required upper bound. However, in general, it is hard to have this nice Gaussian structure. One way to generalize the analysis is to use the finite sample Bernstein-von Mises theorem to compute a Gaussian approximation of the posterior with the aim of computing a finite sample lower bound on the probability of over-estimating the best arm. We leverage such result derived in~\cite{Spokoiny2012,Panov2015} (and reproduced in~Lemma~\ref{lem:FBVM}]) to compute the required bound.

    In the next lemma, we bound the term $\bbE \left[   \frac{1-p_{k,\tau_{u}+1}}{p_{k,\tau_{u}+1}}     \right]$. The proof technique for the second result in the following lemma is adapted from~\cite[Lemma 2.13]{agrawal2013further} and uses Lemma~\ref{lem:Post}. The first result requires developing new techniques using Lemma~\ref{lem:FBVM} and is a generalization to the similar results established in~\cite[Lemma 2.13]{agrawal2013further} and~\cite[Lemma 15]{mazumdar20a}.  
        
    \begin{lemma}~\label{lem:Int2}
        Fix $\alpha\in(0,1)$ with $C <  \sqrt{2}$, where $C$ is the constant of Lemma~\ref{lem:FBVM}.  Then, under Assumptions~\ref{ass:reward},~\ref{ass:prior2},~\ref{ass:Spokoiny}, and~\ref{Ass:Prior2}, $L_k(T)= \frac{\max\{\log(TM),\,1\}}{C(\alpha)\delta_k^2}$ defined for any $M>0$ ,  we have for each $k\neq i^*$ and $\tau_{u}\leq T$, 
            \begin{align}
                \bbE \left[   \frac{1-p_{k,\tau_{u}+1}}{p_{k,\tau_{u}+1}}     \right] \leq \begin{cases}
                    (r_0(C,\alpha,n_0,\mathbb{C})+2) & \forall  u \geq 1 
                    \\
                    \left(MT \right)^{-1}
                    & \forall u > L_k(T).
                \end{cases}
            \end{align}
        \end{lemma}
       
\noindent\textbf{Remark.} The constant $r_0 = r_0(C,\alpha,n_0,\mathbb{C},a)$ in Lemma~\ref{lem:Int2} is finite whenever $C <  \sqrt{2}$ and depends only on the model constants $C$, $\alpha$, $n_0$, and $\mathbb{C}$ (together with the free parameter $a$ used in its proof); in particular, it does not depend on the suboptimality gap $\Delta_k$ or the horizon $T$. It therefore enters the regret bounds of Theorems~\ref{thm:PIRB0}--\ref{thm:RB-PD} only as a multiplicative constant and does not affect their dependence on $T$ or $\Delta_k$. We do not optimize its value.

        Consequently, using the results above, we bound~\eqref{eq:MAIN} to first derive the instance-dependent regret bound of~Theorem~\ref{thm:PIRB0}.  In its derivation, we fix $M=C(\alpha)\delta_k^2$ in~Lemma~\ref{lem:Int2} and Lemma~\ref{lem:Int4}. We compute instance-independent bound from the instance-dependent regret bound of~Theorem~\ref{thm:PIRB0} by decomposing the expected regret on the set of arms that satisfies $\Delta_k > (e\sqrt{K\log(K)})/\sqrt{TC(\alpha)}$ and its complement. To  derive our final 
        instance-dependent regret bound in~Theorem~\ref{thm:RB-PD}, we follow the same steps as used in the proof of ~Theorem~\ref{thm:PIRB0} but with $M=1$ while using~Lemmas~\ref{lem:Int2} and~\ref{lem:Int4}.

\section{Examples}~\label{sec:Exam}
Below we specify two families of distribution that satisfies all the regularity conditions that we have used to derive the regret bounds.  The first family is the popular sub-Gaussian family of distributions. one specifies the regularity conditions for a sub-Gaussian reward. \begin{definition}[Sub-Gaussian]~\label{def:SubReward}
   For any arbitrary $\Theta$ and for each arm $k$, we assume that
    the reward distribution $P_{\theta_k}$ is sub-Gaussian with parameter $\sigma=1$, that is $\bbE_{P_{\theta_k}} [e^{s x_k}] \leq e^{s \theta_k + s^2/2}$ for all $s\in \bbR$.
    \end{definition}
    For the sub-Gaussian family defined above, we show that Assumption~\ref{ass:Renyi} is satisfied for $D=1$ in Lemma~\ref{lem:Pin}. With priors having strictly positive, bounded,  and continuous density on $\Theta$, we show in Proposition~\ref{prop:SG} that this family also satisfies Assumption~\ref{ass:prior2} for $\kappa= \frac{\max_{\theta\in\Theta} \pi_k(\theta) \sqrt{D\tilde{C}}}{\min_{B(\theta_{0,k},1)} \pi_k(\theta) \sqrt{C(\alpha)}}$ for some constant $\tilde{C}>0$. Furthermore, in Proposition~\ref{prop:SpokSG}, we show that the sub-Gaussian family also satisfies the conditions in Assumption~\ref{ass:Spokoiny}. 

Next, we specify the conditions for exponential family reward models. 
    \begin{definition}[1-d Exponential]~\label{def:rewardExp}
        We assume for $\Theta \subseteq \bbR$ that 
        \begin{itemize}
            \item[i.] $p(x|\theta_k) = e^{x \theta_k   -  A(\theta) + C(x) }$, with mean as $ A'(\theta)$ and variance as $A''( \theta)$, where $A(\cdot)$ is the log-partition function and $C(\cdot)$ is some known mapping. We also define the link function $g(\cdot)=A'(\cdot)$.  
            \item[ii.] the link function is Lipschitz continuous, that is for any $x,y\in \Theta$, there exists a constant $C_g>0$, such that
             $   |g(x)-g(y)|\leq C_g|x-y|$, and
            \item[iii.] the log-partition function is strongly convex with parameter $m$, that is for any $\gamma\in (0,1)$, $x,y\in \Theta$, there exists an $m>0$, such that
            \begin{align*}
                \gamma A(x) + (1-\gamma)A(y ) - A(\gamma x + (1-\gamma)y ) 
                \geq \frac{1}{2} \gamma (1-\gamma) m |x-y|^2.
            \end{align*}
        \end{itemize}
    \end{definition} 
  Typically, if $\max_{x\in \Theta}|g'(x)|<\infty$, then the second condition above is satisfied for $C_g= \max_{x\in \Theta}|g'(x)|$. 
    Moreover, when $A''(\cdot)\geq m >0$, then $A(\cdot)$ is strongly convex with parameter $m$. Note that conditions~~(ii) and (iii) in Definition~\ref{def:rewardExp}  above follows when $A''(\cdot)\in [m,C_g]$. This condition is restrictive for exponential family distribution as it requires the variance of a number of exponential family distributions (such as Poisson or Exponential) to lie in a compact space\footnote{For some exponential family distributions such as Gaussian and Poisson, the condition of lower boundedness of the variance for all $\theta\in \Theta$ can be relaxed by assuming a slightly weaker condition, instead of strong convexity of $A(\cdot)$. In particular, we only require that for any $\gamma\in (0,1)$, $\theta,\theta_0\in \Theta$, $
                \gamma A(x) + (1-\gamma)A(y ) - A(\gamma x + (1-\gamma)y ) 
                \geq \frac{1}{2} \gamma (1-\gamma)^2 A''(\theta_0) |\theta-\theta_0|^2.$
           In this case. we only need lower boundedness of $A''(\theta)$ for $\theta=\theta_0$.}. {Moreover, note that according to~\cite{bubeck2013bandits}, reward models with infinite variance must have finite moments of at least of order $1+\zeta, \forall \zeta\in(0,1)$  to have logarithmic regret but the regret bound for such models have sub-optimal dependence on the instance-gap, i.e., \ $(1/\Delta_i)^{1/\zeta}$.~\cite{bubeck2013bandits} also prove a lower bound that shows that this dependence is not improvable.} 
    However, note that the two assumptions above are significantly milder than the requirement of bounded support for the reward distributions, which is typically assumed in the analysis of many MAB algorithm~\citep{agrawal2012analysis,Auer2002}.
    
    For the exponential family defined above, we show that Assumption~\ref{ass:Renyi} is satisfied for $D=\nicefrac{C_g}{m}$ in Lemma~\ref{lem:Exp}. With priors having strictly positive, bounded,  and continuous density on $\Theta$, we show in Proposition~\ref{prop:ExpG} that this family also satisfies Assumption~\ref{ass:prior2} for $\kappa= \frac{\max_{\theta\in \Theta} \pi_k(\theta) \sqrt{D C_g}}{\min_{B(\theta_{0,k},1)} \pi_k(\theta) \sqrt{C(\alpha)m}}$. Furthermore, in Proposition~\ref{prop:SpokExp}, we show that the exponential family also satisfies the conditions in Assumption~\ref{ass:Spokoiny}. 
    
\section{Concluding remarks and open problems}
We believe that this work establishes a connection between the rich literature on posterior concentration in the Bayesian statistics~\citep{GGV,Kleijn2006,Ghosal_2007,bhattacharya2019bayesian,ZG,Spokoiny2012,Panov2015}
and the existing theory for TS to generalize the theoretical analysis of TS. The ideas developed in this work can be developed further by leveraging the generality of the existing posterior concentration results (and also of its variational approximations)~\citep{GGV,ZG,bhattacharya2019bayesian} to analyze other versions of TS that are designed to solve more complex sequential decision-making problems efficiently~\citep{Russo20}. 
We also note that $\alpha$-posteriors are used here as a convenient technical device to obtain regret bounds under a minimal prior mass condition.  However, with additional work, we expect the results to extend to $\alpha =1$. It requires the development of some additional regularity conditions on the prior and the reward distribution that is in line with the theory for standard posterior distribution in~\cite{GGV}. In particular, we need the existence of non-trivial testing/ entropy conditions~[Theorem 2.1 and 7.1]\citep{GGV}. This point has also been discussed in~\citet{bhattacharya2019bayesian}, arguing the simplified concentration analysis of $\alpha$-posterior compared to that of the standard posterior. We provide guidelines to extend our analysis to $\alpha=1$ in~Appendix~\ref{app:StanTS}. 

Furthermore, we conjecture that for any prior distribution under certain assumptions, the problem-independent regret lower-bound for $\alpha$-TS (and TS) will also be of the order $\Omega(\sqrt{K T \log K })$ as derived for the Gaussian priors in~\cite{agrawal2013further}. The analysis would require establishing a lower bound for the posterior concentration, a complementary result to Lemma~\ref{lem:Post}, which is an interesting problem by itself.

\bibliographystyle{plainnat}
\bibliography{refs}

\appendix
\makeatletter
\renewcommand\@seccntformat[1]{%
  \ifnum\pdf@strcmp{#1}{section}=0 \appendixname~\csname the#1\endcsname.\quad
  \else \csname the#1\endcsname\quad\fi}
\makeatother

\section{Notations}
\printnomenclature

\section{Proofs}\label{app:Proofs}

We first prove two technical lemmas that establish the relation between the absolute difference in true and observed means and the $\alpha-$R\'enyi divergence for sub-Gaussian and Exponential family reward distributions. These results could be of independent interest too. 

\begin{lemma}[R\'enyi transportation cost inequality]~\label{lem:Pin}
    Fix $\alpha\in (0,1)$. For any two sub-Gaussian measures $\mu$ and $\nu$ with sub-Gaussian parameter $\sigma_{\mu}$ and $\sigma_{\nu}$ respectively, such that $\nu$ is absolutely continuous wrt $\mu$, that is $\nu << \mu$, the $\alpha-$R\'enyi divergence can be bounded below by the absolute difference between the respective means. In particular, for any random variable $X$ having measure $\mu$ and $\nu$, we have  
    \begin{align}
        |\bbE_{\nu}[X]-\bbE_{\mu}[X]  | &\leq \sqrt{(\sigma_{\mu}^2\alpha+\sigma_{\nu}^2(1-\alpha))} \sqrt{\frac{2}{\alpha} D_{\alpha}(\nu\|\mu) }.
        \end{align} 
\end{lemma}

\begin{proof}[Proof of Lemma~\ref{lem:Pin}]
    For any $\alpha\in (0,1)$, recall the definition of $\alpha-$ R\'enyi divergence  $D_{\alpha}(\nu\|\mu)=   \log \int g^{\alpha}d\mu =\frac{1}{\alpha-1} {\log \int \left(\frac{d\nu}{d\mu}\right)^{\alpha}d\mu}   $, where $g\equiv \frac{d\nu}{d\mu}$. Now observe that
    \begin{align}
    \nonumber
       (\alpha-1) D_{\alpha}(\nu\| \mu) &= \log \bbE_{\nu}[ g^{\alpha-1} e^{-(\alpha-1)X} e^{(\alpha-1)X}  ] = \log \int (ge^{-X})^{\alpha-1}  e^{(\alpha-1)X} d\nu 
       \\
       \nonumber
       &\leq \log \left( \int   e^{(\alpha-1)X} d\nu \right)^{\alpha} \left(\int (ge^{-X})^{-1}  e^{(\alpha-1)X} d\nu \right)^{1-\alpha}
       \\
       \nonumber
       &= \log \left( \bbE_{\nu}[  e^{(\alpha-1)X}]^{\alpha} \bbE_{\nu}[g^{-1}  e^{\alpha X} ]^{1-\alpha} \right)
       \\
       \nonumber
       &= \log \left( \bbE_{\nu}[  e^{(\alpha-1)X}]^{\alpha} \bbE_{\mu}[ e^{\alpha X} ]^{1-\alpha} \right)
       \\
       &= \alpha \log \bbE_{\nu}[  e^{(\alpha-1)X}] + (1-\alpha) \log  \bbE_{\mu}[ e^{\alpha X} ],
    \end{align}
    where the first inequality follows from the H\"older's inequality wrt measure $e^{(\alpha-1)X} d\nu$.

    Now fix $X=s X$ for any $s\in \bbR$ (without loss of generality). Since, $\alpha\in(0,1)$, it follows from the inequality above that
    \begin{align}  
        \nonumber
        D_{\alpha}(\nu\|\mu)  &\geq \left[ \frac{\alpha}{\alpha-1}\log \bbE_{\nu} [e^{(\alpha-1)sX} ] - \log \bbE_{\mu}[\int e^{\alpha sX} ] \right] 
        \\
        \nonumber
        &\geq \left[ \frac{\alpha}{\alpha-1}[s(\alpha-1)\bbE_{\nu}[X]+\sigma_{\nu}^2s^2(\alpha-1)^2/2] - [s\alpha\bbE_{\mu}[X]+\sigma_{\mu}^2s^2\alpha^2/2]\right] 
        \\
        &= \left[ s\alpha[\bbE_{\nu}[X]-\bbE_{\mu}[X]] - \frac{s^2\alpha}{2}[\sigma_{\nu}^2(1-\alpha)+\sigma_{\mu}^2\alpha]\right] ,
    \end{align}
    where the second inequality uses the fact that $\mu$ and $\nu$ are sub-Gaussian measures.
    Recall if $X\sim\mu$, is a sub-Gaussian random variable, then $\bbE_{\mu}[e^{sX}] \leq e^{s\bbE_{\mu}[X]+\sigma_{\mu}^2s^2/2}$ for all $s\in \bbR$.
    Now it follows from above that
    \begin{align}
        \nonumber
        |\bbE_{\nu}[X]-\bbE_{\mu}[X]  | &\leq \inf_{|s|} \frac{D_{\alpha}(\nu\|\mu)}{|s|\alpha} + \frac{|s|}{2}(\sigma_{\mu}^2\alpha+\sigma_{\nu}^2(1-\alpha))
        \\
        &= \sqrt{(\sigma_{\mu}^2\alpha+\sigma_{\nu}^2(1-\alpha))} \sqrt{\frac{2}{\alpha} D_{\alpha}(\nu\|\mu) }.
    \end{align}
    \end{proof}

Next, we derive a similar bound for exponential family reward distributions.
\begin{lemma}~\label{lem:Exp}
    Fix $\alpha\in (0,1)$. Let $r$ be a random variable having its distributions lying an exponentially family with parameters $\theta$ and $\theta_0$ but same $A(\cdot)$ and $C(\cdot)$. Then under Definition~\ref{def:rewardExp} the $\alpha-$R\'enyi divergence can be bounded below by the absolute difference between the respective means. In particular, we have  
    \begin{align}
        |\bbE_{\theta}[r]-\bbE_{\theta_0}[r]  | &\leq \frac{C_g}{\sqrt{m}} \sqrt{\frac{2}{\alpha} D_{\alpha}(\theta,\theta_0) }.
        \end{align} 
\end{lemma}
\begin{proof}[Proof of Lemma~\ref{lem:Exp}]
        Recall that, for $\alpha\in(0,1)$, the $\alpha$-R\'enyi divergence between two exponential family distributions with same $A(\cdot)$ and $C(\cdot)$ with parameters $\theta$ and $\theta_0$, can be expressed as
        \[D_{\alpha}(\theta,\theta_0) = \frac{1}{1-\alpha}\left[\alpha A(\theta) + (1-\alpha)A(\theta_0 ) - A(\alpha \theta + (1-\alpha)\theta_0 ) \right].
        \]
    Using Definition~\ref{def:rewardExp}~(iii), we can lower bound $D_{\alpha}(\theta,\theta_0)$ by $\frac{\alpha m}{2} |\theta-\theta_0|^2$. Now the lemma follows immediately using~Definition~\ref{def:rewardExp}~(ii).
    
    \end{proof}   

Next, we provide proof for the finite sample Bernstein-von Mises for the $\alpha$-posterior distributions.  
\begin{proof}[Proof of Lemma~\ref{lem:FBVM}]
    Using~\cite[Theorem 4]{Panov2015} for any $\eta >0$, we have 
    \begin{align}
        \pi_{\alpha}(\sD_0(\theta- \theta_n^*)\in A|X_n) \geq \exp\left(-2\sqrt{\frac{1+ \eta}{n}} - 8e^{-\eta}\right) \left\{P(\phi/\sqrt{\alpha} \in A)\right\} -e^{-\eta},
    \end{align}
    with $P_0$-probability of at least $1- e^{-\eta}$. For the second assertion, observe that for $\theta^*_n = \theta_0 + \sD_0^{-2} \nabla\sL(\theta_0)$, we have 
    \begin{align}
    |\theta_0- \theta^*_n|  &\leq  |\sD_0^{-1}||\sD_0^{-1} \nabla\sL(\theta_0)| 
    \leq C\sqrt{\frac{1+\eta}{n I(\theta_0)}},
\end{align}
occurs with $P_0$ probability of at least $1- e^{-\eta}$,
where the bound on $|\sD_0^{-1} \nabla\sL(\theta_0)|$ uses the result in display (33) of~\cite[Theorem 9]{Panov2015} for some universal constant $C$ and the definition of $\sD_0$. 
    
\end{proof}

The proof of Lemma~\ref{lem:Post} is motivated from the ideas used in the proof of~\cite[Theorem 3.1]{bhattacharya2019bayesian} and~\cite[Theorem 2.1]{ZG}. 

The proof of Lemma~\ref{lem:Post} relies on the following auxiliary optional-stopping bound, which is the only place in the argument where the randomness of the pull counts interacts with the observed rewards. Recall that $\mathcal{F}_s := \sigma\left(X_s, \{i_r\}_{r=1}^s\right)$ denotes the natural filtration generated by the warm-up samples, the selected arms, and the observed rewards, and that $\bar n_k(s) = n_k(s) + n_{k,0}$.

\begin{lemma}[Adaptively stopped likelihood ratios]\label{lem:MG}
Fix $k \in [K]$ and a \emph{fixed, nonrandom} parameter $\theta_k \in \Theta$. Let $g \geq 0$ be a measurable function on the reward space with $m_g(\theta_k) := \bbE_{x \sim P_{\theta_{0,k}}}[g(x)] \in (0,\infty)$; note that $m_g(\theta_k)$ is a deterministic constant. Then, for any bounded stopping time $\tau \leq T$ with respect to $(\mathcal{F}_s)$, any deterministic $\Lambda \geq 0$, and any $\rho > 0$ with $\rho\, m_g(\theta_k) \leq 1$,
\begin{align}
    \bbE\left[ \rho^{\bar n_k(\tau)} \prod_{i=1}^{\bar n_k(\tau)} g(x_{i,k})\, \ind\{\bar n_k(\tau) \geq \Lambda\} \right] \leq \big(\rho\, m_g(\theta_k)\big)^{\Lambda}.
    \label{eq:MGkey}
\end{align}

\end{lemma}

\begin{proof}
Define
\begin{align}
    W_s := m_g(\theta_k)^{-\bar n_k(s)} \prod_{i=1}^{\bar n_k(s)} g(x_{i,k}), \qquad s=0,1,\ldots,T.
    \label{eq:MGdef}
\end{align}
Then $(W_s)_{s \leq T}$ is a nonnegative martingale with respect to $(\mathcal{F}_s)$ with $\bbE[W_0]=1$. Indeed, the $n_{k,0}$ warm-up samples are drawn i.i.d.\ from $P_{\theta_{0,k}}$ independently of everything else, so $\bbE[W_0]=1$; and at any round $s$, conditionally on $\mathcal{F}_{s-1}$ and on the selected arm $i_s$, the update factor in $W_s$ equals $1$ when $i_s \neq k$, while when $i_s = k$ the newly observed reward is drawn from $P_{\theta_{0,k}}$ independently of the past and of the (possibly randomized) selection rule, so the factor $g(x_{\bar n_k(s),k})/m_g(\theta_k)$ has conditional mean $1$. We emphasize that no independence of the observed rewards conditionally on the action sequence or on the pull counts is used; the adaptivity of the actions is absorbed entirely by the tower property. By the optional stopping theorem, $\bbE[W_{\tau}]=1$. Consequently, since $\rho\, m_g(\theta_k) \leq 1$ is a deterministic constant and $\bar n_k(\tau) \geq \Lambda$ holds on the displayed event,
\begin{align*}
    \bbE\left[ \rho^{\bar n_k(\tau)} \prod_{i=1}^{\bar n_k(\tau)} g(x_{i,k})\, \ind\{\bar n_k(\tau) \geq \Lambda\} \right]
    &= \bbE\left[ W_{\tau}\, \big(\rho\, m_g(\theta_k)\big)^{\bar n_k(\tau)}\, \ind\{\bar n_k(\tau) \geq \Lambda\} \right]
    \\
    &\leq \big(\rho\, m_g(\theta_k)\big)^{\Lambda}\, \bbE[W_{\tau}] = \big(\rho\, m_g(\theta_k)\big)^{\Lambda}.
\end{align*}
\end{proof}

\begin{proof}[Proof of Lemma~\ref{lem:Post}]
    Using Lemma~\ref{lem:Pin} for sub-Gaussian rewards satisfying conditions in Definition~\ref{def:SubReward}   and Lemma~\ref{lem:Exp} for Exponential family rewards satisfying Definition~\ref{def:rewardExp}, we have
    \begin{align}
        \left|   \mu_k-\mu_{0,k}  \right| 
        \leq \frac{1}{\sqrt{D}}\sqrt{\frac{2D_{\alpha}(\theta_k,\theta_{0,k}) }{\alpha}},
        \label{eq:Pins}
        \end{align}
    where $D=1$ for sub-Gaussian rewards and $D=\frac{m}{C_g^2}$ for exponential family rewards. We omit using superscript $t$ in $\mu_k$ and $\theta_k$ for brevity.
    Therefore, it follows from the inequality above that 
    \begin{align}
        E\left[ \ind\{|\mu_k -\mu_{0,k} |^2 \geq \delta_k^2 \} | X_{t} \right] \leq E\left[ \ind\{D_{\alpha}(\theta_k,\theta_{0,k}) \geq D \frac{\alpha}{2} \delta_k^2  \} | X_{t} \right].
        \label{eq:MABDE2}
    \end{align} 
   Now for any $k\in[K]$, let us define a set 
    \begin{align}
        F_k = \left\{\theta_k \in \Theta: D_{\alpha}(\theta_k,\theta_{0,k}) \geq D \frac{\alpha}{2} \delta_k^2  \right\}.
        \label{eq:MABDE2b}
    \end{align}

 and define $\bar n_k(t)=n_k(t)+n_{k,0} =\sum_{s=1}^{t}  \ind(i_s = k) + n_{k,0} $. Observe that 
\begin{align}
    \nonumber
    \Pi_{\alpha}(F_k|X_{t})&= \frac{\int_{F_k} \pi_k(\theta_k) p(X_{k,0}|\theta_{k})^{\alpha}\prod_{s=1}^{t} p(x_{i_s}|\theta_{i_s})^{\alpha \ind(i_s = k)} d\theta_k} {\int \pi_k(\theta_k) p(X_{k,0}|\theta_{k})^{\alpha}\prod_{s=1}^{t} p(x_{i_s} |\theta_{i_s})^{\alpha \ind(i_s = k)} d\theta_k }
        \\
        \nonumber
        &= \frac{\int_{F_k} \pi_k(\theta_k) \prod_{i=1}^{\bar n_k(t)} p(x_{i,k}|\theta_k)^{\alpha }d\theta_k} {\int \pi_k(\theta_k) \prod_{i=1}^{\bar n_k(t)} p(x_{i,k}|\theta_k)^{\alpha }d\theta_k }
        \\
        &= \frac{\int_{F_k} \pi_k(\theta_k) e^{-\alpha \ell_{\bar n_k(t)}(\theta_k,\theta_{0,k})}d\theta_k} {\int \pi_k(\theta_k) e^{-\alpha \ell_{\bar n_k(t)}(\theta_k,\theta_{0,k})}d\theta_k },
        \label{eq:MABDE3}
\end{align}
where $\ell_{\bar n_k(t)}(\theta_k,\theta_{0,k})= \log \frac{\prod_{i=1}^{\bar n_k(t)} p(x_{i,k}|\theta_{0,k})}{\prod_{i=1}^{\bar n_k(t)} p(x_{i,k}|\theta_k)}$.
Next define a set 
\[A_{\bar n_k(t)} = \left\{ X_{k,t}: \int_{\Theta} \tilde \pi_k(\theta_k) e^{-\alpha \ell_{\bar n_k(t)}(\theta_k,\theta_{0,k})}d\theta_k \leq 4^{-\alpha} e^{-D \frac{(\alpha-\alpha^2)}{8} \bar n_k(t)\delta_k^2}  \right\},\] where for any $B\subset \Theta$, $\tilde \Pi_k(B) = \Pi_k\left(B\cap \Theta\right)/\Pi_k\left(B\left(\theta_{0,k},\delta_k \right)\right)$.
Fix a bounded stopping time $\tau \leq T$ with respect to the filtration $\mathcal{F}_s := \sigma\left(X_s, \{i_r\}_{r=1}^s\right)$; recall that $C(\alpha)\delta_k^2 n_{k,0} \geq 1$ by assumption. We abbreviate $\bar n_k := \bar n_k(\tau)$; all pathwise identities above are evaluated at time $\tau$. Our only probabilistic tool is the optional-stopping bound of Lemma~\ref{lem:MG}. 
In view of~\eqref{eq:MABDE2} and~\eqref{eq:MABDE3}, it suffices to bound the compensated posterior mass
\begin{align}
     \bbE\left[\Pi_{\alpha}(F_k|X_{\tau})\, e^{C(\alpha) \bar n_k \delta_k^2}\right] &\leq \bbE\left[\ind\big(A_{\bar n_k}\big)\, e^{C(\alpha) \bar n_k \delta_k^2}\right] \nonumber
     \\
     &\quad + \bbE \left[  \frac{\int_{F_k} \pi_k(\theta_k) e^{-\alpha \ell_{\bar n_k}(\theta_k,\theta_{0,k})}d\theta_k} {\int \pi_k(\theta_k) e^{-\alpha \ell_{\bar n_k}(\theta_k,\theta_{0,k})}d\theta_k } \ind\big(A_{\bar n_k}^\mathtt{c}\big)\, e^{C(\alpha) \bar n_k \delta_k^2} \right].
    \label{eq:MABDE4}
\end{align}

\emph{Second term.} On the set $A_{\bar n_k}^\mathtt{c}$, the definitions of $A_{\bar n_k}$ and of $\tilde \pi_k$ give the pathwise lower bound $\int \pi_k e^{-\alpha \ell_{\bar n_k}} \geq \Pi_k(B\left(\theta_{0,k},\delta_k \right))\, 4^{-\alpha} e^{-D \frac{(\alpha-\alpha^2)}{8} \bar n_k \delta_k^2}$ on the denominator. Writing $F_k= \bigcup_{j\geq 1} F_k^j$ with $F_k^j := \left\{\theta_k \in \Theta: D\frac{\alpha}{2} (j+1) \delta_k^2 \geq D_{\alpha}(\theta_k,\theta_{0,k}) \geq D j \frac{\alpha}{2} \delta_k^2  \right\}$, we obtain the pathwise bound
\begin{align}
    &\frac{\int_{F_k} \pi_k e^{-\alpha \ell_{\bar n_k}}} {\int \pi_k e^{-\alpha \ell_{\bar n_k}} } \ind\big(A_{\bar n_k}^\mathtt{c}\big)\, e^{C(\alpha) \bar n_k \delta_k^2} \nonumber \\
    &\qquad \leq \frac{4^{\alpha}}{\Pi_k(B\left(\theta_{0,k},\delta_k \right))} \sum_{j \geq 1} \int_{F_k^j} \pi_k(\theta_k)\, e^{-\alpha \ell_{\bar n_k}(\theta_k,\theta_{0,k})}\, e^{\left(D \frac{(\alpha-\alpha^2)}{8} + C(\alpha)\right) \bar n_k \delta_k^2}\, d\theta_k.
    \label{eq:MABDE6}
\end{align}
For a fixed $\theta_k$, note that $e^{-\alpha \ell_{\bar n_k}(\theta_k,\theta_{0,k})} = \prod_{i=1}^{\bar n_k} g_1(x_{i,k})$ for $g_1(x) = \left( p(x|\theta_k)/p(x|\theta_{0,k})\right)^{\alpha}$, and
\begin{align}
    m_{g_1}(\theta_k) = \int p(x|\theta_k)^{\alpha}\, p(x|\theta_{0,k})^{1-\alpha}\, dx = e^{-(1-\alpha) D_{\alpha}(\theta_k,\theta_{0,k})}.
    \label{eq:Iden1}
\end{align}
On $F_k^j$ we have $D_{\alpha}(\theta_k,\theta_{0,k}) \geq D j \frac{\alpha}{2}\delta_k^2$, so taking $\rho = e^{\left(D\frac{(\alpha-\alpha^2)}{8} + C(\alpha)\right)\delta_k^2}$ and using $C(\alpha) \leq D\frac{(\alpha-\alpha^2)}{8}$ gives $\rho\, m_{g_1}(\theta_k) \leq e^{-D \frac{(4j-2)(\alpha-\alpha^2)}{8}\delta_k^2} \leq 1$. Applying Lemma~\ref{lem:MG} (with $g = g_1$, this $\rho$, and $\Lambda = n_{k,0}$, whose indicator is vacuous) for each fixed $\theta_k \in F_k^j$ and Fubini--Tonelli,
\begin{align}
    \bbE\left[ \int_{F_k^j} \pi_k(\theta_k)\, e^{-\alpha \ell_{\bar n_k}(\theta_k,\theta_{0,k})}\, e^{\left(D \frac{(\alpha-\alpha^2)}{8} + C(\alpha)\right) \bar n_k \delta_k^2}\, d\theta_k\right] \leq \Pi_k(F_k^j)\, e^{-D \frac{(4j-2)(\alpha-\alpha^2)}{8}\, n_{k,0}\, \delta_k^2 }.
    \label{eq:MABDE6.5}
\end{align}
Therefore, using Assumption~\ref{ass:prior2} to bound the ratios $\Pi_k(F_k^j)/\Pi_k(B\left(\theta_{0,k},\delta_k \right)) \leq \kappa (j+1)$, the expectation of the second term in~\eqref{eq:MABDE4} is bounded by
\begin{align}
    \kappa\, 4^{\alpha}\sum_{j\geq 1} (j+1)\, \xi^{4j-2} \leq \frac{3\kappa 4^{\alpha}}{(1-e^{-4})^2}\, \xi^{2} \leq 4^{-1}.
    \label{eq:MABDE7}
\end{align}
Here $\xi := e^{-D \frac{(\alpha-\alpha^2)}{8}\, n_{k,0}\, \delta_k^2 }$, the series computation $\sum_{j\geq 1}(j+1)\xi^{4j-2} \leq 3 \xi^2/(1-\xi^4)^2$ is elementary, the bound $\xi \leq e^{-C(\alpha) n_{k,0} \delta_k^2} \leq e^{-1}$ holds because $C(\alpha) \leq D(\alpha-\alpha^2)/8$ and $C(\alpha) n_{k,0} \delta_k^2 \geq 1$, and the last inequality uses the assumption $3\kappa 4/(1-e^{-4})^2 \leq 4^{-1}$.

\emph{First term.} On the set $A_{\bar n_k}$ we have, pathwise, $\left[ \int \tilde \pi_k(\theta_k) e^{-\alpha \ell_{\bar n_k}(\theta_k,\theta_{0,k})} d\theta_k \right]^{-1/\alpha} \geq 4\, e^{ D \frac{1-\alpha}{8} \bar n_k \delta_k^2 }$, and hence, using Jensen's inequality for the convex map $x \mapsto x^{-1/\alpha}$,
\begin{align}
    \ind\big(A_{\bar n_k}\big)\, e^{C(\alpha) \bar n_k \delta_k^2} \leq 4^{-1} \int \tilde \pi_k(\theta_k)\, e^{\ell_{\bar n_k}(\theta_k,\theta_{0,k})}\, e^{ -\left(D \frac{1-\alpha}{8} - C(\alpha)\right) \bar n_k \delta_k^2 }\, d\theta_k.
    \label{eq:MABDE8}
\end{align}
For fixed $\theta_k$, $e^{\ell_{\bar n_k}(\theta_k,\theta_{0,k})} = \prod_{i=1}^{\bar n_k} g_2(x_{i,k})$ with $g_2(x) = p(x|\theta_{0,k})/p(x|\theta_k)$ and $m_{g_2}(\theta_k) = e^{D_2 \left( {\theta_{0,k}}\| {\theta_k} \right)}$. On the support of $\tilde \pi_k$ we have $D_2 \left( {\theta_{0,k}}\| {\theta_k} \right) \leq D \frac{(\alpha-\alpha^2)}{8} \delta_k^2$ by the definition of the set $B\left(\theta_{0,k},\delta_k \right)$, so with $\rho = e^{-\left(D \frac{1-\alpha}{8} - C(\alpha)\right)\delta_k^2}$ and $C(\alpha) \leq D\frac{(1-\alpha)^2}{16}$ we get $\rho\, m_{g_2}(\theta_k) \leq e^{-D \frac{(1-\alpha)^2}{16}\delta_k^2} \leq 1$. Applying Lemma~\ref{lem:MG} (with $g = g_2$ and $\Lambda = n_{k,0}$) and Fubini--Tonelli once more,
\begin{align}
    \bbE\left[\ind\big(A_{\bar n_k}\big)\, e^{C(\alpha) \bar n_k \delta_k^2}\right] \leq 4^{-1} \big(\rho\, m_{g_2}(\theta_k)\big)^{n_{k,0}} \leq 4^{-1}.
    \label{eq:MABDE8b}
\end{align}
Combining~\eqref{eq:MABDE7} and~\eqref{eq:MABDE8b} in~\eqref{eq:MABDE4},
\begin{align}
    \bbE\left[\Pi_{\alpha}(F_k|X_{\tau})\, e^{C(\alpha) \bar n_k \delta_k^2}\right]
    \leq 4^{-1} + 4^{-1} = 2^{-1},
    \label{eq:MABDE9}
\end{align}
which, together with~\eqref{eq:MABDE2}, proves~\eqref{eq:postcon}. The indicator form~\eqref{eq:postconL} holds under its own hypothesis $C(\alpha) \Lambda \delta_k^2 \geq 1$ alone, without requiring $C(\alpha) n_{k,0} \delta_k^2 \geq 1$: the identical argument applies with the compensation factor $e^{C(\alpha)\bar n_k \delta_k^2}$ replaced throughout by the indicator $\ind\{\bar n_k \geq \Lambda\}$ (so the choices of $\rho$ lose their $C(\alpha)$ terms) and with Lemma~\ref{lem:MG} invoked at threshold $\Lambda$ in place of $n_{k,0}$; every estimate then holds with $\Lambda$ in place of $n_{k,0}$, and the series bound uses $\xi = e^{-D\frac{(\alpha-\alpha^2)}{8} \Lambda \delta_k^2} \leq e^{-C(\alpha) \Lambda \delta_k^2} \leq e^{-1}$, yielding $\bbE[\Pi_{\alpha}(F_k|X_{\tau})\, \ind\{\bar n_k \geq \Lambda\}] \leq 2^{-1} e^{-2C(\alpha) \Lambda \delta_k^2} \leq 2^{-1} e^{-C(\alpha) \Lambda \delta_k^2}$.
\end{proof}

\begin{proof}[Proof of Lemma~\ref{lem:Int1}]
Observe that 
\begin{align}
     P( i_t=k,\mu_k\leq y_k  | X_{t-1}) = P(i_t=k | \mu_k\leq y_k , X_{t-1} ) P(\mu_k\leq y_k| X_{t-1}).
\end{align}
Now using the fact that on the set $\{\mu_k\leq y_k\}$, $i_t=k$ for some $k\neq i^*$, only if $\mu_j \leq y_k$ for all $j\in[K]$, we have for any $k\neq i^*$
\begin{align}
\nonumber
P(i_t=k | \mu_k\leq y_k , X_{t-1} ) &\leq P(\forall j \in [K], \mu_j \leq y_k  | \mu_k\leq y_k , X_{t-1} )
\\
\nonumber
&= P(\mu_{i^*} \leq y_k  | \mu_k\leq y_k , X_{t-1} ) P(\forall j \neq i^*, \mu_j \leq y_k  | \mu_k\leq y_k , X_{t-1} )
\\
\nonumber
&= \pi_{\alpha}(\mu_{i^*} \leq y_k  | X_{t-1} ) P(\forall j \neq i^*, \mu_j \leq y_k  | \mu_k\leq y_k , X_{t-1} )
\\
&= ( 1-p_{k,t} ) P(\forall j \neq i^*, \mu_j \leq y_k  | \mu_k\leq y_k , X_{t-1} ),
\label{eq:MA2}
\end{align}
where the penultimate inequality follows due to conditional independence of $\mu_k$ and $\mu_{i^*}$. Similarly, we have 
\begin{align}
\nonumber
P(i_t=i^* | \mu_k\leq y_k , X_{t-1} ) &\geq P(\forall j \neq i^*, \mu_i^* > y_k \geq \mu_j  | \mu_k\leq y_k , X_{t-1} )
\\
\nonumber
&= P(\mu_{i^*} > y_k  | \mu_k\leq y_k , X_{t-1} ) P(\forall j \neq i^*, \mu_j \leq y_k  | \mu_k\leq y_k , X_{t-1} )
\\
\nonumber
&= \pi_{\alpha}(\mu_{i^*} > y_k  | X_{t-1} ) P(\forall j \neq i^*, \mu_j \leq y_k  | \mu_k\leq y_k , X_{t-1} )
\\
&= p_{k,t} P(\forall j \neq i^*, \mu_j \leq y_k  | \mu_k\leq y_k , X_{t-1} ).
\label{eq:MA2.5}
\end{align}
Combining~\eqref{eq:MA2} and~\eqref{eq:MA2.5} implies that for any $k\neq i^*$
\begin{align*}
P(i_t=k , \mu_k\leq y_k | X_{t-1} ) &= P(i_t=k | \mu_k\leq y_k , X_{t-1} )P(\mu_k\leq y_k| X_{t-1})  
\\
&\leq  \frac{1-p_{k,t}}{p_{k,t}} P(i_t=i^* , \mu_k\leq y_k | X_{t-1} ).
\end{align*}
\end{proof}

\begin{proof}[Proof of Lemma~\ref{lem:Int2}]
For brevity, we omit $i^*$ from the subscript of $\mu_{i^*}$, as it is clear that proof in this lemma is for the best arm $i^*$. Now, let $G_u$ denote a geometric random variable denoting the number of consecutive independent trials until $\mu > y_k$ (We explicitly write $\mu(u)$ as $\mu$ to show its dependence on $u$, however, it is omitted in the note for brevity). Then, observe that $p_{k,\tau_{u} +1} = \pi_{\alpha}(\mu > y_k|X_{\tau_{u}})$, and
\begin{align}
    \bbE\left[\frac{1-p_{k,\tau_{u}+1}}{p_{k,\tau_{u}+1}}\right] = \bbE\left[E[G_u|X_{\tau_{u}}]\right] = \sum_{r=1}^{\infty} \bbE\left[\pi_{\alpha}(G_u \geq r|X_{\tau_{u}})\right].
    \label{eq:G1}
\end{align}

Denoting $\mu^{m}$ as the $m^{th}$ i.i.d sample of $\mu|X_{\tau_{u}}$, observe that 
\begin{align}
    \nonumber
    \bbE\left[\pi_{\alpha}(G_u < r|X_{\tau_{u}})\right] \geq \bbE\left[\pi_{\alpha}(\max_{m=1}^{r}\mu^{m} > y_k |X_{\tau_{u}})\right] &= 1- \bbE\left[\pi_{\alpha}(\max_{m=1}^{r}\mu^{m} \leq y_k |X_{\tau_{u}}) \right]
    \\
    &= 1-\bbE\left[\left(\pi_{\alpha}(\mu \leq y_k |X_{\tau_{u}})\right)^{r}\right]
    \nonumber
    \\
    &= 1-\bbE\left[\left(1-p_{k,\tau_u+1}\right)^{r}\right],
    \label{eq:G2}
\end{align}
Fix $\delta_k=\frac{\mu_{0,i^*}- \mu_{0,k}}{3}$. Now setting $y_k=\mu_{0,i^*}-\delta_k$, and using Lemma~\ref{lem:Post}, we have for any $u > L_k(T)$ (here the indicator form~\eqref{eq:postconL} of Lemma~\ref{lem:Post} is applied at the bounded stopping time $\tau_u \wedge T$ with $\Lambda = u$; the indicator equals one because $\bar n_{i^*}(\tau_u) = u$ by the definition of $\tau_u$, and its admissibility condition holds because $C(\alpha)\delta_k^2 u > C(\alpha)\delta_k^2 L_k(T) = \max\{\log(MT),\,1\} \geq 1$, which requires no condition on the warm-up size),  
\begin{align*}
    \bbE \left[   \frac{1-p_{k,\tau_{u}+1}}{p_{k,\tau_{u}+1}}     \right] \leq e^{- C(\alpha)u\delta_k^2} \sum_{r=1}^{\infty} 2^{-r} =  e^{- C(\alpha)u\delta_k^2} \leq \frac{1}{MT},
\end{align*}
where the last step uses $C(\alpha)\delta_k^2 u > \max\{\log(MT),\,1\} \geq \log(MT)$. This establishes the second assertion.

Note that we still need to bound $\bbE \left[   \frac{1-p_{k,\tau_{u}+1}}{p_{k,\tau_{u}+1}}     \right]$ for all $u < \frac{\log M u}{C(\alpha)\delta_k^2}$. Since $\delta_k^2$ can be arbitrarily small, this implies that we need to bound $\bbE \left[   \frac{1-p_{k,\tau_{u}+1}}{p_{k,\tau_{u}+1}}     \right]$ for all $u \geq 1$.

In view of~\eqref{eq:G1} and~\eqref{eq:G2}, it suffices to bound $\bbE\left[\left(1-p_{k,\tau_u+1}\right)^{r}\right]$ for each $r \geq 1$. For $r < r_0$, with $r_0$ defined below, we use the trivial bound $\bbE\left[\left(1-p_{k,\tau_u+1}\right)^{r}\right] \leq 1$; for $r \geq r_0$ we combine a pathwise lower bound on $p_{k,\tau_u+1}$, valid on a high-probability event, with the trivial bound on the complement of that event.

Assuming $\mu=f(\theta)$, where $f$ is an (strictly) increasing and continuous function (which is true for exponential family and sub-Gaussian models considered in this manuscript), we have for $\theta_{u}^*$ defined in Lemma~\ref{lem:FBVM} that,
\begin{align}
    \pi_{\alpha}(\mu\geq \mu_0|X_{\tau_{u}}) = \pi_{\alpha}(\sD_0(\theta-\theta_{u}^*) \geq \sD_0 (\theta_0-\theta_{u}^*)|X_{\tau_{u}}) \geq \pi_{\alpha}(\sD_0(\theta-\theta_{u}^*) \geq \sD_0 |\theta_0-\theta_{u}^*||X_{\tau_{u}}).
\end{align}  
Now using the second assertion of the Lemma~\ref{lem:FBVM} combined with the observation above, we have with $P_0$-probability of at least $1-e^{-\eta}$ for any $\eta>0$ that 
\begin{align}
    \pi_{\alpha}(\mu\geq \mu_0|X_{\tau_{u}}) \geq \pi_{\alpha}(\sD_0(\theta-\theta_{u}^*) \geq C\sqrt{1+\eta} |X_{\tau_{u}}) .
\end{align}
Next, using the first assertion of the Lemma~\ref{lem:FBVM} (for $p=1$) combined with the observation above, we have with $P_0$-probability of at least $1-2 e^{-\eta}$, 
\begin{align}
\nonumber
    \pi_{\alpha}&(\mu\geq \mu_0|X_{\tau_{u}}) 
    \geq \pi_{\alpha}(\sD_0(\theta-\theta_{u}^*) \geq C\sqrt{1+\eta} |X_{\tau_{u}})  \\
    \nonumber
    &
    \geq \mathbb{C}\exp\left(-2\sqrt{\frac{1+\eta}{u}} - 8e^{-\eta}\right) \left\{P(\phi/\sqrt{\alpha} > C\sqrt{1+\eta} ) \right\}-e^{-\eta}
    \\
    &\geq 
     \frac{\mathbb{C}\exp\left(-2\sqrt{\frac{1+\eta}{u}} - 8e^{-\eta}\right) }{\sqrt{2\pi}}\left\{ \frac{C\sqrt{\alpha (1+\eta)}}{1+C^2{\alpha (1+\eta)}}e^{-C^2{\alpha(1+ \eta)}/2}  \right\}-e^{-\eta}
   ,
\end{align}
where in the last inequality, we used the Gaussian lower tail bound (Mill's ratio).

Fix $r \geq 2$, fix a constant $a > 1$, and set $\eta = a\log r$; let $\mathbf{E}_r$ denote the intersection of the two events above for this choice of $\eta$, so that $P_0(\mathbf{E}_r^C) \leq 2e^{-\eta} = 2r^{-a}$ and the display above holds pathwise on $\mathbf{E}_r$. Using $u \geq n_0$ to lower bound the first exponential factor, define
\begin{align}
    \psi_r := \frac{\mathbb{C}\exp\left(-2\sqrt{\frac{1+a\log r}{n_0}} - 8r^{-a}\right) }{\sqrt{2\pi}}\left\{ \frac{C\sqrt{\alpha (1+a\log r)}}{1+C^2{\alpha (1+a\log r)}}e^{-C^2{\alpha(1+a \log r)/2}}  \right\}-r^{-a}.
    \label{eq:G4}
\end{align}
Then, pathwise on $\mathbf{E}_r$, for every $u \geq n_0$,
\begin{align}
    p_{k,\tau_u+1} = \pi_{\alpha}(\mu > y_k|X_{\tau_u}) \geq \pi_{\alpha}(\mu \geq \mu_0|X_{\tau_u}) \geq \psi_r,
    \nonumber
\end{align}
where the first inequality holds because $y_k < \mu_0$. Consequently, splitting the expectation on $\mathbf{E}_r$, bounding $\left(1-p_{k,\tau_u+1}\right)^{r}$ by one on $\mathbf{E}_r^C$, and using $(1-x)^r \leq e^{-rx}$ for $x \leq 1$,
\begin{align}
    \bbE\left[\left(1-p_{k,\tau_{u}+1}\right)^{r}\right] \leq \left(1-\psi_r\right)^{r} + P_0(\mathbf{E}_r^C) \leq e^{-r \psi_r} + 2r^{-a}.
    \label{eq:G5}
\end{align}
Now let $r_0 = r_0(C,\alpha,n_0,\mathbb{C},a) \geq 2$ be the smallest integer such that $e^{-r\psi_r} \leq r^{-a}$ holds for all $r \geq r_0$ and, in addition, $\sum_{r \geq r_0} 3r^{-a} \leq 2$ (such a finite integer exists under the assumed condition $C^2\alpha < 2$); as before, the middle factor of $\psi_r$ may be simplified using $x/(1+x^2) \geq 1/(2x)$ for $x \geq 1$. Then, for all $u \geq n_0$ and all $r \geq r_0$,
\begin{align}
    \bbE\left[\left(1-p_{k,\tau_{u}+1}\right)^{r}\right] \leq 3r^{-a}.
    \label{eq:G6}
\end{align}

Finally, bounding the terms with $r < r_0$ trivially by one, it follows from~\eqref{eq:G1},~\eqref{eq:G2}, and~\eqref{eq:G6} that
\begin{align}
    \bbE\left[\frac{1-p_{k,\tau_{u}+1}}{p_{k,\tau_{u}+1}}\right] \leq \left(r_0(C,\alpha,n_0,\mathbb{C},a) - 1\right) + \sum_{r=r_0}^{\infty} 3r^{-a} \leq r_0(C,\alpha,n_0,\mathbb{C},a) +2,
\end{align}
where the last inequality uses the defining property $\sum_{r \geq r_0} 3r^{-a} \leq 2$ of $r_0$.
\end{proof}

Here $C$ denotes the constant of the score-norm deviation bound in the second assertion of Lemma~\ref{lem:FBVM}. Whenever the normalized score is sub-Gaussian with parameter $\nu$, the constant of condition $(e\!d_0)$ in Assumption~\ref{ass:Spokoiny}, a Chernoff bound gives this deviation bound with $C^2 = 2\nu^2$; here $\nu \geq 1$, while the universal constant of~\citet{Panov2015} corresponds to the conservative choice $\nu^2 = 3$. This covers the sub-Gaussian family and every exponential family with a bounded sufficient statistic (including the Bernoulli and categorical reward models of our examples); for exponential families with an unbounded sufficient statistic the score is sub-exponential, and the bound holds in the moderate-deviation range that is all the argument uses. For the Gaussian reward model $p_\eta = N(0,1)$, condition $(e\!d_0)$ holds with $\nu = 1$ exactly. Indeed, $h_\eta(z) = -z^2/2$ gives $h'_\eta(\eta) = -\eta$ and $\mathtt{h}^2 = -\bbE[h''_\eta(\eta)] = 1$, so the normalized score is $h'_\eta(\eta)/\mathtt{h} = -\eta \sim N(0,1)$ and $\log \bbE \exp(\mu\, h'_\eta(\eta)/\mathtt{h}) = \mu^2/2$; that is, Assumption~\ref{ass:AddSG}(1) holds with $v_0 = 1$, and Proposition~\ref{prop:SpokSG} then gives $(e\!d_0)$ with $\nu = v_0 = 1$. 

\begin{proof}[Proof of Lemma~\ref{lem:Int4}]
Since, $y_k>u_k$, therefore,
\begin{align}
    \bbE\left[  \ind(i_t=k )  \ind\{\mu_k > y_k  \}   \right] \leq \bbE\left[  \ind(i_t=k )  \ind\{\mu_k > u_k  \}   \right] 
\end{align}

Now for $u_k = \mu_{0,k}+\delta_k$, where $\delta_k=\Delta_k/3$, we have
\begin{align}
    \nonumber
     \sum_{t=1}^{T}&\bbE\left[  \ind(i_t=k )  \ind\{\mu_k > u_k  \}   \right]  
     \\
     \nonumber
     &= \sum_{t=1}^{T} \bbE\left[  \ind(i_t=k )  \ind\{\mu_k > \mu_{0,k}+\delta_k  \}   \right] 
     \\
     \nonumber
     &\le \sum_{t=1}^{T} \bbE\left[  \ind(i_t=k )  \ind\{|\mu_k - \mu_{0,k}|^2>\delta_k^2  \}   \right]
     \\
     \nonumber
     &=\sum_{t=1}^{T} \bbE\left[  \ind(i_t=k )  \ind\{|\mu_k - \mu_{0,k}|^2>\delta_k^2  \}  \ind\{\bar n_k(t) \leq L_k(T)\} \right] 
     \\
     \nonumber
      &\quad +\sum_{t=1}^{T} \bbE\left[  \ind(i_t=k )  \ind\{|\mu_k - \mu_{0,k}|^2>\delta_k^2  \}  \ind\{\bar n_k(t) > L_k(T)\} \right]
     \\
     \nonumber
     &\leq  \bbE\left[  \sum_{t=1}^{T} \ind(i_t=k )    \ind\{\bar n_k(t) \leq L_k(T)\} \right] +\sum_{t=1}^{T} \bbE\left[  \ind(i_t=k )  \ind\{|\mu_k - \mu_{0,k}|^2>\delta_k^2  \}  \ind\{\bar n_k(t) > L_k(T)\} \right]
     \\
     &\leq  L_k(T) +\sum_{t=1}^{T} \bbE\left[   \ind\{|\mu_k - \mu_{0,k}|^2>\delta_k^2  \}  \ind\{\bar n_k(t) > L_k(T)\} \right],
     \label{eq:MA4}
\end{align}
where the second inequality follows because $\bbE\left[ \sum_{t=1}^{T} \ind(i_t=k )  \ind\{\bar n_k(t) \leq L_k(T)\} \right]$ is trivially bounded by $L_k(T)$, since on any sample path of observation, if for any $t_1\in[T]$, $\bar n_k(t_1)> L_k(T))$, then for all $t\geq t_1$, $\bar n_k(t)>L_k(T)$. Therefore, on those sample paths $\sum_{t=t_1}^{T} \ind(i_t=k)  \ind\{\bar n_k(t) \leq L_k(T)\}=0$, and thus $\sum_{t=1}^{T} \ind(i_t=k )  \ind\{\bar n_k(t) \leq L_k(T)\}$ must be less than $L_k(T)$.
Now using the indicator form~\eqref{eq:postconL} of Lemma~\ref{lem:Post}, applied at the deterministic time $t$ with $\Lambda = L_k(T)$ (admissible since $C(\alpha)\delta_k^2 L_k(T) = \max\{\log(MT),\,1\} \geq 1$ by the definition of $L_k(T)$, with no further hypothesis; a constant time is trivially a bounded stopping time, since $\{t \leq s\}$ is either empty or the whole sample space, both of which lie in $\mathcal{F}_s$), the summand in the second term of~\eqref{eq:MA4} satisfies
\begin{align}
    \nonumber
    &\bbE\left[   \ind\{|\mu_k - \mu_{0,k}|^2>\delta_k^2  \}  \ind\{\bar n_k(t) > L_k(T)\} \right] \nonumber
    \\
    &\quad\leq \bbE\left[  \bbE\left[ \ind\{|\mu_k - \mu_{0,k}|^2\geq  \delta_k^2\} | X_{t} \right] \ind\{\bar n_k(t) \geq L_k(T)\}\right]
    \\
    &\quad\leq 2^{-1} e^{ - C(\alpha) L_k(T)\delta_k^2 } = 2^{-1} e^{-\max\{\log(MT),\,1\}} \leq \frac{2^{-1}}{MT}. \nonumber
    \end{align}
 Therefore, it follows from~\eqref{eq:MA4} and the equation above that 
\begin{align*}
    \sum_{t=1}^{T}\bbE\left[  \ind(i_t=k )  \ind\{\mu_k \geq y_k  \}   \right]  &\leq \sum_{t=1}^{T}\bbE\left[  \ind(i_t=k )  \ind\{\mu_k > u_k  \}   \right] 
    \\
    &\leq L_k(T)  + \frac{1}{2M} = \frac{\max\{\log(MT),\,1\}}{C(\alpha)\delta_k^2}  + \frac{1}{2M}.
\end{align*}
\end{proof}

\begin{proof}[Proof of Theorem~\ref{thm:PIRB0}]
    For $\Delta_k = \mu^0_{i^*} -  \mu_{0,k}$, note that
    \begin{align}
        \bbE  \left[ \sum_{t=1}^{T}\mu^0_{i^*} -  \mu^0_{i_t}  \right] 
        &=  \sum_{\forall k \neq i^*}  \Delta_k \bbE\left[\sum_{t=1}^{T}\ind( i_t=k )\right] 
        \label{eq:MA0}
    \end{align}
    
    In what follows, we analyze $\sum_{t=1}^{T}\bbE\left[ \ind( i_t=k ) \right]$. Observe that
    \begin{align} 
        \sum_{t=1}^{T} \bbE\left[  \ind(i_t=k )\right] &=  \sum_{t=1}^{T} \bbE\left[  \ind(i_t=k )  \ind\{\mu_k \leq y_k  \}   \right] + \sum_{t=1}^{T} \bbE\left[  \ind(i_t=k )  \ind\{\mu_k > y_k  \}   \right],
        \label{eq:MA1}
        \end{align}
        where for any $k\neq i^*$, we assume that there exists $\{u_k,y_k\}$ such that $\mu_{0,k} < u_k < y_k < \mu^0_{i^*}$. Now for any $k \neq i^*$ recall the definition of $p_{k,t}:= \pi_{\alpha}( \mu_{i^*} \geq y_k | X_{t-1} )$.

    Using Lemma~\ref{lem:Int1}, the first term in~\eqref{eq:MA1} can be bounded as
    \begin{align}
        \nonumber
      \sum_{t=1}^{T} \bbE \left[ E\left[  \ind(i_t=k )  \ind\{\mu_k \leq y_k  \}  | X_{t-1}  \right] \right] &\leq \sum_{t=1}^{T} \bbE \left[ E\left[  \frac{1-p_{k,t}}{p_{k,t}} \ind(i_t=i^* )  \ind\{\mu_k \leq y_k  \}  | X_{t-1}  \right] \right]
      \\
      &\leq \sum_{t=1}^{T} \bbE \left[   \frac{1-p_{k,t}}{p_{k,t}} \ind(i_t=i^* )   \right].
    \end{align}
    Recall that $\tau_u$ denotes the round at which the total number of observations from the best arm $i^*$, counting the $n_{i^*,0}$ warm-up samples together with the pulls made within the horizon, first equals $u$; equivalently $\bar n_{i^*}(\tau_u) = u$, with the convention $\tau_{n_{i^*,0}} = 0$. Note that $\tau_{n_{i^*,0}+T} \geq T$. Therefore, it follows from the inequality above that
    \begin{align}
        \nonumber
      \sum_{t=1}^{T} \bbE \left[ E\left[  \ind(i_t=k )  \ind\{\mu_k \leq y_k  \}  | X_{t-1}  \right] \right] &\leq \sum_{t=1}^{\tau_{n_{i^*,0}+T}} \bbE \left[   \frac{1-p_{k,t}}{p_{k,t}} \ind(i_t=i^* )   \right]
      \\
      \nonumber
      &= \sum_{u=n_{i^*,0}}^{n_{i^*,0}+T-1} \sum_{t=\tau_{u}+1}^{\tau_{u+1}}  \bbE \left[   \frac{1-p_{k,t}}{p_{k,t}}  \ind(i_t=i^* )   \right]
      \\ 
      &= \sum_{u=n_{i^*,0}}^{n_{i^*,0}+T-1}  \bbE \left[   \frac{1-p_{k,\tau_{u}+1}}{p_{k,\tau_{u}+1}}  \sum_{t=\tau_{u}+1}^{\tau_{u+1}}  \ind(i_t=i^* )   \right]
      \nonumber
      \\
      &= \sum_{u=n_{i^*,0}}^{n_{i^*,0}+T-1}  \bbE \left[   \frac{1-p_{k,\tau_{u}+1}}{p_{k,\tau_{u}+1}}     \right] ,
      \label{eq:MA3}
    \end{align}
    where the second equality follows since $p_{k,t}$ does not change (and is equals to $p_{k,\tau_{u}+1}$) in the time interval $[\tau_{u}+1,\tau_{u+1}]$, because the posterior of arm $i^*$ is updated only when that arm is pulled. Consequently, it follows from~\eqref{eq:MA3} and Lemma~\ref{lem:Int2} that
   \begin{align}
    \nonumber
    \sum_{t=1}^{T} \bbE \left[ E\left[  \ind(i_t=k )  \ind\{\mu_k \leq y_k  \}  | X_{t-1}  \right] \right] &\leq \sum_{u=n_{i^*,0}}^{n_{i^*,0}+T-1}  \bbE \left[   \frac{1-p_{k,\tau_{u}+1}}{p_{k,\tau_{u}+1}}     \right] 
    \\
    \nonumber
    &\leq \sum_{u=n_{i^*,0}}^{\lfloor L_k(T) \rfloor} \bbE \left[   \frac{1-p_{k,\tau_{u}+1}}{p_{k,\tau_{u}+1}}     \right]   + \sum_{u=\lfloor L_k(T) \rfloor + 1}^{n_{i^*,0}+T-1} \bbE \left[   \frac{1-p_{k,\tau_{u}+1}}{p_{k,\tau_{u}+1}}     \right]
    \\
    &\leq  r_0(L_k(T)+1) + \frac{1}{M},
    \label{eq:MA6} 
   \end{align}
   where the first sum has at most $L_k(T)+1$ terms and the second has at most $T$ terms, each bounded by $(MT)^{-1}$. Both counts hold for an arbitrary constant warm-up $n_{i^*,0}=n_0\geq1$: if $\lfloor L_k(T)\rfloor < n_0$ the first sum is empty and the second has exactly $T$ terms, while if $\lfloor L_k(T)\rfloor \geq n_0$ the first has $\lfloor L_k(T)\rfloor-n_0+1 \leq L_k(T)+1$ terms and the second has $n_0+T-1-\lfloor L_k(T)\rfloor \leq T$ terms. In particular no lower bound on $n_0$ is used anywhere. 

Now consider the second term in~\eqref{eq:MA1}. Using Lemma~\ref{lem:Int4}, the second term in~\eqref{eq:MA1} is bounded by
\begin{align}
    \sum_{t=1}^{T}\bbE\left[  \ind(i_t=k )  \ind\{\mu_k \geq y_k  \}   \right]  \leq \sum_{t=1}^{T}\bbE\left[  \ind(i_t=k )  \ind\{\mu_k > u_k  \}   \right]   \leq L_k(T)+\frac{1}{2M}.
    \label{eq:MA9}
\end{align}

The assertion of the theorem follows from substituting~\eqref{eq:MA6} and~\eqref{eq:MA9} into~\eqref{eq:MA1} and then into~\eqref{eq:MA0} for $M=C(\alpha)\delta_k^2$ .
In particular, we have
 \begin{align}
     \nonumber
    \Delta_k  &\bbE[n_k(T)] 
    \\
    \nonumber
    &\leq   \Delta_k  \left(r_0 \frac{\max\{\log(TC(\alpha)\delta_k^2),\,1\}}{C(\alpha)\delta_k^2} + r_0 + \frac{1}{C(\alpha)\delta_k^2} + \frac{\max\{\log(TC(\alpha)\delta_k^2),\,1\}}{C(\alpha)\delta_k^2}  + \frac{1}{2C(\alpha)\delta_k^2} \right)
    \\
    \nonumber
    &=    \left(\frac{9(r_0+1) \max\{\log(T /9 C(\alpha)\Delta_k^2),\,1\}}{C(\alpha)\Delta_k} + r_0 \Delta_k + \frac{27}{2C(\alpha) \Delta_k}  \right).
\end{align}
The upper bound above decreases for  $\Delta_k > \frac{e}{\sqrt{TC(\alpha)}}$, therefore, the result follows by dividing the summand over $\Delta_k > \frac{e\sqrt{K\log(K)}}{\sqrt{TC(\alpha)}}$ and its complement. The $\max$ with $1$ only enlarges the bound and does not affect this monotonicity: on the range $\Delta_k > \nicefrac{e}{\sqrt{TC(\alpha)}}$ the quantity $\log(\nicefrac{T C(\alpha)\Delta_k^2}{9})$ is increasing in $\Delta_k$, so the $\max$ is attained by the constant $1$ on an initial sub-interval and by the logarithm thereafter, and the displayed bound is decreasing in $\Delta_k$ on each piece.
\end{proof}

\begin{proof}[Proof of Theorem~\ref{thm:PIRB}]
First, using the assumption that $\Delta_k\leq 1$, observe that the instance dependent upper-bound in~Theorem~\ref{thm:PIRB0}
\begin{align} 
\nonumber
&\frac{9(r_0+1)\max\{\log(\frac{T}{9} C(\alpha)\Delta_k^2),\,1\}}{C(\alpha)\Delta_k} + r_0 \Delta_k + \frac{27}{2 C(\alpha) \Delta_k}
\\
    &\leq  \frac{9(r_0+1) \max\{\log(T C(\alpha)\Delta_k^2),\,1\}}{C(\alpha)\Delta_k} + r_0 + \frac{27}{2 C(\alpha) \Delta_k}.
    \end{align}
The upper bound above decreases for  $\Delta_k > \frac{e}{\sqrt{TC(\alpha)}}$, therefore, for $\Delta_k > \frac{e\sqrt{K\log(K)}}{\sqrt{TC(\alpha)}}$
, the upper bound above is bounded by (the $\max$ with $1$ is inactive at this threshold, since there $T C(\alpha)\Delta_k^2 \geq e^2 K\log K \geq e^2 \log 2 > e$ for every $K \geq 2$, hence $\log(e^2K\log K) > 1$) 
\begin{align}
    \nonumber
   & \left(\frac{9(r_0+1) \max\{\log(T C(\alpha)\Delta_k^2),\,1\}}{C(\alpha)\Delta_k} + r_0 + \frac{27}{2 C(\alpha) \Delta_k}  \right)
    \\
    \nonumber
    &\leq  \left(\frac{9(r_0+1) \max\{\log(e^2 K \log(K)),\,1\}}{C(\alpha)} \frac{\sqrt{TC(\alpha)}} {e\sqrt{K\log(K)}} + r_0 + \frac{27}{2 C(\alpha) } \frac{\sqrt{TC(\alpha)}} {e\sqrt{K\log(K)}} \right)
    \\
    \nonumber
    &\leq \left(\frac{18(r_0+1)}{e}+\frac{18(r_0+1)+27/2}{ e}   \right) \frac{\sqrt{T\log(K)}}{\sqrt{C(\alpha)K}} + r_0 
    \\
    & \leq 19(r_0+1) \frac{\sqrt{T\log(K)}}{\sqrt{C(\alpha)K}} + r_0.
    \label{eq:MA11}
\end{align}

Moreover, since the total number of pulls satisfies $\sum_{k \neq i^*} \bbE[n_k(T)] \leq T$, the arms with $\Delta_k \leq  \frac{e\sqrt{K\log(K)}}{\sqrt{TC(\alpha)}}$ jointly contribute
\begin{align}
  \sum_{k \neq i^*} \bbI\left(\Delta_k \leq \tfrac{e\sqrt{K\log(K)}}{\sqrt{TC(\alpha)}}\right) \Delta_k  \bbE[n_k(T)] \leq \frac{e\sqrt{K\log(K)}}{\sqrt{TC(\alpha)}} \sum_{k \neq i^*} \bbE[n_k(T)] \leq \frac{e\sqrt{T K\log(K)}}{\sqrt{C(\alpha)}}.
  \label{eq:MA12}
\end{align}

Now observe that for $M(T)=\frac{e\sqrt{K\log(K)}}{\sqrt{TC(\alpha)}}$, it follows from~\eqref{eq:MA11},~\eqref{eq:MA12}, and~\eqref{eq:MA0} that 
\begin{align}
    \nonumber
    \bbE  \left[ \sum_{t=1}^{T} (\mu^0_{i^*} -  \mu^0_{i_t})  \right] &= \sum_{\forall k \neq i^*} \Delta_k  \bbE [n_{k}(T)] 
    \\
    \nonumber
    &= \sum_{k\in[K] \setminus \{i^*\}} \left[ \bbI\left(\Delta_k \leq M(T) \right) \Delta_k \bbE [n_{k}(T)]  + \bbI\left(\Delta_k > M(T)\right) \Delta_k \bbE [n_{k}(T)] \right]
    \\
    \nonumber
    &\leq  \frac{e\sqrt{K\log(K)}}{\sqrt{TC(\alpha)}} \sum_{k\in[K] \setminus \{i^*\}}\bbE [n_{k}(T)]  + \sum_{k\in[K] \setminus \{i^*\}} \left[ 19(r_0+1) \frac{\sqrt{T\log(K)}}{\sqrt{C(\alpha)K}} + r_0  \right]
    \\
    &=  \mathcal{O}\left( K + \frac{\sqrt{KT\log(K)}}{\sqrt{C(\alpha)}} \right).
\end{align}

Also, for $\frac{K}{\log(K)}\leq \frac{T}{C(\alpha)}$, $\bbE  \left[ \sum_{t=1}^{T} (\mu^0_{i^*} -  \mu^0_{i_t})  \right] = \mathcal{O}\left(\frac{\sqrt{KT\log(K)}}{\sqrt{C(\alpha)}} \right) $.

\end{proof}

\begin{proof}[Proof of Theorem~\ref{thm:RB-PD}]
    For $\Delta_k = \mu^0_{i^*} -  \mu_{0,k}$, note that
    \begin{align}
        \bbE  \left[ \sum_{t=1}^{T}\mu^0_{i^*} -  \mu^0_{i_t}  \right] = \sum_{\forall k \neq i^*}  \Delta_k \bbE[n_{k}(T)] = \sum_{\forall k \neq i^*}  \Delta_k \bbE\left[\sum_{t=1}^{T}\ind( i_t=k )\right] = \sum_{\forall k \neq i^*}  \Delta_k  \sum_{t=1}^{T}\bbP\left[ i_t=k \right]
        \label{eq:MA20}
    \end{align}
    
    In what follows, we analyze $\sum_{t=1}^{T}\bbP\left[ i_t=k \right]$. Observe that
    \begin{align} 
        \sum_{t=1}^{T} \bbE\left[  \ind(i_t=k )\right] &=  \sum_{t=1}^{T} \bbE\left[  \ind(i_t=k )  \ind\{\mu_k \leq y_k  \}   \right] + \sum_{t=1}^{T} \bbE\left[  \ind(i_t=k )  \ind\{\mu_k > y_k  \}   \right],
        \label{eq:MA21}
        \end{align}
        where for any $k\neq i^*$, we assume that there exists $\{u_k,y_k\}$ such that $\mu_{0,k} < u_k < y_k < \mu^0_{i^*}$. Now for any $k \neq i^*$ define $p_{k,t}:= \pi_{\alpha}( \mu_{i^*} \geq y_k | X_{t-1} )$.

    Therefore, using Lemma~\ref{lem:Int1}, the first term in~\eqref{eq:MA21} can be bounded as
    \begin{align}
        \nonumber
      \sum_{t=1}^{T} \bbE \left[ E\left[  \ind(i_t=k )  \ind\{\mu_k \leq y_k  \}  | X_{t-1}  \right] \right] &\leq \sum_{t=1}^{T} \bbE \left[ E\left[  \frac{1-p_{k,t}}{p_{k,t}} \ind(i_t=i^* )  \ind\{\mu_k \leq y_k  \}  | X_{t-1}  \right] \right]
      \\
      &\leq \sum_{t=1}^{T} \bbE \left[   \frac{1-p_{k,t}}{p_{k,t}} \ind(i_t=i^* )   \right].
    \end{align}
    Recall that $\tau_u$ is the round at which the total number of observations from the best arm $i^*$, including the $n_{i^*,0}$ warm-up samples, first equals $u$, so that $\bar n_{i^*}(\tau_u) = u$, with the convention $\tau_{n_{i^*,0}} = 0$. Note that $\tau_{n_{i^*,0}+T} \geq T$. Therefore, it follows from the inequality above that
    \begin{align}
        \nonumber
      \sum_{t=1}^{T} \bbE \left[ E\left[  \ind(i_t=k )  \ind\{\mu_k \leq y_k  \}  | X_{t-1}  \right] \right] &\leq \sum_{t=1}^{\tau_{n_{i^*,0}+T}} \bbE \left[   \frac{1-p_{k,t}}{p_{k,t}} \ind(i_t=i^* )   \right]
      \\
      \nonumber
      &= \sum_{u=n_{i^*,0}}^{n_{i^*,0}+T-1} \sum_{t=\tau_{u}+1}^{\tau_{u+1}}  \bbE \left[   \frac{1-p_{k,t}}{p_{k,t}}  \ind(i_t=i^* )   \right]
      \\
      \nonumber
      &= \sum_{u=n_{i^*,0}}^{n_{i^*,0}+T-1}  \bbE \left[   \frac{1-p_{k,\tau_{u}+1}}{p_{k,\tau_{u}+1}}  \sum_{t=\tau_{u}+1}^{\tau_{u+1}}  \ind(i_t=i^* )   \right]
      \\
      &= \sum_{u=n_{i^*,0}}^{n_{i^*,0}+T-1}  \bbE \left[   \frac{1-p_{k,\tau_{u}+1}}{p_{k,\tau_{u}+1}}     \right] ,
      \label{eq:MA23}
    \end{align}
    where the second equality follows since $p_{k,t}$ does not change (and is equals to $p_{k,\tau_{u}+1}$) in the time interval $[\tau_{u}+1,\tau_{u+1}]$. 

    Consequently, it follows from~\eqref{eq:MA23} and Lemma~\ref{lem:Int2} that
   \begin{align}
    \nonumber
    \sum_{t=1}^{T} \bbE \left[ E\left[  \ind(i_t=k )  \ind\{\mu_k \leq y_k  \}  | X_{t-1}  \right] \right] &\leq \sum_{u=n_{i^*,0}}^{n_{i^*,0}+T-1}  \bbE \left[   \frac{1-p_{k,\tau_{u}+1}}{p_{k,\tau_{u}+1}}     \right] 
    \\
    \nonumber
    &\leq \sum_{u=n_{i^*,0}}^{\lfloor L_k(T) \rfloor} \bbE \left[   \frac{1-p_{k,\tau_{u}+1}}{p_{k,\tau_{u}+1}}     \right]   + \sum_{u=\lfloor L_k(T) \rfloor + 1}^{n_{i^*,0}+T-1} \bbE \left[   \frac{1-p_{k,\tau_{u}+1}}{p_{k,\tau_{u}+1}}     \right]
    \\
    &\leq  r_0(L_k(T)+1) + 1.
    \label{eq:MA26} 
   \end{align} 
   As in the proof of Theorem~\ref{thm:PIRB0}, the two term counts hold for an arbitrary constant warm-up $n_{i^*,0}=n_0\geq1$, and no lower bound on $n_0$ is used.

Now consider the second term in~\eqref{eq:MA1}. It follows from Lemma~\ref{lem:Int4} (for $M=1$, for which $L_k(T)=\nicefrac{\max\{\log T,\,1\}}{(C(\alpha)\delta_k^2)}=\nicefrac{\log T}{(C(\alpha)\delta_k^2)}$ whenever $T \geq e$) that
\begin{align}
    \sum_{t=1}^{T}\bbE\left[  \ind(i_t=k )  \ind\{\mu_k \geq y_k  \}   \right]  \leq \sum_{t=1}^{T}\bbE\left[  \ind(i_t=k )  \ind\{\mu_k > u_k  \}   \right]   \leq \frac{\log(T)}{C(\alpha)\delta_k^2}  + \frac{1}{2}.
    \label{eq:MA29}
\end{align}

Consequently, the result follows from substituting~\eqref{eq:MA26} and~\eqref{eq:MA29} into~\eqref{eq:MA21} and then into~\eqref{eq:MA20} and noting that 
\begin{align}
    \Delta_k  \bbE[n_k(T)] &\leq  \Delta_k  \left(9(r_0+1)  \frac{\log(T)}{C(\alpha)\Delta_k^2}  + \frac{2r_0+3}{2} \right).
\end{align}
\end{proof}
\subsection{Exponential and sub-Gaussian Family satisfy Assumption~\ref{ass:prior2}}~\label{App:Assumption34}

\begin{proposition}~\label{prop:ExpG}
    The exponential family of reward distributions satisfying conditions in Definition~\ref{def:rewardExp} with priors having strictly positive, bounded  and continuous density on $\Theta$ satisfy Assumption~\ref{ass:prior2}  for $C(\alpha)\leq C_g$ and $\kappa= \frac{\max \pi_k(\theta) \sqrt{DC_g}}{\min_{B(\theta_{0,k},1)} \pi_k(\theta) \sqrt{C(\alpha)m}}$.
\end{proposition}

\begin{proof}[Proof of Proposition~\ref{prop:ExpG}]
    Recall the definition of $F_k^j$ and $B\left(\theta_{0,k},C(\alpha)\delta_k^2 \right)$ and observe that
    \begin{align}
    \nonumber
        \frac{\pi_k(F_k^j)}{\pi_k(B\left(\theta_{0,k},C(\alpha)\delta_k^2 \right))}&= \frac{\pi_k(D \frac{\alpha}{2}\delta_k^2 (j+1)\geq D_{\alpha}(\theta_k,\theta_{0,k}) \geq D \frac{\alpha}{2} \delta_k^2j)}{\pi_k(D_{2}(\theta_k,\theta_{0,k}) 
        \leq C(\alpha) \delta_k^2)} 
        \\
        \nonumber
        &\leq \frac{\pi_k(D_{\alpha}(\theta_k,\theta_{0,k}) \leq D \frac{\alpha}{2}(j+1) \delta_k^2)}{\pi_k(D_{2}(\theta_k,\theta_{0,k}) \leq C(\alpha) \delta_k^2)} 
        \\
        \nonumber
        &\leq \frac{\pi_k(\frac{\alpha m}{2}|\theta-\theta_{0,k}|^2 \leq D \frac{\alpha}{2} (j+1) \delta_k^2) }{\pi_k(C_g|\theta-\theta_{0,k}|^2 \leq C(\alpha) \delta_k^2)}
        \\
        &\leq   \frac{\max \pi_k(\theta)}{\min_{B(\theta_{0,k},1)} \pi_k(\theta)} \sqrt{\frac{D(j+1)/m}{C(\alpha)/C_g}},
        \end{align}
    where the second inequality follows because for $\alpha\in(0,1)$, $D_{\alpha}(\theta_k,\theta_{0,k})=\frac{1}{1-\alpha}\Big[\alpha A(\theta) + (1-\alpha)A(\theta_{0,k} ) - A(\alpha \theta + (1-\alpha)\theta_{0,k} ) \Big] \geq \frac{\alpha m}{2}|\theta-\theta_{0,k}|^2$ by Definition~\ref{def:rewardExp}~(iii) and $D_{2}(\theta_k,\theta_{0,k}) \leq C_g|\theta-\theta_{0,k}|^2$ due to second order mean value theorem and Definition~\ref{def:rewardExp}~(ii). The numerator of the third inequality follows because the prior is bounded, and the denominator follows because $\pi_k(C_g|\theta-\theta_{0,k}|^2 \leq C(\alpha) \delta_k^2) \geq  \min_{B(\theta_{0,k},\sqrt{C(\alpha)/C_g}\delta_k)} \pi_k(\theta) 2\sqrt{C(\alpha)/C_g} \delta_k \geq \min_{B(\theta_{0,k},1)} \pi_k(\theta) 2\sqrt{C(\alpha)/C_g} \delta_k$ since $\sqrt{C(\alpha)/C_g}\delta_k \leq 1$. Note that $\delta_k$ will cancel out in the last step. Therefore, the result follows for $\kappa= \frac{\max \pi_k(\theta) \sqrt{DC_g}}{\min_{B(\theta_{0,k},1)} \pi_k(\theta) \sqrt{C(\alpha)m}}$ and the fact that $\sqrt{j+1}\leq (j+1)$.
\end{proof}

Next, we provide a technical result that bounds 2-R\'enyi divergence between two random variables that are modeled with additive noise.

\begin{lemma}~\label{lem:QuadApp}
    For $\theta \in \Theta\subseteq \bbR$, let $X = g(\theta) + \eta$, where $g(\cdot)$ is a known twice differentiable mapping and $\eta$ is a random variable with density $p(\cdot)$. Then for any $\e>0$, there exist a $\theta^*(\epsilon)$ in $B(\theta_0,\e)$, a convex ball centred at $\theta_0$ with radius $\e$, such that for any $\theta\in B(\theta_0,\e)$,
    \begin{align}
        D_2(\theta,\theta_0) = (\theta-\theta_0)^2  I(\theta^*(\e)),
    \end{align}
    where 
   $I(\theta ) = \bbE_{\theta} \left[ ( \nabla_{u} \log p(x-g(u)) ) ( \nabla_{u}\log p(x-g(u)) )^\top \Big |_{u = \theta}   \right] $ and $I(\theta_0)$ is the Fisher information about $\theta_0$ that $X$ contains. 
 \end{lemma}
 \begin{proof}
     Recall from the definition of the $2-$R\'enyi divergence that, for a given $a \in \sA$,
     \begin{align}
         { D_2(\theta,\theta_0)} = \log   \int p_{\theta}(x)^2p_{\theta_0}(x)^{-1}dx   = \log \bbE_{\theta_0} \left[ \frac{p_{\theta}(x)^2}{p_{\theta_0}(x)^2} \right] = \log \bbE_{\theta_0} \left[ \frac{p(x-g(\theta))^2}{p(x-g(\theta_0))^2} \right].
     \end{align}
     For brevity, we denote $\dot{f}$ and $\ddot{f}$ as the first and second derivative of $f$ for $f=\{g,p\}$. It is straightforward to compute the gradient of $D_2$ with respect to $\theta$ as
     \begin{align}
         \nabla_{\theta} D_2 = - \left( \bbE_{\theta_0} \left[ \frac{p(x-g(\theta))^2}{p(x-g(\theta_0))^2} \right]\right)^{-1}   \bbE_{\theta_0} \left[  \frac{2 p(x-g(\theta))\dot{p}(x-g(\theta))\dot{g}(\theta)  }{p(x-g(\theta_0))^2}  \right]. 
     \end{align}
     Similarly, the Hessian of $D_2$ is computed as, 
     \begin{align}
         \nonumber
         \nabla_{\theta}^2 D_2 &=  -\left( \bbE_{\theta_0} \left[ \frac{p(x-g(\theta))^2}{p(x-g(\theta_0))^2} \right]\right)^{-2}   \left(\bbE_{\theta_0} \left[  \frac{2 p(x-g(\theta))\dot{p}(x-g(\theta))\dot{g}(\theta)  }{p(x-g(\theta_0))^2}  \right]\right)^2 
         \\
         \nonumber
         &+ \left( \bbE_{\theta_0} \left[ \frac{p(x-g(\theta))^2}{p(x-g(\theta_0))^2} \right] \right)^{-1} 
         \\
         &  \bbE_{\theta_0} \left[ \frac{
         \begin{matrix}
         2 p(x-g(\theta))\ddot{p}(x-g(\theta)) \dot{g}(\theta)^2 + 2 \left( \dot{p}(x-g(\theta)) \dot{g}(\theta) \right)^2 \\ 
         - 2 p(x-g(\theta))\dot{p}(x-g(\theta))\ddot{g}(\theta) 
         \end{matrix}}{p(x-g(\theta_0))^2}  \right].
     \end{align}
      When we evaluate the above Hessian at $\theta=\theta_0$ using the fact that $\nabla_{\theta}\bbE_{\theta_0}[\log p(x-g(\theta))] \Big|_{\theta=\theta_0} = 0$ and $\bbE_{\theta_0} \left[ \frac{2  \nabla_{\theta}^2 {p}(x-g(\theta))  }{p(x-g(\theta_0))} \right] = \nabla_{\theta}^2 \int p_{\theta}(x)dx = 0$, then 
     \begin{align}
         \nabla_{\theta}^2 D_2 \Big| _{\theta= \theta_0} &=   \bbE_{\theta_0} \left[  2  ( \nabla_{u} \log p(x-g(u)) ) ( \nabla_{u}\log p(x-g(u)) )^\top \Big |_{u = \theta_0}   \right] 
          =     2 I(\theta_0).
     \end{align}
     Now, using second-order Taylor's theorem, it follows that for any $\e>0$, there exists a $\theta^*(\epsilon)$ in $B(\theta_0,\e)$ such that for any $\theta\in B(\theta_0,\e)$,
     \begin{align}
         D_2(\theta,\theta_0) =  2 (\theta-\theta_0)^\top  I(\theta^*(\e))a(\theta-\theta_0).
     \end{align}
 \end{proof}

\begin{proposition}~\label{prop:SG}
    The sub-Gaussian family of reward distributions as defined in~Definition~\ref{def:SubReward} with priors having strictly positive, bounded  and continuous density on $\Theta$ satisfy Assumption~\ref{ass:prior2}  for $C(\alpha)\leq \tilde C$ and $\kappa= \frac{\max \pi_k(\theta) \sqrt{D\tilde{C}}}{\min_{B(\theta_{0,k},1)} \pi_k(\theta) \sqrt{C(\alpha)}}$, where $\tilde{C} = \max_{\theta\in B(\theta_{0,k},1)} I(\theta)$.
\end{proposition}
\begin{proof}[Proof of Proposition~\ref{prop:SG}]
    The proof follows similar steps to those used for the exponential family models. Observe that for $\alpha\in(0,1)$, $D_{\alpha}(\theta_k,\theta_{0,k}) \geq \frac{\alpha}{2}|\theta-\theta_{0,k}|^2$ by Lemma~\ref{lem:Pin}.
         Using Lemma~\ref{lem:QuadApp}, there exist a $\theta^*(1) \in B(\theta_{0,k},1)$ (defined later), a convex ball centred at $\theta_{0,k}$ with radius $1$, such that for any $\theta\in B( \theta_{0,k},1)$, $D_{2}(\theta, \theta_{0,k}) 
        \leq \tilde C  \|\theta-\theta_{0,k}\|^2$.
  In particular,  for $C(\alpha)\leq \tilde C$,  $\pi_k\left(D_{2}(\theta_k,\theta_{0,k}) 
        \leq C(\alpha) \delta_k^2\right) \geq \pi_k\left(B(\theta_{0,k},1),D_{2}(\theta_k,\theta_{0,k}) 
        \leq C(\alpha) \delta_k^2\right) \geq \pi_k\left(B(\theta_{0,k},1), \tilde C  \|\theta-\theta_0\|^2
        \leq C(\alpha) \delta_k^2\right) = \pi_k\left(\|\theta-\theta_0\|^2
        \leq C(\alpha) \delta_k^2/\tilde C \right) $. The result will follow using the above two observations for $\kappa= \frac{\max \pi_k(\theta) \sqrt{D\tilde C}}{\min_{B(\theta_{0,k},1)} \pi_k(\theta) \sqrt{C(\alpha)}}$.
\end{proof}

\subsection{Exponential and sub-Gaussian families satisfy Assumption~\ref{ass:Spokoiny}}~\label{app:Spokoiny}

To show that the exponential family satisfies Assumption~\ref{ass:Spokoiny}, the conditions required for the finite BvM to hold, we first write down various expressions used in their definition for exponential family models.
First recall the definition of the conditional log-likelihood of generating $X_n$, that is $\sL(\theta) = \sum_{i=1}^{n} \ell(x_i,\theta)= \sum_{i=1}^{n} (x_i\theta-A(\theta)+C(x_i))$.
Using this definition, the gradient of the log-likelihood can be derived as
\begin{align}
    \nabla \sL(\theta) = \sum_{i=1}^{n} \left({x_i    -  A'(  \theta) }\right).
\end{align}
Similarly, $\nabla^2 \sL(\theta) = -n A''(  \theta)$. Now observe that \[\bbE[\nabla \sL(\theta)] =  n\left(A'(  \theta_0)-A'(  \theta)\right).\]
and $\sD_0^{2}= n A''(\theta_0)$ and $\sD_0^{2}(\theta)
     = 
    - \nabla^{2} \bbE[ \sL(\theta)]  = n A''(\theta) $.
The stochastic part of the conditional log-likelihood is denoted as $\zeta (\theta):= \sL(\theta) - \bbE[\sL(\theta)]$ and $\nabla \zeta (\theta) = \nabla \sL(\theta) - \bbE[\nabla \sL(\theta)]= \sum_{i=1}^{n} \left({x_i    -  A'(\theta) }\right) - \bbE[ \sum_{i=1}^{n} \left({x_i    -  A'(  \theta) }\right)] = \sum_{i=1}^{n} \left(x_i  - \bbE[x_i]\right) = \sum_{i=1}^{n} \left(x_i  - A'(\theta_0)\right)=: \zeta_1 (\theta)$ and therefore $\bbE[\nabla \zeta(\theta_0)]=0$. Note that $\nabla \zeta(\theta)$ is independent of $\theta$, therefore $\nabla^2 \zeta(\theta)=0$. 
First, we present a technical lemma that is satisfied by exponential families.
\begin{lemma}~\label{lem:Aux1}
        There exists some constant $v_0>0$ and $\mathtt{g}_1>0$, for every $i$ a constant  $\sigma_i^2=A''( \theta_0)$  such that $\bbE[(x_i - A'(\theta_0))^2/\sigma_i^2]\leq 1$ and 
        \begin{align}
            \log \bbE [ \exp(\lambda (x_i - A'(\theta_0))/\sigma_i) ] \leq v_0^2 \lambda^2/2 , ~~|\lambda| < \mathtt{g}_1.
        \end{align}
    \end{lemma}
    \begin{proof}
        The proof is a direct consequence of Lemma 2.14~\cite{Spokoiny2012}(in their supplement).
    \end{proof}
Next, we have the main result that verifies all the conditions in Assumption~\ref{ass:Spokoiny}.
\begin{proposition}~\label{prop:SpokExp}
    The exponential family of distribution (Definition~\ref{def:rewardExp}) satisfies Assumption~\ref{ass:Spokoiny}.
\end{proposition}
\begin{proof}
    We derive all the conditions in seriatim. The proofs are adapted from~\cite{Panov2015}, and~\cite{Spokoiny2012}.
  \begin{enumerate}
    \item \({(e\!d_{0})} \):
    Observe using the definition of $\nabla \zeta_1(\theta_0)$ that 
        \begin{align}
            \bbE \exp\left\{
              \mathtt{m} \frac{\langle \nabla \zeta_1(\theta_0),\gamma \rangle}
              {\| \sD_0 \gamma \|}
              \right\} &= \bbE \exp\left\{
              \mathtt{m} \frac{  \left({x_i \gamma   -  A'(  \theta_0) \gamma}\right)}
              {\| \sV_0 \gamma \|}
              \right\} = \bbE \exp\left\{
              \mathtt{m} \frac{  \left({x_i   -  A'(  \theta_0) }\right)}
              {\sV_0 }
              \right\}
              \label{eq:ED0}
        \end{align}
    Now, the result follows immediately using Lemma~\ref{lem:Aux1} for $\lambda=\mathtt{m}$, $\mathtt{g}=\mathtt{g}_1$, and $\sigma_i=\sV_0$.
    \item \( {(e\!d_{1})} \):
      This assumption follows immediately for any $\omega>0$ because $\nabla^2\zeta(\theta) =0$ by definition. 
    \item Proof for \({(\ell_0)} \):
       For 1-d exponential family models, $-\bbE[\sL(\theta,\theta_0)]= \scKL(\theta_0,\theta) = A(\theta)-A(\theta_0) -A'(\theta_0)(\theta-\theta_0) = (\theta-\theta_0)^2I(\theta^*)/2$, where last inequality uses the second order Taylor expansion of $-\bbE[\sL(\theta,\theta_0)]$ and the fact that $\theta_0$ is the extreme point of $-\bbE[\sL(\theta_0,\theta)]$, and $\theta^*$ is a point between $\theta$ and $\theta_0$. To show $(\ell_0)$, note that $\left|\frac{\scKL(\theta_0,\theta)}{(\theta-\theta_0)^2I(\theta_0)} -1 \right|= \left|\frac{I(\theta^*)}{I(\theta_0)}-1\right|$ and $I(\cdot)$ is continuous, therefore there must exists a $\delta^*$ such that $\left|\frac{I(\theta^*)}{I(\theta_0)}-1\right|<\delta^* \mathtt{u}$.
      \item \( {(\ell{\mathtt{r}})} \): Using the expression for $\textsc{KL}(\theta,\theta_0)$ derived to prove $\ell_0$, we have $\frac{\textsc{KL}(\theta,\theta_0)}{\|\sV_0 (\theta - \theta_0)\|^2} = \frac{I(\theta^*)}{I(\theta_0)}$ for some $\theta^*$ between $\theta$ and $\theta_0$. Since, $\theta$ must satisfy $\|\sV_0(\theta-\theta_0)\|=u$,
        $\theta^*$ is a mapping from $\mathtt{u}$ and so is $\frac{I(\theta^*)}{I(\theta_0)}$. Therefore, we can always find a mapping $\mathtt{b}(\mathtt{u})$ such that $\frac{I(\theta^*)}{I(\theta_0)}\geq \mathtt{b}(\mathtt{u})$ for any $\mathtt{u}>0$. 
    \item \({(e\mathtt{u})} \): For any $\theta\in \Theta_0(\mathtt{u})$, observe that $\nabla \zeta_1(\theta) = \nabla \zeta_1(\theta_0)$. Therefore, the result follows using Lemma~\ref{lem:Aux1} using similar steps as we used for proving \({(e\!d_0)} \). It will follow for $\mathtt{g}_1(\mathtt{u})= \mathtt{g}_1$.
  \end{enumerate}
\end{proof}

To show that the sub-Gaussian family satisfies Assumption~\ref{ass:Spokoiny}, we first write down various expressions used in their definition for this family.
The log-likelihood of generating $X_n$ as $\sL(\theta) = \sum_{i=1}^{n}\log  \left[p_{\eta}(x_i-\theta)\right] $, where $p_{\eta}$ is the density of the sub-Gaussian error with sub-Gaussian parameter $1$. Also, denote $\log p_{\eta} := h_{\eta}$.
The stochastic part of the log-likelihood is denoted as $\zeta (\theta):= \sL(\theta) - \bbE[\sL(\theta)]$ and $\nabla \zeta (\theta) = \nabla \sL(\theta) - \bbE[\nabla \sL(\theta)]= \sum_{i=1}^{n} \left(-h_{\eta}'(x_i-\theta) +\bbE[ h_{\eta}'(x_i-\theta)] \right)$ and $\bbE[\nabla \zeta(\theta_0)]=- \sum_{i=1}^{n} h_{\eta}'(x_i-\theta_0)$ because $\bbE[\nabla \sL(\theta_0)]=0$. 
We assume that $h_\eta$ is twice continuously differentiable and let
$\mathtt{h}^2:= -\int h_\eta''(z)p_{\eta}(z)dz <\infty$.
Also, 
\begin{align}
\sD_0^2 = \sD_0^{2}(\theta_0)
     = 
    - \nabla^{2} \bbE[ \sL(\theta_0)] = -\sum_{i=1}^{n} \bbE[h_{\eta}''(x_i-\theta_0)]  = n\mathtt{h}^2.
    \end{align}
Similarly, $\sD_0^{2}(\theta) = -\sum_{i=1}^{n} \bbE[h_{\eta}''(x_i-\theta)]a_s  = -\sum_{i=1}^{n} \bbE[h_{\eta}''(\eta + (\theta_0-\theta))]  $. Moreover, $\sV_0^2=\sD_0^2/n= \mathtt{h}^2$.
Next, we assume two other technical conditions on the error density $p_{\eta}$.
\begin{assumption}~\label{ass:AddSG}
\begin{enumerate}
    \item There exists some constant $v_0$ and $\mathtt{g}_1>0$, such that a random variable $\eta \sim p_\eta$, it holds that
    \begin{align}
        \log \bbE \exp(\mu h'_{\eta}(\eta)/ \mathtt{h}) \leq v_0^2\mu^2/2, \quad  |\mu|<\mathtt{g}_1.
    \end{align}
    \item There exists some constant $v_0$ and for every $\mathtt{u}>0$, there exists $\mathtt{g}_1(\mathtt{u})>0$, such that for all $\delta$ with $|\delta|<\mathtt{u}/\mathtt{h}$ it holds that
    \begin{align}
        \log \bbE \left[\exp
    \left(\frac{\mu}{\mathtt{h}} \left\{h'_{\eta}(\eta+\delta) -\bbE[h'_{\eta}(\eta+\delta)] \right\}\right) \right] \leq v_0^2\mu^2/2, \quad  |\mu|<\mathtt{g}_1(\mathtt{u}).
    \end{align}
\end{enumerate}   
\end{assumption}
The above assumption essentially requires that the error distribution has an exponentially decaying tail. Since we assume that the error distribution is sub-Gaussian due to Assumption~\ref{ass:reward}, the above condition is automatically satisfied.
\begin{proposition}~\label{prop:SpokSG}
    The sub-Gaussian family of distributions (Definition~\ref{ass:reward}) satisfies Assumption~\ref{ass:Spokoiny} under Assumption~\ref{ass:AddSG}.
\end{proposition}
\begin{proof}
    We derive all the conditions in seriatim. The proofs are adapted from~\cite{Panov2015}.
  \begin{enumerate}
    \item \({(e\!d_{0})} \):
        Using the definition of $\nabla \zeta_1(\theta_0)$ and $\sV_0$, observe that
        \begin{align}
            \bbE \exp\left\{
              \mathtt{m} \frac{\langle \nabla \zeta_1(\theta_0),\gamma \rangle}
              {\| \sV_0 \gamma \|}
              \right\} &= \bbE \exp\left\{-
              \mathtt{m} \frac{  h_{\eta}'(x_i-\theta_0) }
              {\sV_0}
              \right\}
             = \bbE \exp\left\{- \mathtt{m}\frac{h_{\eta}'(\eta)}{\mathtt{h}}.
              \right\}
        \end{align}
        The result follows immediately using Assumption~\ref{ass:AddSG}~(1).
    \item Proof for \( {(e\!d_{1})} \):
        Using the definition of $\zeta_1(\theta)$, observe that
        \begin{align}
        \nonumber
            \bbE &\exp\left\{
        \frac{\mathtt{m}}{\omega} \frac{\gamma^{\top}[\nabla \zeta_1(\theta) - \nabla \zeta_1(\theta_0)  ] }{ \mathtt{u} \|\sV_0 \gamma\|}
        \right\} 
        \\
        \nonumber
        &= \bbE \exp\left\{
        \frac{\mathtt{m}}{\omega} \frac{[h_{\eta}'(\eta)-h_{\eta}'(\eta+\theta_0-\theta) +\bbE[ h_{\eta}'(\eta+\theta_0-\theta)- h_{\eta}'(\eta)]  ] }{ \mathtt{u} \sV_0}
        \right\}
        \\
        \nonumber
        &\le\bbE \exp\left\{
        \frac{\mathtt{m}}{\omega} \frac{\sV_0|\theta-\theta_0|[h_{\eta}''(\eta+\delta) -\bbE[ h_{\eta}''(\eta+\delta)]  ] }{ \mathtt{u} \sV_0^2}
        \right\}
        \\
        &\le \bbE \exp\left\{
        \frac{\mathtt{m}}{\omega} \frac{[h_{\eta}''(\eta+\delta) -\bbE[ h_{\eta}''(\eta+\delta)]  ] }{  \mathtt{h}^2}
        \right\}
        .
        \end{align}
        where in the first inequality, we used the mean-value theorem for $|\delta|\leq \mathtt{u}/\mathtt{h}$, and in the second inequality, we used the definition of $\Theta_0(\mathtt{u})$.
        For $\mu = \frac{\mathtt{m}}{\omega }$ such that $|\mu| \leq  \mathtt{g}_1(\mathtt{u})$, it follows from Assumption~\ref{ass:AddSG}(2), that
        \begin{align}
            \bbE &\exp\left\{
        \frac{\mathtt{m}}{\omega} \frac{\gamma^{\top}[\nabla \zeta_1(\theta) - \nabla \zeta_1(\theta_0)  ] }{ \mathtt{u} \|\sV_0 \gamma\|}
        \right\}   \leq \exp\left( \frac{v_0^2\mathtt{m}^2}{2\omega^2}  \right).
        \end{align}
        By fixing $\omega=1$, the result follows.
    \item \({(\ell_0)} \):
        First note that 
          \begin{align}
          \nonumber
              -\textsc{KL}(\theta,\theta_0)&= \bbE[\sL_1(\theta)-\sL_1(\theta_0)] 
              \\
              \nonumber
              = \bbE&\left[ h_{\eta}(x_1- \theta)- h_{\eta}(x_1- \theta_0)\right] 
              \\
              \nonumber
              =   \bbE&\left[ -h_{\eta}'(x_1- \theta_0) (\theta-\theta_0) + h_{\eta}''(x_1- \theta^*)(\theta-\theta_0)^2 \right]
              \\
              = \bbE&[h_{\eta}''(x_1- \theta^*)](\theta-\theta_0)^2 = -\sV_0^2(\theta^*)(\theta-\theta_0)^2=I(\theta^*)(\theta-\theta_0)^2,
          \end{align}
          where the second equality uses second-order Taylor's theorem for a $\theta^*$ that lies between $\theta$ and $\theta_0$ and the penultimate inequality uses the fact that $\bbE\left[ -h_{\eta}'(x_1- \theta_0)\right]= \nabla\bbE[\sL_1(\theta_0)]=0$. To show $(\ell_0)$, note that $\left|\frac{\scKL(\theta_0,\theta)}{(\theta-\theta_0)^2I(\theta_0)} -1 \right|= \left|\frac{I(\theta^*)}{I(\theta_0)}-1\right|$ and $I(\cdot)$ is continuous, therefore there must exists a $\delta^*$ such that $\left|\frac{I(\theta^*)}{I(\theta_0)}-1\right|<\delta^* \mathtt{u}$.
    \item \( {(\ell{\mathtt{u}})} \):
      Using the expression for $\textsc{KL}(\theta,\theta_0)$ derived to prove $\ell_0$, we have $\frac{\textsc{KL}(\theta,\theta_0)}{\|\sV_0 (\theta - \theta_0)\|^2} = \frac{I(\theta^*)}{I(\theta_0)}$ for some $\theta^*$ between $\theta$ and $\theta_0$. Since, $\theta$ must satisfy $\|\sV_0(\theta-\theta_0)\|=u$,
        $\theta^*$ is a mapping from $\mathtt{u}$ and so is $\frac{I(\theta^*)}{I(\theta_0)}$. Therefore, we can always find a mapping $\mathtt{b}(\mathtt{u})$ such that $\frac{I(\theta^*)}{I(\theta_0)}\geq \mathtt{b}(\mathtt{u})$ for any $\mathtt{u}>0$. 
    \item \({(e\mathtt{u})} \): Using the definition of $\nabla \zeta_1(\theta)$ and $\sV_0$, observe that
        \begin{align}
        \nonumber
            \bbE \exp\left\{
              \mathtt{m} \frac{\langle \nabla \zeta_1(\theta),\gamma \rangle}
              {\| \sV_0 \gamma \|}
              \right\} 
              &= \bbE \exp\left\{-
              \mathtt{m} \frac{  h_{\eta}'(x_i-\theta) - \bbE[h_{\eta}'(x_i-\theta)] }
              {\sV_0}
              \right\}
              \\
              \nonumber
             &= \bbE \exp\left\{- \mathtt{m}\frac{h_{\eta}'(\eta+ (\theta-\theta_0)) - \bbE[h_{\eta}'(\eta+(\theta-\theta_0))] }{\mathtt{h}}
              \right\}.
        \end{align}
    Now the result follows immediately using Assumption~\ref{App:Assumption34}(2).
  \end{enumerate}
  \end{proof}

\subsection{More examples  and comparisons with baseline algorithms}\label{app:More}

\begin{figure}[ht]
\centering
    \includegraphics[clip,trim=0.3cm 0cm 0.0cm 0cm,width=0.49\textwidth]{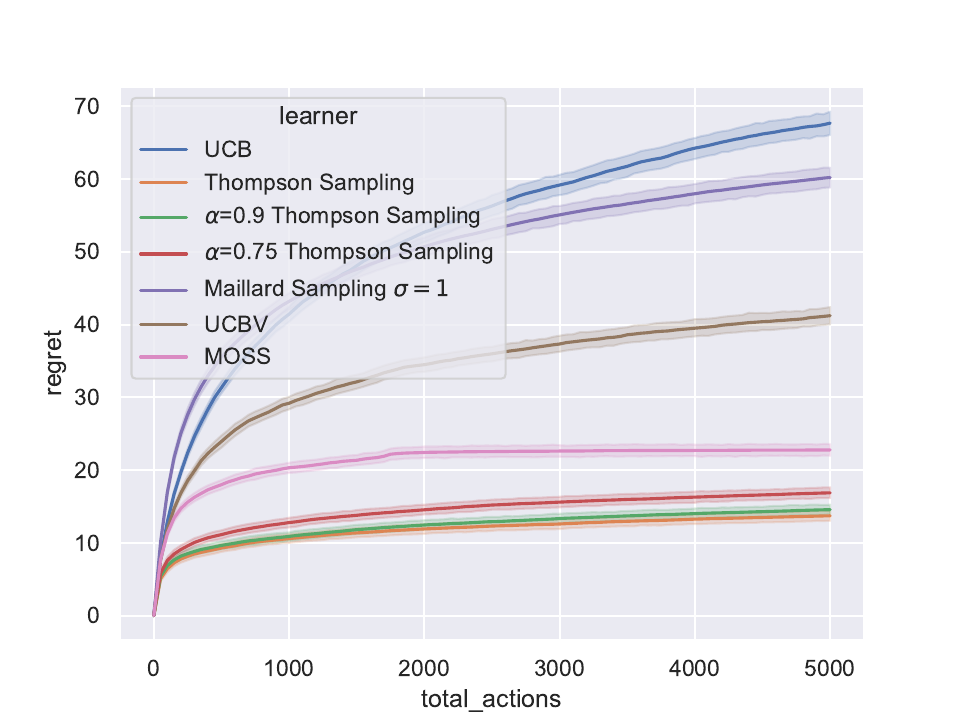}
    \includegraphics[width=0.49\textwidth]{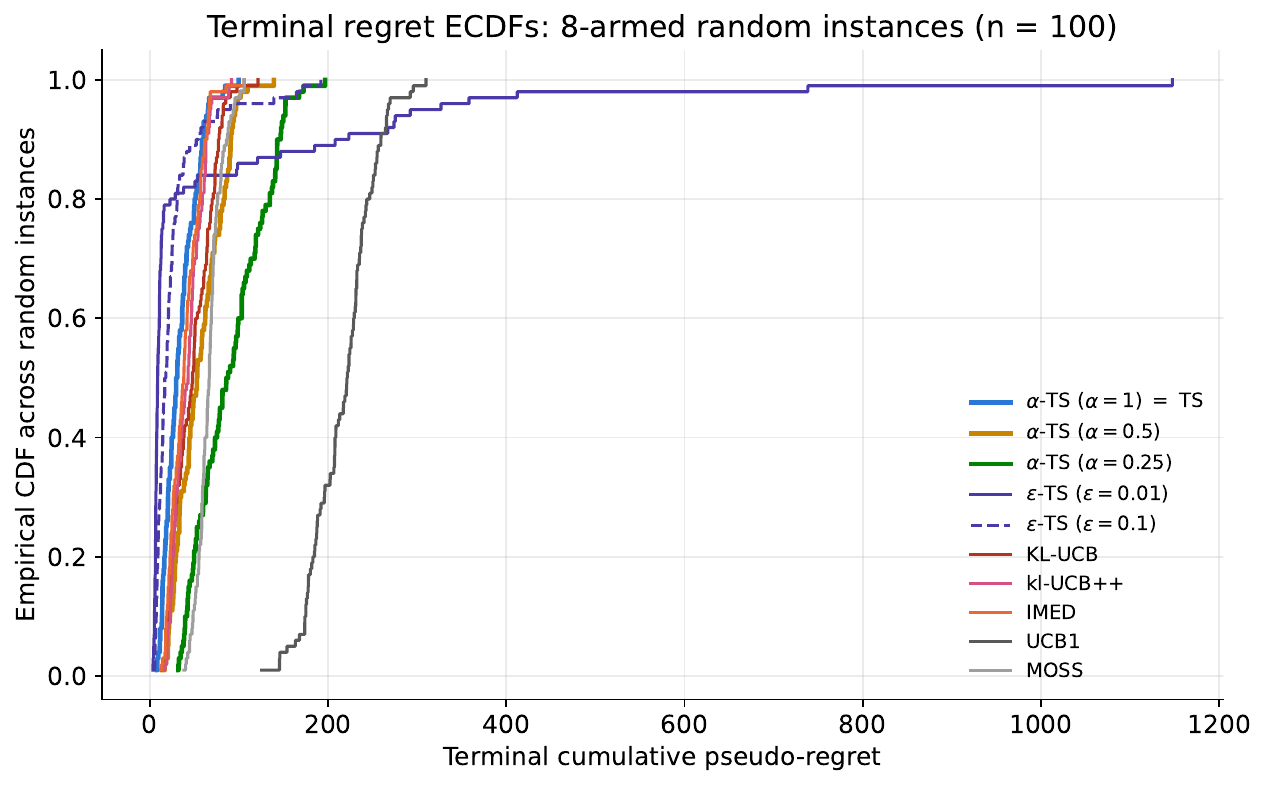}
    \caption{(left) Regret plot for Beta-Bernoulli $\alpha$-TS and Bernoulli UCB~\cite{Auer2002}, UCBV~\cite{audibert2009exploration}, MOSS~\cite{audibert2009minimax} for  rewards with 3 arms having unique true mean rewards. Solid lines are the median regrets with ($10^{th}$ and $90^{th}$ \%iles) computed using 100 replicates of the experiment (with mean rewards $0.3,0.33,0.7$). (right) Empirical CDFs of terminal regret for 100 randomly generated mean rewards with 8 arms
    }
        \label{fig:BernoulliU}
\end{figure}

We compare the empirical performance of $\alpha$-TS (Beta prior) with UCB, MOSS, UCBV for Bernoulli rewards and plot it in~Figure~\ref{fig:BernoulliU}.

\paragraph{Comparison with modern baselines: KL-UCB, kl-UCB++, IMED, and $\varepsilon$-TS.}\label{app:Baselines}

\begin{figure}[ht]
\centering
    \includegraphics[width=0.99\textwidth]{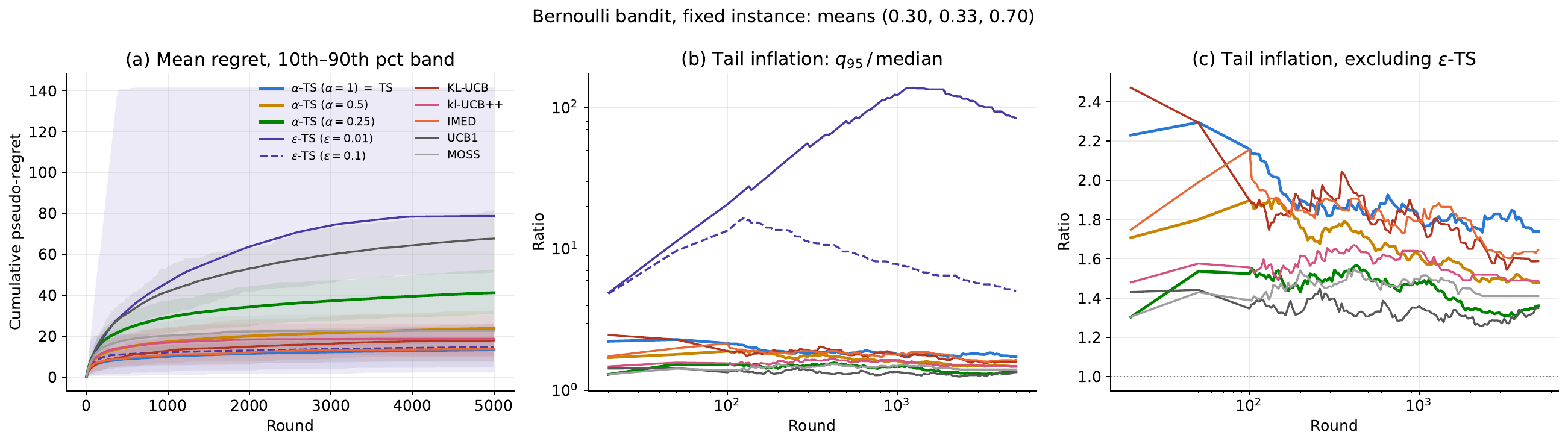}
    \caption{Fixed Bernoulli instance (means $0.3, 0.33, 0.7$). (a) Mean cumulative regret with $10$th--$90$th percentile bands. (b) Tail inflation $q_{95}/\mathrm{median}$ (log scale). (c) The same ratio excluding $\varepsilon$-TS. Standard TS ($\alpha$-TS with $\alpha = 1$, mean terminal regret $13.5$) and IMED ($13.8$) lead, ahead of KL-UCB ($18.1$) and kl-UCB++ ($18.8$). $\varepsilon$-TS at $\varepsilon = 0.01$ has the best median ($5.0$) but by far the heaviest tail ($q_{95} = 424$, inflation $84\times$); at $\varepsilon = 0.1$ the inflation is $5.1\times$, still triple that of any well-explored method ($1.3$--$1.7\times$).}
        \label{fig:Baselines}
\end{figure}

We extend the comparison above with the standard asymptotically optimal index policies for Bernoulli bandits, namely KL-UCB~\citep{garivier2011kl}, kl-UCB++~\citep{menard2017minimax}, and IMED~\citep{honda2015imed}, and with $\varepsilon$-exploring Thompson Sampling ($\varepsilon$-TS)~\citep{jin2023eps}, which plays greedily on empirical means with probability $1-\varepsilon$ and uses a posterior sample with probability $\varepsilon$. All policies are run on the fixed instance of Figure~\ref{fig:BernoulliU} (means $0.3, 0.33, 0.7$, $T = 5{,}000$, $100$ replications on shared reward tables, so comparisons are paired). Figure~\ref{fig:Baselines} reports the mean regret with $10$th--$90$th percentile bands, together with the \emph{tail-inflation} ratio $q_{95}/\mathrm{median}$ of cumulative regret, a scale-free heavy-tail diagnostic: an algorithm that is merely slow (UCB1) has a ratio near $1$, whereas an algorithm whose rare runs are catastrophically worse than its typical run has a large ratio.

\begin{figure}[ht]
\centering
        \includegraphics[width=0.6\textwidth]{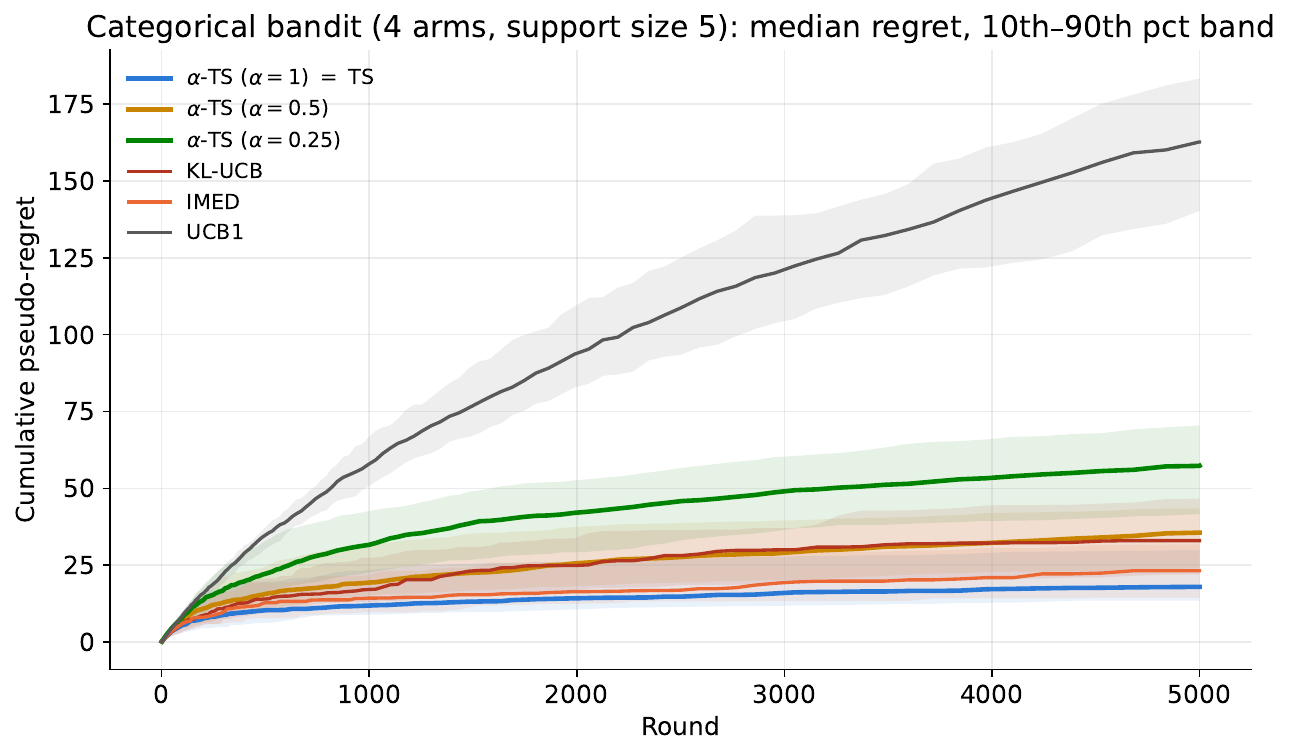}
    \caption{Categorical bandit ($4$ arms, support $\{0, 0.25, 0.5, 0.75, 1\}$): median cumulative regret with $10$th--$90$th percentile bands for Dirichlet--Categorical $\alpha$-TS, empirical KL-UCB, IMED (both via $\mathcal{K}_{\inf}$), and UCB1, over $40$ replications.}
        \label{fig:CatBaselines}
\end{figure}

\begin{example}{Dirichlet-Categorical MAB}

        In this example, we model the reward function as a Categorical distribution with support $[0,1,2,\ldots d-1]$, that is for any $k\in [K]$, $r_k \sim Categorical(\theta_k)$, where $\theta_k$ lies in a $d$-dimensional simplex, and $\theta_{0,k}$ is the true ( but unknown) parameter.
        We posit a $\text{Dirichlet}(\textbf{a})$ prior on $\theta_k$, for each arm with $\textbf{a} =\{1\}^d$. Let $c_k(t) \in \bbN^d$ denote the vector recording the number of time a category is sampled from the $k^{th}$ arm. Using these notations, the $\alpha$-posterior distribution for each arm $k$ can be expressed as
        $\pi_{\alpha}(\theta_k|X_{t},\alpha-\text{TS}) \equiv \text{Dirichlet}(\textbf{a}+\alpha*c_k(t))$. Note that in this example $\mu_k=\sum_{i=0}^{d-1}i\theta_k^{i+1}$. 
        In Figure~\ref{fig:CatBaselines}, we plot the Regret$(T,\alpha-\text{TS})$ for various values of $\alpha$, with the support rescaled to $[0,1]$ and together with the empirical KL-UCB and IMED baselines described above.

\end{example}

Two observations. First, the stronger baselines sharpen rather than change the comparative picture: TS-type methods match or beat the KL-UCB family, and the same ordering holds in the categorical setting, where we implement \emph{empirical} KL-UCB and IMED for bounded rewards via the $\mathcal{K}_{\inf}$ optimization of~\citet{honda2010asymptotically,cappe2013kullback} (Figure~\ref{fig:CatBaselines}): Dirichlet--Categorical TS (mean terminal regret $21.0$) leads IMED ($23.1$) and empirical KL-UCB ($33.8$) at a fraction of their per-round computational cost. Second, $\varepsilon$-TS is an extreme point of an exploration--tail trade-off: trimming exploration buys average-case speed and pays in the upper quantiles, because a nearly greedy policy that locks onto a suboptimal arm generates almost no exploration with which to correct itself. 
Appendix~\ref{app:AlphaRole} shows the corresponding behavior of $\alpha$-TS as $\alpha$ varies. Notably, panel (c) of Figure~\ref{fig:Baselines} shows that within the $\alpha$-TS family tail inflation \emph{decreases} with $\alpha$ ($1.7$ at $\alpha = 1$ down to $1.4$ at $\alpha = 0.25$): tempering compresses the regret distribution even in this flat-prior instance where it costs mean regret.

\subsection{The role of $\alpha$: prior retention and tail risk}\label{app:AlphaRole}

\begin{figure}[ht]
\centering
    \includegraphics[width=0.99\textwidth]{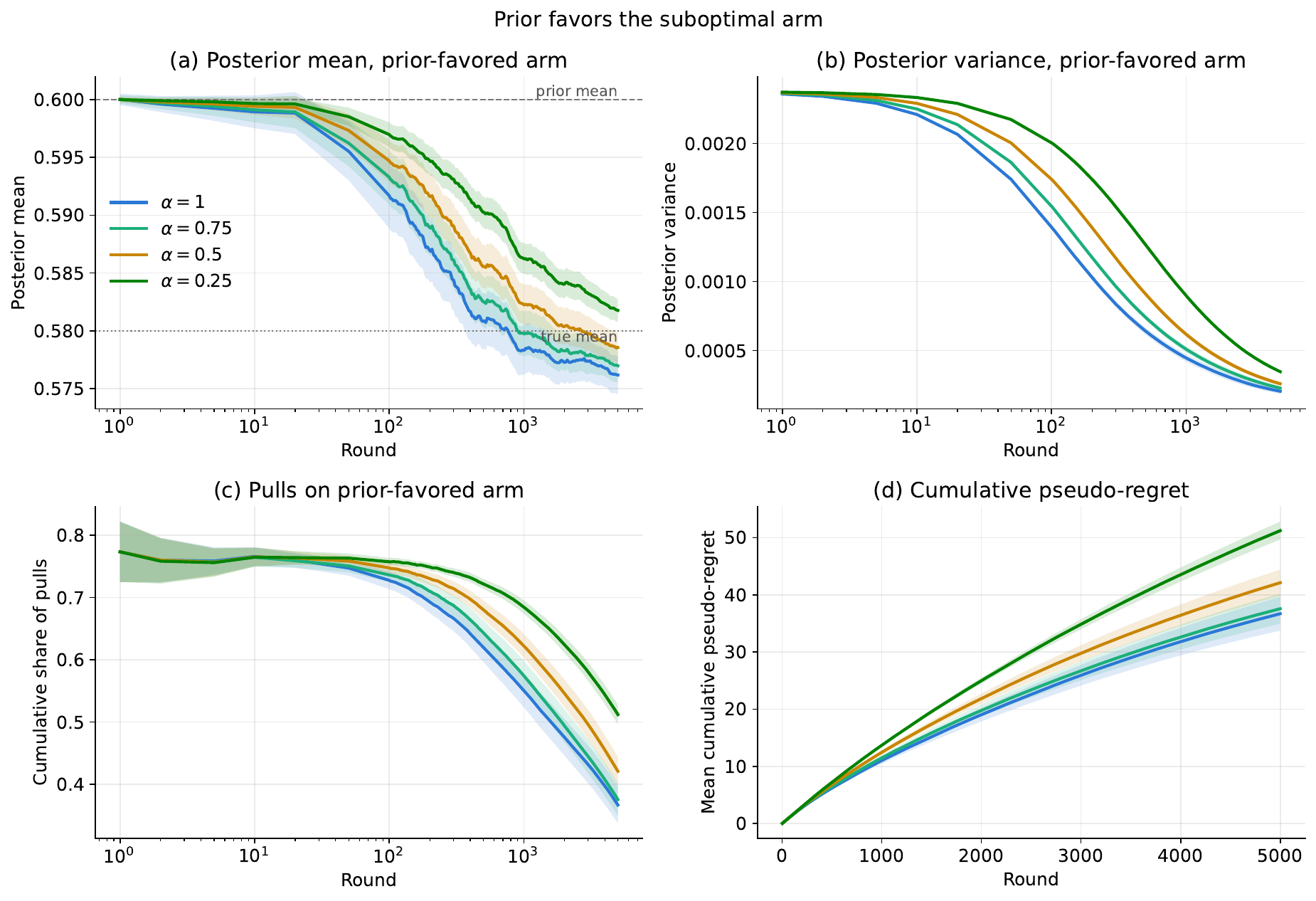}
    \caption{Informative prior favoring the suboptimal arm. (a) Posterior mean and (b) posterior variance of the prior-favored arm: smaller $\alpha$ moves beliefs toward the truth more slowly (effective sample size $\alpha n$). (c) Cumulative share of pulls on the prior-favored arm: belief-level retention becomes action-level retention. (d) Mean cumulative pseudo-regret, ordered in $\alpha$ (terminal means $36.7$ at $\alpha = 1$ vs.\ $51.2$ at $\alpha = 0.25$). Bands are $\pm 2$ standard errors. In the aligned environment (prior favoring the true best arm; not shown) mean terminal regret is statistically flat in $\alpha$ ($21.7$--$22.6$) while the $95$th percentile \emph{falls} from $59.2$ to $41.2$ as $\alpha$ decreases: retaining a correct prior is free on average and strictly better in the tail.}
        \label{fig:PriorRetention}
\end{figure}

This appendix develops in full the experiments summarized in Section~\ref{sec:Main Results}, which examine the role of $\alpha$ as the prior-retention dial introduced in Section~\ref{sec:AlphaTS}. In the spirit of Figure~3 of~\citet{chapelle2011empirical}, we visualize how $\alpha$ influences behavior and regret in a two-armed Bernoulli instance with gap $\Delta = 0.02$ and an informative prior of strength $100$ pseudo-observations per arm ($T = 5{,}000$, $300$ paired replications), under two environments that differ only in whether the prior favors the true best arm. Figure~\ref{fig:PriorRetention} traces the mechanism round by round, and Figure~\ref{fig:TailAlpha} (reported in the main text) summarizes the resulting mean and upper-quantile regret as a function of $\alpha$.

\begin{figure}[ht]
\centering
    \includegraphics[width=\textwidth]{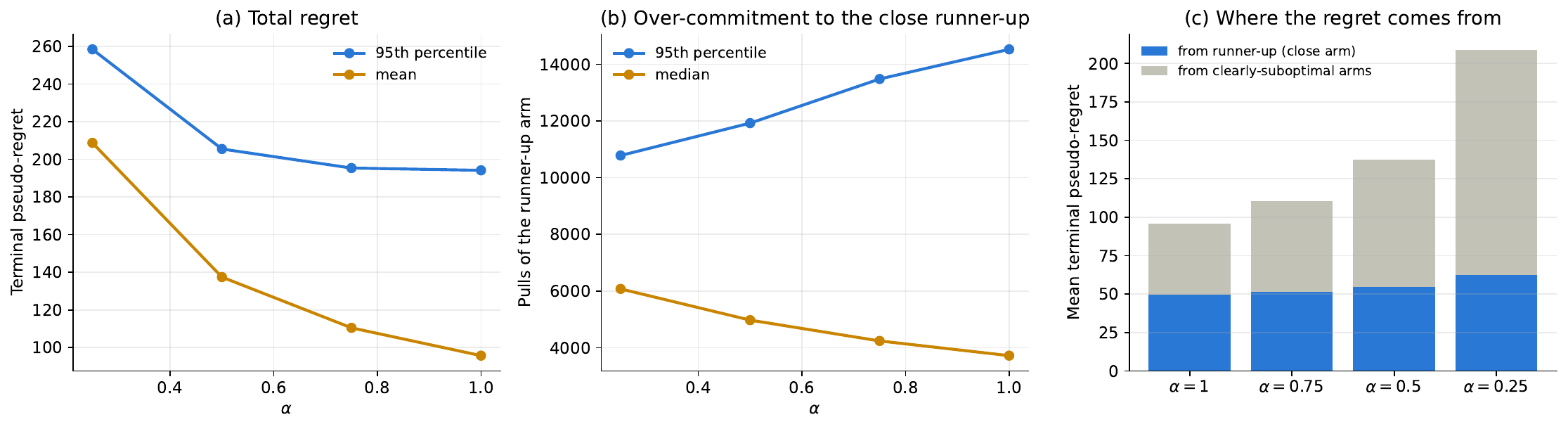}
    \caption{Eight arms with true means $(0.70, 0.69, 0.40, 0.35, 0.30, 0.25, 0.20, 0.15)$, non-informative prior, $T = 20{,}000$, $800$ replications. (a) Mean and $95$th percentile of total terminal regret: both increase as $\alpha$ decreases. (b) Pulls of the close runner-up: the $95$th percentile \emph{decreases} with $\alpha$, so tempering reduces over-commitment in the unlucky runs, while the median increases. (c) Decomposition of mean terminal regret: the contribution of the close runner-up is roughly stable across $\alpha$, whereas the contribution of the clearly-suboptimal arms grows sharply, which is what drives the total in panel (a).}
        \label{fig:ManyArm}
\end{figure}

\paragraph{Why the two-armed instance of Figure~\ref{fig:Bernoulli} isolates the effect.} In a two-armed bandit, confusing the two arms is the only way to incur regret, so the mechanism above is visible directly in the total regret. With more arms the same mechanism still operates, but it is masked. Figure~\ref{fig:ManyArm} repeats the experiment on eight arms whose top two are close, $(0.70, 0.69)$, and whose remaining arms are clearly worse (all at most $0.40$, so they are eliminated quickly), over a horizon $T = 20{,}000$ long enough for the tempered versions to converge as well. Panel (b) shows that tempering does exactly what one would expect of it: the distribution of pulls on the close runner-up concentrates, with the $95$th percentile falling from $14{,}529$ at $\alpha = 1$ to $10{,}782$ at $\alpha = 0.25$, a reduction of $26\%$ in the worst runs, while the median rises. The share of regret attributable to that arm falls correspondingly from $52\%$ to $30\%$. Panel (a) nevertheless shows the $95$th percentile of \emph{total} regret increasing as $\alpha$ decreases, because, as panel (c) makes explicit, tempering pays for this protection by exploring the six clearly-suboptimal arms far longer, and that cost outweighs what is saved on the runner-up. The tail benefit is therefore visible in total regret precisely when confusion between the leading arms is the dominant contributor to regret, and it is otherwise present but hidden.

Together these experiments yield a practical guideline for selecting the tempering parameter: choose $\alpha$ by prior trust and risk preference. A vetted informative prior argues for small $\alpha$, which leaves mean regret essentially unchanged while reducing upper-quantile regret; an unvetted prior with average performance as the objective argues for $\alpha$ near $1$; when tail risk matters more than average performance, moderate tempering $\alpha \in [0.2, 0.5]$ buys a large reduction in upper-quantile regret at a quantifiable mean cost. Extreme tempering ($\alpha \lesssim 0.1$) is dominated in the Pareto sense: it reduces the regret tail no further than moderate tempering, because the tail saturates at the level of an algorithm that never updates away from its prior, while its mean cost keeps growing, so no trade-off between average and tail performance favors it.

\subsection{Beyond conjugate priors: Gaussian-mixture and logit-normal priors}\label{app:Fancy}

Our analysis requires no conjugacy, and $\alpha$-TS runs unchanged whenever one $\alpha$-posterior draw per arm per round is available. We illustrate this with the two prior families below; in both, the prior-retention behavior of Appendix~\ref{app:AlphaRole} transfers quantitatively.

\begin{figure}[ht]
\centering
    \includegraphics[width=0.99\textwidth]{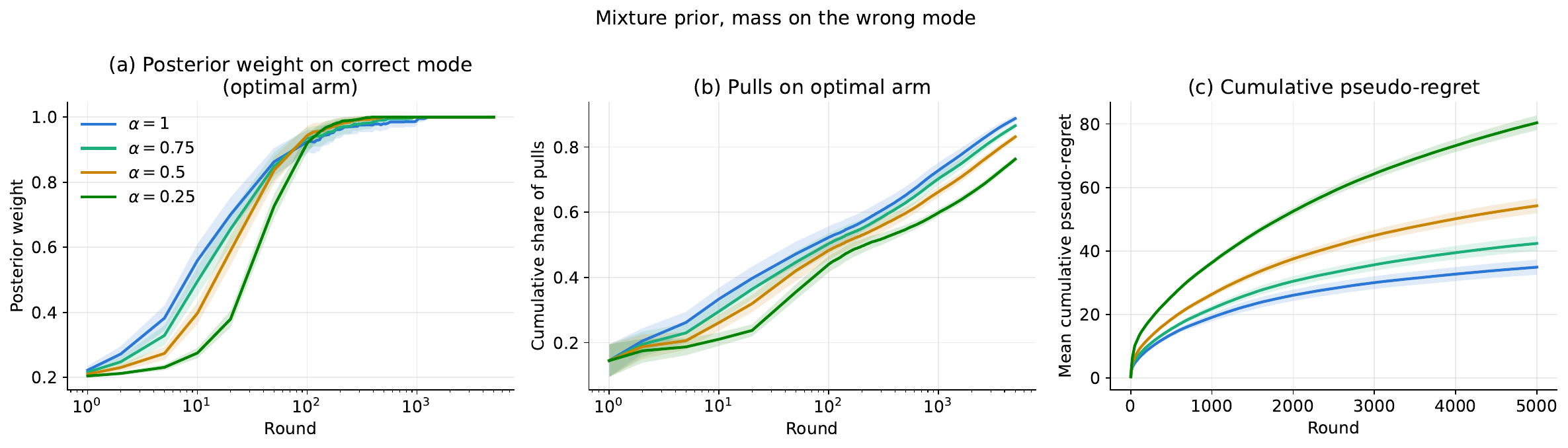}
    \caption{Gaussian-mixture prior with mass $0.8$ on the wrong mode of each arm. (a) Posterior weight on the correct mode of the optimal arm: the mode switch is delayed by $\approx 1/\alpha$, the mixture analog of prior retention. (b) Cumulative optimal-arm fraction. (c) Mean cumulative regret. Bands are $\pm 2$ standard errors over $200$ replications.}
        \label{fig:Mixture}
\end{figure}

\textbf{Gaussian-mixture prior (mode retention).} Arms pay $X \sim \mathcal{N}(\mu_k, 0.5^2)$ and each arm's mean carries the bimodal prior $w_{\mathrm{lo}}\mathcal{N}(0.2, 0.05^2) + w_{\mathrm{hi}}\mathcal{N}(0.8, 0.05^2)$: ``the arm either works or it does not,'' the structured prior family of~\citet{hong22b} discussed in Section~2. The $\alpha$-posterior of a mixture prior is again a mixture: each component updates by Gaussian conjugacy at effective sample size $\alpha n$ and the mixture weights update analytically through the component evidences, so posterior sampling is exact. With true means $(0.80, 0.75, 0.20)$ and prior mass $0.8$ on the \emph{wrong} mode of every arm, the posterior must switch modes before acting well, and the evidence required scales as $1/\alpha$ (Figure~\ref{fig:Mixture}): mean terminal regret over $200$ replications orders from $34.9$ ($\alpha = 1$) to $80.4$ ($\alpha = 0.25$).

\begin{figure}[ht]
\centering
    \includegraphics[width=0.85\textwidth]{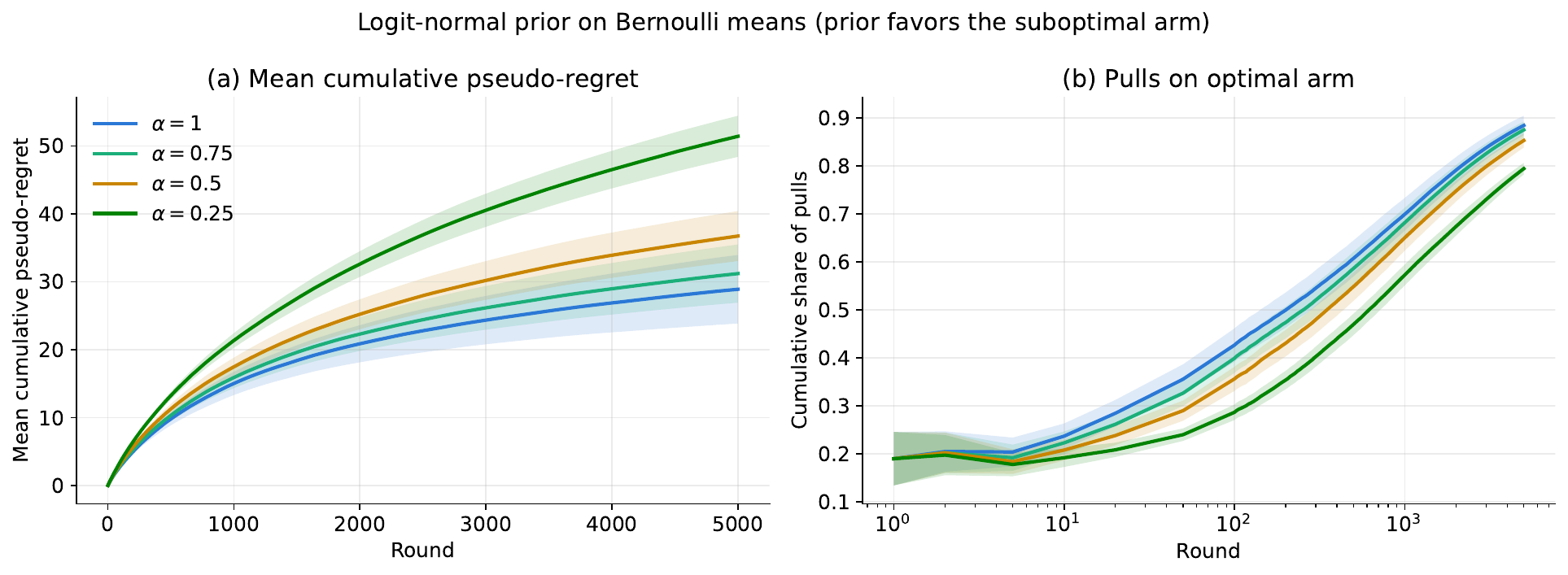}
    \caption{Logit-normal prior on Bernoulli means (non-conjugate; grid-quadrature sampling), with the prior favoring the suboptimal arm. (left) Mean cumulative regret. (right) Cumulative optimal-arm fraction. Bands are $\pm 2$ standard errors over $200$ replications.}
        \label{fig:LogitNormal}
\end{figure}

\textbf{Logit-normal prior (genuinely non-conjugate).} Bernoulli rewards with $\theta_k = (1 + e^{-Z_k})^{-1}$, $Z_k \sim \mathcal{N}(m_k, 1)$, yield an $\alpha$-posterior in no standard family; we evaluate it on a $2{,}048$-point grid over $(0,1)$ and sample by inverse-CDF, exact to grid resolution and costing microseconds per draw. With true means $(0.65, 0.60)$ and an informative-but-wrong prior ordering (prior centered at $0.4$ for the optimal arm and $0.7$ for the suboptimal one), mean terminal regret over $200$ replications reproduces the conjugate misspecified-prior signature, ordering in $\alpha$ from $28.9$ ($\alpha = 1$) to $51.4$ ($\alpha = 0.25$) (Figure~\ref{fig:LogitNormal}). The retention phenomenon is a property of likelihood tempering itself, not of the Beta--Bernoulli pair.

\paragraph{A note on the assumptions.} Both settings satisfy the conditions of our theory. The Gaussian-mixture setting requires no qualification: Gaussian rewards with known variance are sub-Gaussian with a reward variance that does not depend on the parameter, so the reward-side conditions hold on all of $\Theta = \mathbb{R}$ by the verifications in Appendices~\ref{App:Assumption34} and~\ref{app:Spokoiny}. The Bernoulli setting is the one requiring care, since the reward variance $\theta(1-\theta)$ vanishes at the boundary of $(0,1)$; we therefore take $\Theta = (\epsilon, 1-\epsilon)$ for a small $\epsilon > 0$, which bounds the variance above and below and, since the prior is supported on $\Theta$, restricts the $\alpha$-posterior to $\Theta$ as well, leaving the algorithm unchanged. The restriction is numerically immaterial: the logit-normal priors place mass below $10^{-9}$ outside $(10^{-3}, 1-10^{-3})$, and the quadrature grid used for sampling already lies inside $[2^{-12}, 1-2^{-12}]$, so the implementation coincides with the restricted model. Both priors have continuous, strictly positive, and bounded densities on $\Theta$, which gives Assumption~\ref{Ass:Prior2} directly and the prior-thickness condition (Assumption~\ref{ass:prior2}) via the verification in Appendix~\ref{App:Assumption34}, which covers any bounded prior with continuous density; the Gaussian mixture is precisely the prior class of~\citet{hong22b} discussed in Section~2. More generally, fractional posteriors concentrate under such prior-mass conditions alone, without the testing constructions required by standard posteriors~\citep{bhattacharya2019bayesian,Alquier2020}, which is what makes non-conjugate priors compatible with the analysis.

\section{Way forward for $\alpha=1$ (Standard TS)}~\label{app:StanTS}
To extend our result to $\alpha=1$, the standard TS, the only Lemma that needs to be adapted is Lemma~\ref{lem:Post}, which will require a substantial amount of work and is beyond the scope of this manuscript. We also note that the exclusion of $\alpha = 1$ is consistent with our experiments, in which the upper tail of the regret distribution is heaviest near $\alpha = 1$, even under a non-informative prior (Figures~\ref{fig:TailAlpha}). Next, we provide the guiding sketch of the proof to extend this Lemma for $\alpha=1$ case. We omit $\alpha$ from $\pi_{\alpha}$ to denote standard posterior ($\alpha=1$). The proof essentially follows similar steps up until the display~\eqref{eq:MABDE3} and the definition of the set $A$. As in the proof of Lemma~\ref{lem:Post}, the sketch below is stated at an arbitrary bounded stopping time $\tau \leq T$ and is indexed by the effective count $\bar n_k(\tau)$ rather than by a deterministic time. The adaptive, data-dependent number of pulls is handled exactly as it is there, through the optional-stopping bound of Lemma~\ref{lem:MG}, so that no conditional independence of the observed rewards is used at any point below. Thereafter, we need a slightly different definition of the set $A_{\bar n_k(\tau)}$ which is independent of $\alpha$. In particular, we now fix $A_{\bar n_k(\tau)} = \left\{ X_{k,\tau}: \int_{\Theta} \tilde \pi_k(\theta_k) e^{-\ell_{\bar n_k(\tau)}(\theta_k,\theta_{0,k})}d\theta_k \leq  e^{-D c\, \bar n_k(\tau)\delta_k^2}  \right\}$ for some positive constant $c$. Now we bifurcate the expectation of the set $F_k$ (see~\eqref{eq:MABDE2b}) as
\begin{align}
\nonumber
 \bbE [\Pi(F_k|X_{\tau})] &\leq  \bbE\left[\phi_{\bar n_k(\tau),\Delta_k}\right] + \bbE\left[\bb1(A_{\bar n_k(\tau)})\right] 
 \\
 &\quad + \bbE \left[ (1-\phi_{\bar n_k(\tau),\Delta_k})  \frac{\int_{F_k} \pi_k(\theta_k) e^{- \ell_{\bar n_k(\tau)}(\theta_k,\theta_{0,k})}d\theta_k} {\int \pi_k(\theta_k) e^{- \ell_{\bar n_k(\tau)}(\theta_k,\theta_{0,k})}d\theta_k } \bb1(A_{\bar n_k(\tau)}^C) \right],
 \label{eq:MABDE4C}
\end{align}
where $\phi_{\bar n_k(\tau),\Delta_k} \in [0,1]$ are a sequence of test functions and it follows by dividing the expectation using $\phi_{\bar n_k(\tau),\Delta_k}$ and $\bb1(A_{\bar n_k(\tau)})$ and also using the fact that $\Pi(F_k|X_{\tau})\leq 1$. At a high level, these tests are nothing but the test of an hypothesis that the data is generated from true model or not.  
Interested readers can refer to~\citep{Sc1965,GGV,ZG} for more details regarding these test function and their usage for studying posterior concentration.

First, let us analyze the third term in~\eqref{eq:MABDE4C}. Note that on the set $A_{\bar n_k(\tau)}^C$,
\[\int \pi_k(\theta_k) e^{-\ell_{\bar n_k(\tau)}(\theta_k,\theta_{0,k})}d\theta_k \geq \pi_k(B\left(\theta_{0,k},\delta_k \right))e^{-D c\, \bar n_k(\tau) \delta_k^2} .\]
 Therefore, using the definition of the set $A_{\bar n_k(\tau)}^C$, the third term in~\eqref{eq:MABDE4C} can be bounded by
\begin{align}
\nonumber
   &\frac{e^{D c\, \bar n_k(\tau) \delta_k^2}}{\pi_k(B\left(\theta_{0,k},\delta_k \right))} \bbE \left[ \left(1-\phi_{\bar n_k(\tau),\Delta_k}\right)  {  \int_{F_k} \pi_k(\theta_k) e^{-\ell_{\bar n_k(\tau)}(\theta_k,\theta_{0,k})}d\theta_k } \right]
   \\
   \nonumber
   &\leq  \frac{e^{D c\, \bar n_k(\tau) \delta_k^2}}{\pi_k(B\left(\theta_{0,k},\delta_k \right))} \int_{F_k} \pi_k(\theta_k) \bbE \left[ \left(1-\phi_{\bar n_k(\tau),\Delta_k}\right)     e^{-\ell_{\bar n_k(\tau)}(\theta_k,\theta_{0,k})}\right] d\theta_k  
   \\
   \nonumber
   &=  \frac{e^{D c\, \bar n_k(\tau) \delta_k^2}}{\pi_k(B\left(\theta_{0,k},\delta_k \right))} \int_{F_k} \pi_k(\theta_k) \bbE_{P_{\theta_k}} \left[ \left(1-\phi_{\bar n_k(\tau),\Delta_k}\right)  \right] d\theta_k 
   \\
   &\leq \frac{e^{D c\, \bar n_k(\tau) \delta_k^2}}{\pi_k(B\left(\theta_{0,k},\delta_k \right))} \left( \int_{F_k\bigcap \Theta(\Delta_k)} \pi_k(\theta_k) \bbE_{P_{\theta_k}} \left[ \left(1-\phi_{\bar n_k(\tau),\Delta_k}\right)  \right] d\theta_k + \pi_k\left(\Theta(\Delta_k)^C\right)\right),
   \label{eq:MABDE11C}
\end{align}
where the first inequality follows using Fubini's theorem by interchanging integral and expectation, and the first equality uses the definition (after display~\eqref{eq:MABDE3}) of $e^{-\ell_{\bar n_k(\tau)}(\theta_k,\theta_{0,k})}$. Note the $\bbE_{P_{\theta_k}}$ in the first equality is with respect to any model $P_{\theta_k}$, not the true model. The set $\Theta(\Delta_k)$ is a subset of $\Theta_k$. 
The second term in~\eqref{eq:MABDE4C} can be bounded using similar steps as used in~\eqref{eq:MABDE8}. Bounding~the first and the third term (through~\eqref{eq:MABDE11C}) requires two separate assumptions as used in~\cite{GGV,ZG,Yang2020}. The first condition assumes that, at every fixed sample size $n \geq n_{k,0}$, $\bbE_{P_{\theta_{0,k}}}\left[\Pi(F_k|X_{k,n})\right]$ and $\bbE_{P_{\theta_k}} \left[ 1-\phi_{n,\Delta_k}\right]$, for all $\theta_k \in F_k\cap \Theta(\Delta_k)$, are bounded by $e^{-C_1 n\Delta_k^2}$, and the second condition would require bounding $\nicefrac{\pi_k\left(\Theta(\Delta_k)^C\right)}{\pi_k(B\left(\theta_{0,k},\delta_k \right))}$ also by $e^{-C_2 n\Delta_k^2}$, for some positive constants $C_1$ and $C_2$. Passing from these fixed sample size statements to the stopping time $\tau$ is then precisely the optional-stopping step of Lemma~\ref{lem:MG}, which is what delivers the exponentially compensated form in which Lemma~\ref{lem:Post} is stated. Then, the proof steps will follow a similar argument as used to derive~\eqref{eq:MABDE7}. The details about this assumption can be found in~\cite {ZG,Yang2020}, which are motivated by the pioneering work in~\cite{GGV}. Also, verifying these new assumptions for exponential family and sub-Gaussian distribution requires a separate set of arguments.

\end{document}